\documentclass{article}
\usepackage{arxiv}

\usepackage[T1]{fontenc}
\usepackage[utf8]{inputenc}
\usepackage{amsmath,amssymb,amsthm}
\usepackage{graphicx}
\usepackage{booktabs}
\usepackage{tabularx}
\usepackage{xcolor}
\usepackage{caption}
\usepackage[round,authoryear]{natbib}
\usepackage[colorlinks=true,linkcolor=blue!60!black,citecolor=blue!60!black,urlcolor=blue!60!black]{hyperref}
\usepackage{authblk}

\usepackage{subcaption}
\usepackage{makecell}
\usepackage{multirow}
\usepackage{float}
\usepackage{wrapfig}

\newtheorem{proposition}{Proposition}

\newcommand{\Prob}{\mathbb{P}}

\newcommand{\z}{\mathbf{z}}
\newcommand{\yv}{\mathbf{y}}
\newcommand{\C}{\mathcal{C}}
\newcommand{\bhat}{\hat{b}}
\newcommand{\x}{\mathbf{x}}

\begin{document}

\title{Quantile-Based Sequential Learning and Optimization}

\renewcommand\Authfont{\bfseries}
\setlength{\affilsep}{0em}
\newbox{\orcid}\sbox{\orcid}{\includegraphics[scale=0.06]{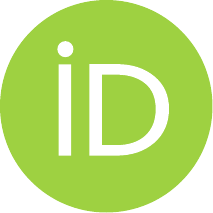}}
\author[1]{\href{https://orcid.org/0009-0005-2870-9825}{\usebox{\orcid}\hspace{1mm}Nele Bertling}\thanks{Corresponding author: \texttt{nele.bertling@univie.ac.at}}}
\author[1]{\href{https://orcid.org/0000-0002-5931-4907}{\usebox{\orcid}\hspace{1mm}Kevin Tierney}\thanks{\texttt{kevin.tierney@univie.ac.at}}}
\author[2]{\href{https://orcid.org/0000-0001-8369-7939}{\usebox{\orcid}\hspace{1mm}Michael R{\"o}mer}\thanks{\texttt{michael.roemer@uni-bielefeld.de}}}
\affil[1]{University of Vienna, Vienna, Austria}
\affil[2]{Bielefeld University, Bielefeld, Germany}
\renewcommand{\shorttitle}{\textit{Quantile-Based Sequential Learning and Optimization}}
\hypersetup{
pdftitle={Quantile-Based Sequential Learning and Optimization},
pdfauthor={Nele Bertling, Kevin Tierney, Michael R{\"o}mer},
}
\date{}

\maketitle

\begin{abstract}
Predict-then-optimize problems arise in resource allocation, production planning, and scheduling, where uncertain parameters must be predicted from contextual information before a decision is made. 
Unlike uncertainty in objective coefficients, uncertainty in the constraints can render solutions infeasible, and the cost of correcting must be accounted for. 
Integrated learning and optimization (ILO) methods address this by embedding the optimization problem into the training process, which requires repeated optimization and problem-specific differentiable approximations of the otherwise non-differentiable decision loss.
This paper proposes a Quantile-based Sequential Learning and Optimization (QSLO) approach that predicts a conditional quantile of the uncertain parameters and uses the quantile level as a single control parameter for the feasibility vs. (post-hoc) regret trade-off, which, unlike a mean prediction, can be adjusted as correction costs change.
Compared against ODECE, a state-of-the-art feasibility-aware ILO method, QSLO significantly outperforms it on two multi-dimensional knapsack variants with uncertain weights or capacities and performs comparably on a brass alloy production problem with uncertain ore quantities, while training one to almost three orders of magnitude faster when the inner optimization is expensive.
These results indicate that choosing a suitable quantile and a model that fits the data is sufficient for a competitive trade-off, without the differentiable learning pipelines advocated in the ILO literature.

\end{abstract}

\section{Introduction}\label{QSLO::intro}
Routing, resource allocation, and other combinatorial optimization (CO) problems frequently involve uncertain input parameters that depend on \textit{contextual information} such as the time of day, weather conditions, demand patterns, or operational environments. This class of problems is defined as contextual stochastic (combinatorial) optimization (CSO) problems by \citep{sadanaSurveyContextualOptimization2025}. One of the central challenges for solving these problems is that decisions must be made before the true parameter values are revealed, yet the quality of a decision depends directly on how well the uncertainty is accounted for. Combining machine learning (ML) with combinatorial optimization methods to predict uncertain parameters and inform optimization decisions has emerged as a natural and increasingly popular approach to this challenge, giving rise to two broad paradigms.

In the sequential learning and optimization (SLO) framework, prediction and decision-making are treated as separate stages. Within SLO, one can either learn a conditional distribution of the uncertain parameters and generate scenarios to solve a stochastic programming (SP) model, or use a point estimate in a deterministic model, also known as the predict-then-optimize (PtO) pipeline \citep{sadanaSurveyContextualOptimization2025}. While SP models provide theoretically grounded solutions under uncertainty \citep{birge1997introduction, Shapiro2014Lectures}, they scale poorly with the number of scenarios, particularly for combinatorial problems where the extensive form quickly becomes intractable \citep{Kleywegt2002The}. The PtO pipeline instead uses machine learning to inform the parameter estimates within a deterministic model, offering a scalable alternative that is widely used in operations research and industry.

However, prediction errors in PtO generally translate into suboptimal decisions, and empirical evidence shows that minimizing prediction loss does not reliably minimize decision loss \citep{mandiDecisionFocusedLearningFoundations2024, tangPyEPOPyTorchbasedEndend2024}. These limitations motivate the second paradigm, integrated learning and optimization (ILO), also referred to as predict-and-optimize (PnO) or decision focused learning (DFL).  ILO methods directly minimize a decision-aware loss, often defined as \textit{regret}, by integrating the optimization problem into the learning process. They typically rely on differentiable surrogates (problem-specific, differentiable approximations of the otherwise non-differentiable decision loss) or perturbations during training to remain compatible with gradient-based learning \citep{mandiDecisionFocusedLearningFoundations2024, sadanaSurveyContextualOptimization2025, elmachtoubSmartPredictThen2022, berthetLearningDifferentiablePerturbed2020}.

Most ILO methods focus on uncertainty in objective coefficients, as uncertainty in constraint parameters introduces the additional challenge that prediction errors may render solutions infeasible \citep{mandiDecisionFocusedLearningFoundations2024}. Infeasible decisions will inevitably occur and must be systematically handled \citep{Kononova2022The, Diz2018A}, and the cost of restoring feasibility can be substantial. Since strictly enforcing feasibility often leads to overly conservative solutions \citep{Meselhi2022Modified, Jatty2023Diagnosis}, evaluating methods solely on feasibility or \textit{regret} of feasible solutions risks optimizing for a criterion that does not reflect the true operational cost.

To address this, \citet{huBranchLearnPosthoc2023, huPredict+OptimizePackingCovering2023, hu2023twostage, huTwoStagePredict+OptimizeMixed2023} introduce the \textit{post-hoc regret}, which explicitly accounts for correction costs and reduces to standard regret when solutions are feasible, providing a more complete and practically meaningful evaluation criterion. However, its exact computation relies on recursive solvers and problem-specific correction and penalty functions. To avoid the need for such domain knowledge, \citet{mandiFeasibilityAwareDecisionFocusedLearning2025} propose ODECE,
a feasibility-aware ILO method that balances infeasibility and regret through a tunable infeasibility aversion coefficient, without requiring domain-specific correction mechanisms. 
Nevertheless, both approaches require an inner optimization loop during training, which limits their scalability.

ILO methods incur substantial computational overhead by requiring the optimization problem to be solved and differentiated for each training instance in each epoch, and in practice they are typically trained with linear or shallow models~\citep{mandiFeasibilityAwareDecisionFocusedLearning2025,tangPyEPOPyTorchbasedEndend2024}. Backpropagation through the optimization layer further restricts the choice to gradient-trained predictive models, excluding strong off-the-shelf learners such as tree-based ensembles~\citep{breiman2001random,friedman2001greedy,grinsztajn2022tree}, distributional models such as NGBoost~\citep{duan2020ngboost}, and tabular foundation models such as TabPFN~\citep{hollmann2025tabpfn}. \citet{elmachtoubEstimateThenOptimizeIntegratedEstimationOptimizationSample2023} show that, for problems with known constraints, estimating a well-specified model first and optimizing afterwards asymptotically dominates integrated estimation and optimization, which pays off mainly when the model class is misspecified and the decision loss compensates for a poor fit. Under constraint \textit{uncertainty}, however, a well-fitted model alone is not sufficient, since the decision also depends on which statistic of the prediction enters the optimization problem~\citep{cameron2022perils, schutteSufficientDecisionProxies2026}.

Conditional quantiles offer such a statistic. Predicting the conditional $\tau$-quantile of an uncertain right-hand side and using it in a deterministic model is well known in chance-constrained programming (CCP)  to obtain a tractable reformulation of the corresponding chance constraint at level $1-\tau$ \citep{CharnesCooper1959, Charnes1963, Prekopa1995}, so that in this case $\tau$ is exactly the tolerated violation probability. We discuss this connection  and where it stops being exact in Section~\ref{qslo::chance_constraint_view}. Rather than predicting expected parameter values, for which a large share of instances can remain infeasible, predicting a given quantile of the uncertain parameters allows the decision-maker to continuously shift between conservative predictions that favor feasibility and aggressive predictions that favor objective performance~\citep{Steinwart2011Estimating}. The quantile level thus plays an analogous role to ODECE's infeasibility aversion coefficient, but within a simpler and computationally lighter framework that requires no differentiable surrogates, no repeated optimization during training, and no problem-specific penalty functions. Because prediction and optimization are fully decoupled, any model capable of producing quantile estimates or predictive distributions can be used .

This motivates our work, whose contributions we summarize as follows:
\begin{enumerate}
\item We propose Quantile-based Sequential Learning and Optimization (QSLO), a predict-then-optimize approach for constraint uncertainty that uses the quantile level of the predicted constraint parameters as a single control parameter to navigate the feasibility vs.\ (post-hoc) regret trade-off.

\item We demonstrate that QSLO significantly outperforms ODECE on the feasibility vs.\ (post-hoc) regret Pareto frontier across the two synthetic benchmark problems, the multi-dimensional knapsack problem with uncertain weights and uncertain capacities \citep{elmachtoubSmartPredictThen2022, mandiFeasibilityAwareDecisionFocusedLearning2025}, and performs comparably on the benchmark based on real data, the brass alloy production problem \citep{huTwoStagePredict+OptimizeMixed2023, mandiFeasibilityAwareDecisionFocusedLearning2025}, while training up to almost three orders of magnitude faster when the inner optimization is expensive.

\item We find that the decisive factor is quantile control paired with an adequate model fit, not a decision-focused loss, since QSLO already outperforms ODECE with a linear predictor of the same model class, challenging the assumption that decision-focused training is necessary in constrained settings.

\end{enumerate}

The remainder of this paper is structured as follows. Section~\ref{SQLO::literature_review} reviews the literature on contextual stochastic optimization, decision-focused learning, and quantile regression in robust and stochastic optimization. In Section~\ref{QSLO::methods}, we formalize the CSO problem, show how predicting a conditional quantile of an uncertain constraint parameter corresponds to imposing a chance constraint, and describe the proposed QSLO methodology together with its QSLO-P and QSLO-D variants. The predictive models, experimental setup, and computational results follow in Section~\ref{QSLO::experiments}, covering the selection of the regression model, the feasibility vs.\ (post-hoc) regret comparison with ODECE, the training and tuning times, and an analysis of how the predictions and the optimal operating point respond to the correction cost. We conclude in Section~\ref{QSLO::conclusion} with a discussion of the method's limitations and open questions.

\section{Literature Review}\label{SQLO::literature_review}
Interest in combining machine learning and combinatorial optimization for problems under uncertainty has grown in recent years, and several distinct methodological frameworks for contextual stochastic optimization have been proposed. The surveys by \citet{sadanaSurveyContextualOptimization2025} and \citet{mandiDecisionFocusedLearningFoundations2024} provide systematic overviews of approaches that connect predictive modeling with optimization, and both touch on the constraint-uncertain setting we consider here. 
A detailed review of these frameworks is beyond the scope of this paper, and we refer the reader to these surveys for a comprehensive treatment. Both surveys find that ILO methods tend to outperform SLO in decision quality, but at the cost of more complex training procedures. Because the decision loss depends on solving a discrete, non-smooth optimization problems, ILO methods require differentiable surrogates or perturbation-based approximations and repeated optimization during training, which adds significant computational overhead.

\subsection{Sequential and Integrated Learning-Optimization Frameworks}

SLO treats prediction and decision-making as separate stages, where a model is trained to minimize a standard predictive loss, and its output is then passed to the optimization problem \citep{mandiDecisionFocusedLearningFoundations2024}. Since prediction error and decision error are not necessarily aligned, better forecasts do not guarantee better downstream decisions, particularly when decisions are sensitive to tails, constraints, or rare events \citep{mandiDecisionFocusedLearningFoundations2024, elmachtoubEstimateThenOptimizeIntegratedEstimationOptimizationSample2023}. This misalignment is especially problematic when the predictions parameterize the constraints, where small errors can cause infeasibility or high regret \citep{kotaryEndtoEndConstrainedOptimization2021}. The a priori choice of which statistic to predict can have a large impact on decision quality, and predicting the expectation can lead to arbitrarily poor decisions compared with end-to-end training~\citep{cameron2022perils}. In contrast, ILO methods address this by optimizing (a surrogate of) the decision loss during training, but the resulting loss depends on solving a discrete, non-smooth optimization problem, leading to non-differentiability, scalability issues, and reliance on problem-specific relaxations \citep{kotaryEndtoEndConstrainedOptimization2021, tangPyEPOPyTorchbasedEndend2024}. Empirical studies confirm that ILO methods have mainly been demonstrated on relatively simple or convex problems and remain computationally demanding for large-scale combinatorial instances \citep{gengBenchmarkingPtOPnO, tangPyEPOPyTorchbasedEndend2024}.

Neither paradigm is uniformly superior. \citet{elmachtoubEstimateThenOptimizeIntegratedEstimationOptimizationSample2023} prove that when the predictive model is well specified and sufficient data is provided, SLO-style predict-then-optimize can asymptotically stochastically dominate integrated prediction and optimization in terms of regret, while the ordering reverses under model misspecification in settings where only the objective is uncertain and the constraints are known. Empirical studies in cost-sensitive prediction similarly find that decision quality often depends more on how costs and thresholds are incorporated into the decision stage than on using a decision-focused loss during training, and that simple SLO pipelines can be competitive if the downstream policy is well designed~\citep{vanderschuerenPredictthenoptimizePredictandoptimizeEmpirical2022}. 
At the same time, broader benchmarks and tools such as PyEPO indicate that integrated methods frequently outperform SLO on many tasks, but suffer from scalability and robustness issues~\citep{gengBenchmarkingPtOPnO,tangPyEPOPyTorchbasedEndend2024}. Notably, these comparisons typically employ the same linear and shallow neural network architectures for both paradigms, leaving open whether ILO's advantage persists when SLO is paired with strong off-the-shelf learners such as tree ensembles or tabular foundation models, which are incompatible with gradient-based end-to-end training. Taken together, the evidence points to a nuanced trade-off rather than clear dominance of one paradigm over the other.

\subsection{ILO for Uncertainty in the Objective}

The majority of ILO research addresses uncertainty in objective coefficients and a wide range of methods has been proposed. Approaches can broadly be divided into approximate methods, which replace the true decision loss with a smooth surrogate, and exact methods, which differentiate through relaxed or structured optimization procedures \citep{huBranchLearnRecursively2022}. On the approximate side, the SPO+ loss \citep{elmachtoubSmartPredictThen2022} has become a canonical benchmark by providing a convex, decision-aware surrogate for polyhedral problems. Perturbation-based methods such as the Perturbed Fenchel--Young Loss (PFYL) \citep{berthetLearningDifferentiablePerturbed2020} avoid requiring ground-truth optimal solutions during training. Black-box differentiation \citep{vlastelicaDifferentiationBlackboxCombinatorial2020} enables end-to-end training through unmodified combinatorial solvers. Further work has extended these ideas towards robust regret objectives \citep{schutteRobustLossesDecisionFocused2024}, task-aware loss learning \citep{shahDecisionFocusedLearningDifferentiable}, and solver-free approaches that exploit polytope geometry \citep{berdenSolverFreeDecisionFocusedLearning2025}. On the exact side, methods differentiate through KKT conditions \citep{wilderMeldingDataDecisionsPipeline2018}, interior-point barriers \citep{mandiInteriorPointSolving2020}, dynamic programming structure \citep{demiroviDynamicProgrammingPredictOptimise2020}, or recursive algorithmic decompositions \citep{huBranchLearnRecursively2022}. PyEPO provides a unified benchmarking framework for many of these methods \citep{tangPyEPOPyTorchbasedEndend2024}. 

A distinct line of work bypasses parameter prediction and maps contextual features directly to decisions, as in prescriptive analytics and weighted sample-average-approximation policies \citep{li2025ridehailing}. Such maps do not extend to constraint uncertainty, where defining the feasible region and computing post-hoc correction costs both require the actual values of the uncertain constraint parameters rather than a decision recommendation.

Despite their variety, these approaches remain computationally demanding, largely restricted to uncertainty in the objective coefficients, and tailored to specific problems. This motivates the work on uncertainty in the constraints.

\subsection{ILO for Uncertainty in the Constraints}\label{QSLO::litreview::il_ouncertain_constraints}

Decision-focused learning under constraint uncertainty poses challenges beyond those arising from objective uncertainty. When predictions parametrize constraints, even small errors can render solutions infeasible, and the cost of correcting such decisions directly affects downstream performance. Standard ILO methods designed for objective uncertainty, such as SPO+ \citep{elmachtoubSmartPredictThen2022} or PFYL \citep{berthetLearningDifferentiablePerturbed2020}, cannot be straightforwardly applied in this setting, as their loss functions are designed to measure suboptimality of feasible solutions and do not account for infeasibility or its correction costs. Adapting these losses to handle constraint uncertainty in a principled way is non-trivial, motivating the development of dedicated frameworks such as those proposed by \citet{huBranchLearnPosthoc2023, huPredict+OptimizePackingCovering2023, hu2023twostage,huTwoStagePredict+OptimizeMixed2023}.

\citet{hu2023twostage} introduce a canonical two-stage ILO framework for mixed-integer linear programs later extending it to a multi-stage framework \citep{huTwoStagePredict+OptimizeMixed2023} in which predictions are first used to generate an initial solution that is subsequently repaired through problem-specific, post-hoc correction and penalty functions once the true parameters are revealed. This framework generalizes earlier approaches developed for packing and covering linear programs and consistently outperforms classical regression-based pipelines. However, the required correction mechanisms are domain-specific and can lead to unstable learning dynamics or high regret under model misspecification \citep{huBranchLearnPosthoc2023,huPredict+OptimizePackingCovering2023}. \citet{huBranchLearnPosthoc2023} further show that the resulting post-hoc regret depends heavily on the choice of correction and penalty functions, often yielding piecewise non-linear loss landscapes that complicate gradient-based optimization.

\citet{mandiFeasibilityAwareDecisionFocusedLearning2025} propose ODECE, a feasibility-aware DFL framework that explicitly balances feasibility and optimality via a tunable trade-off parameter improving upon the baselines of \citet{hu2023twostage}. ODECE optimizes a weighted combination of the Infeasibility Penalty Loss (IPL) and the Optimality-Preserving Loss (OPL), with IPL penalizing predictions that lead to infeasible solutions, while OPL penalizes predictions that would cause the true optimal solution to become infeasible. The user-defined infeasibility aversion coefficient $\alpha$ balances these objectives, making the approach applicable to generic constrained problems without assuming LP or ILP structure.

ODECE integrates feasibility awareness into the training loss through an infeasibility aversion coefficient, whereas \citet{huBranchLearnPosthoc2023} train on the post-hoc regret, which includes correction costs. \citet{imSmartSurrogateLosses2025} instead handle constraint uncertainty in the optimization model. They build contextual uncertainty sets for the uncertain constraint parameters and solve a robust problem, which they describe as a safe approximation of the corresponding contextual chance-constrained problem, while the objective cost vector is learned end-to-end. Other ILO approaches handle uncertain constraints without problem-specific surrogates. \citet{garciaApplicationDrivenLearningClosedLoop2024} use bilevel KKT-based optimization, combining exact and heuristic methods, and show that learning point forecasts that are deliberately biased toward lower operating and reliability costs can outperform purely accuracy-driven prediction. \citet{silvestriScoreFunctionGradient2024} develop score-function gradient estimators with stochastic smoothing that apply to constraint uncertainty without requiring problem-specific structure. Their method typically needs more training epochs but performs on par with specialized methods and particularly well under constraint uncertainty. QSLO, much like \citet{garciaApplicationDrivenLearningClosedLoop2024}, builds this bias into the prediction itself and, like \citet{imSmartSurrogateLosses2025}, has a chance-constraint interpretation, which is exact for a single uncertain right-hand side (Section~\ref{qslo::chance_constraint_view}), but it requires neither a decision loss nor solving the optimization problem during training.

The a priori choice of which statistic to predict can lead to arbitrarily poor decisions compared with end-to-end training~\citep{cameron2022perils}, and a single point prediction may not capture all information relevant for the decision. \citet{schutteSufficientDecisionProxies2026} therefore study when a single predicted scenario, i.e., a deterministic proxy, is sufficient and when a distribution is needed. They propose alternative proxies and show their effectiveness in experiments with uncertainty in both the objective and the constraints. Other decision-focused approaches predict full distributions and solve a stochastic problem over samples from them~\citep{zhaoDiffusionDFLDecisionfocusedDiffusion2025}. \citet{homem-de-melloForecastingOutsideBox2026} show that two-stage stochastic programs with a fixed recourse matrix, fixed costs, and a linear second stage can be solved with a single ``optimal scenario'', which need not be unique and may lie outside the support of the distribution. They further show that decision-focused learning via a bilevel formulation yields pointwise forecasts that asymptotically best approximate this optimal scenario within the chosen class of forecast functions. A single well-chosen point, not necessarily the mean, can therefore be decision-optimal. QSLO follows the same idea by plugging a single conditional quantile instead of the mean into the deterministic model. With a distributional model, QSLO could in addition feed a stochastic version of the problem at evaluation time without retraining, which we leave for future work.

\subsection{Fine-Tuning SLO Using ILO}
Hybrid approaches combine the robustness of SLO pretraining with the decision alignment of ILO through a subsequent fine-tuning phase. \citet{beichterDecisionFocusedRetrainingForecast2024} propose \emph{Decision-Focused Retraining}, which first trains predictive models using standard base losses and then fine-tunes them with surrogate decision losses, using strong SLO initialization to mitigate the prediction-decision misalignment. \citet{yangDFFDecisionFocusedFinetuning2025} introduce \emph{Decision-Focused Fine-tuning} (DFF), which incorporates bias-controlled trust-region constraints to preserve predictive performance during ILO-based optimization, enabling stable decision-focused learning even in limited-data regimes and settings with constraint uncertainty. \citet{sunPredictthenCalibrateNewPerspective} take a different approach and decouple prediction and uncertainty quantification entirely by first training a base model without concern for the downstream risk profile, and subsequently calibrating its uncertainty in post-processing.

\subsection{Conditional Quantiles}
Conditional quantiles have increasingly been used for incorporating asymmetric cost structures and robustness considerations into predict-then-optimize pipelines. \citet{wangPredictThenOptimize2022} use quantile regression at a grid of quantile levels to build an equally weighted scenario set that approximates the conditional distribution, and then solve a sample-average problem for objectives that are nonlinear in the uncertain parameter. The approach is proposed without numerical experiments, with applications in multimodal transportation in mind. Closest to our setting, \citet{mahmutogullariLeveragingContextualInformation2023} predict demand quantiles from contextual information for a capacitated vehicle routing problem with time windows, so that the uncertainty enters the capacity constraints. They interpret quantiles as an implicit robustness mechanism and show that appropriately chosen quantile levels in a deterministic model reach robustness levels comparable to those of robust optimization models. QSLO builds on this idea of a single quantile as a plug-in for uncertain constraint parameters, but treats the quantile level as a control parameter that traverses the feasibility vs.\ (post-hoc) regret trade-off for general constrained problems.

Related approaches exploit quantiles for decision-making and decision-aware forecasting. Two of them embed quantiles in two-stage problems under demand uncertainty. \citet{romer2023predict} propose a predict, tune and optimize framework for shift scheduling, in which quantile levels are jointly optimized alongside other model parameters to maximize expected decision quality. \citet{ScenarioPredictthenOptimizeDataDriven} instead use predicted demand quantiles to generate scenarios in a scenario predict-then-optimize framework for data-driven inventory routing, which avoids strong distributional assumptions. Other works use quantiles as forecasts without a downstream optimization model: \citet{brandoUsingQuantileRegression2022} use high quantiles of multicore contention as safe, conservative estimates in real-time systems, and \citet{wangCostSensitiveOperationalManagement2026} use an asymmetric quantile loss in demand forecasting to account for the differing costs of under- and overestimation, which proves robust across both statistical and business cost metrics.
In contrast to both groups, QSLO plugs a single conditional quantile of the uncertain constraint parameters into the deterministic model and uses the quantile level to navigate the trade-off between feasibility and post-hoc regret.

\section{Method}\label{QSLO::methods}
This section formalizes the methodological foundations of our approach. We first introduce the CSO problem and the reformulation that reduces it to a standard inequality form, so that constraint uncertainty, whether on the left- or right-hand side, enters the optimization through a single predicted quantity $\hat{\mathbf{y}}$.

Afterwards, we define the key evaluation criteria, the regret and post-hoc regret, that guide our experimental analysis. We recall that predicting the conditional quantile of an uncertain right-hand side reproduces the corresponding chance constraint exactly, which gives the quantile level a precise probabilistic meaning,  and we delimit how far this equivalence extends once several uncertain parameters or several constraints are involved. Finally, we provide the necessary background on quantile regression and describe how it can be used to traverse the feasibility vs. (post-hoc) regret trade-off, motivating the two variants of our proposed method QSLO.

\subsection{Problem Formulation}\label{df_dfl::problem_formualtion}
Following \citet{mandiDecisionFocusedLearningFoundations2024}, we consider contextual stochastic optimization (CSO) problems formulated as parametric constrained optimization problems where uncertain parameters $\mathbf{y}$ appear in the objective and constraints:

\begin{subequations}\label{eq:cso_problem}
\begin{align}
    \mathbf{z}^\star(\mathbf{y}) = \underset{\mathbf{z}}{\arg\min} \quad &f(\mathbf{z}, \mathbf{y}) \label{eq:objective}\\
    \text{s.t. } \quad &g_i(\mathbf{z}, \mathbf{y}) \leq 0 \quad \forall \ i \in \mathcal{E} \label{eq:inequality_constraints}\\
    &h_i(\mathbf{z}, \mathbf{y}) = 0 \quad  \forall \ i \in \mathcal{I}.\label{eq:equality_constraints}
\end{align}
\end{subequations}

The goal is to find an optimal solution $\mathbf{z}^\star(\mathbf{y}) \in \mathbb{R}^n$ that minimizes the objective function $f$ subject to the constraints $g_i$ and $h_i$. The set of constraints $\mathcal{C}=\mathcal{E} \cup \mathcal{I}$ entails the set of equality constraints $\mathcal{E}$ and inequality constraints $\mathcal{I}$. To predict the uncertain parameters $\mathbf{y} \in \mathbb{R}^p$, we use historical data comprising contextual observed characteristics $\mathbf{x} \in \mathbb{R}^d$, represented in the form $D=\{(\mathbf{x}_i,\mathbf{y}_i)\}_{i=1}^N$, with $N$ denoting the number of instances \citep{sadanaSurveyContextualOptimization2025, mandiDecisionFocusedLearningFoundations2024}.

Following \cite{mandiFeasibilityAwareDecisionFocusedLearning2025}, equality constraints can be handled by two inequality constraints $h_i(\mathbf{z}, \mathbf{y}) \leq 0$ and $-h_i(\mathbf{z}, \mathbf{y}) \leq 0$. Additionally, right-hand side (RHS) uncertainties $b_i(\mathbf{z}, \mathbf{y})$ in a constraint $i$ of the form $g_i(\mathbf{z}, \mathbf{y}) \leq b_i(\mathbf{z}, \mathbf{y})$ can be incorporated into the standard form by subtracting them from the left-hand side (LHS), yielding $g_i(\mathbf{z}, \mathbf{y}) - b_i(\mathbf{z}, \mathbf{y}) \leq 0$. This reformulation allows us to handle both left-hand side and right-hand side parameter uncertainties in a unified framework without loss of generality:
\begin{subequations}\label{eq:cso_problem_reformed}
\begin{align}
    \mathbf{z}^\star(\mathbf{y}) = \underset{\mathbf{z}}{\arg\min} \quad &f(\mathbf{z}, \mathbf{y}) \label{qslo::eq:objective_short}\\
\text{s.t. } \quad &g_i(\mathbf{z}, \mathbf{y})-b_i(\mathbf{z}, \mathbf{y}) \leq 0 \quad \forall \ i \in \mathcal{C} \label{qslo::eq:reformed_inequality_constraints}
\end{align}
\end{subequations}

\begin{figure}[t!]
    \centering
    \includegraphics[width=.8\linewidth]{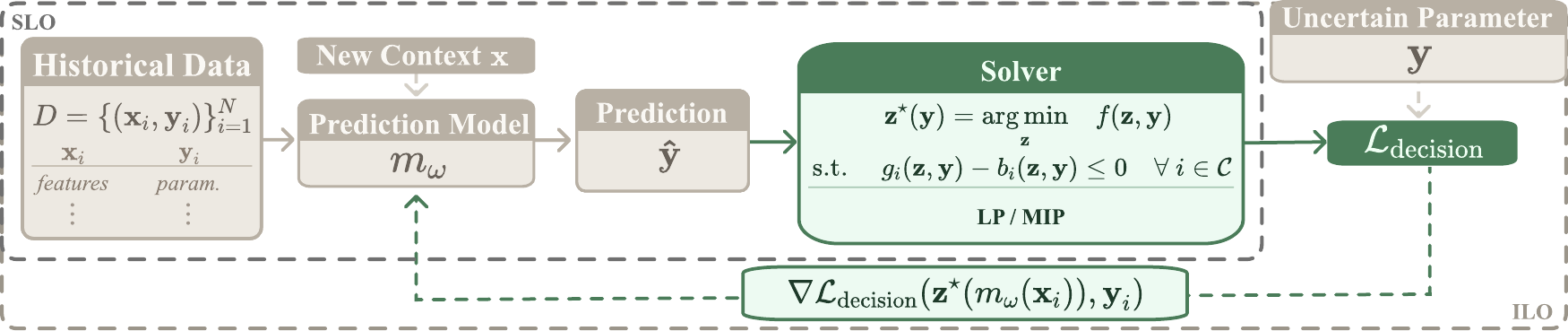}
        \caption{
        The inner block (SLO) trains the predictive model $m_{\omega}$ on a standard prediction loss $\ell$ (e.g.\ MSE) and passes its prediction $\hat{\mathbf{y}}$ to the solver, which returns the decision $\mathbf{z}^\star(\hat{\mathbf{y}})$. ILO adds the outer loop, where the decision is evaluated by a decision loss $\mathcal{L}_{\text{decision}}$ and the optimization problem is fed back into training, requiring $\mathbf{z}^\star(\cdot)$ to be solved repeatedly and differentiated through \citep{sadanaSurveyContextualOptimization2025}. } 
    \label{qslo::fig:SLO_ILO_overview}
\end{figure}

To solve the optimization problem in Equations~\eqref{qslo::eq:objective_short} and~\eqref{qslo::eq:reformed_inequality_constraints}, the predicted parameters $\hat{\mathbf{y}} = m_{\omega}(\mathbf{x})$ are used to obtain a prescriptive decision $\mathbf{z}^\star(\hat{\mathbf{y}})$. As illustrated in Figure~\ref{qslo::fig:SLO_ILO_overview}, SLO and ILO differ in \emph{where} the optimization problem enters the learning pipeline. In SLO approaches, prediction and optimization are decoupled. A predictive model $m_{\omega}: \mathbb{R}^d \rightarrow \mathbb{R}^p$ with parameters $\omega$ is trained by minimizing a prediction loss $\min_{\omega} \sum_{i=1}^N \ell(\mathbf{y}_i, m_{\omega}(\mathbf{x}_i))$ on the contextual data $D$, where $\ell$ is a loss function such as the mean squared error (MSE). The resulting predictions are only passed to the optimizer afterwards. ILO approaches embed the optimization problem into training, directly optimizing the downstream decision quality with $\min_{\omega} \sum_{i=1}^N \mathcal{L}_{\text{decision}}(\mathbf{z}^\star(m_{\omega}(\mathbf{x}_i)), \mathbf{y}_i)$, where $\mathcal{L}_{\text{decision}}$ evaluates the quality of the decision $\mathbf{z}^\star(m_{\omega}(\mathbf{x}_i))$ in comparison to the true parameters $\mathbf{y}_i$. This requires solving $\mathbf{z}^\star(\cdot)$ repeatedly during training and differentiating through it.

Depending on the problem setting, different decision losses $\mathcal{L}_{\text{decision}}$ are used. When uncertain parameters appear only in the objective function, the standard regret, defined in Eq.~\eqref{df_dfl::eq::regret} (also called the true $\mathit{SPO}_{\mathit{loss}}$) is most commonly used \citep{tangPyEPOPyTorchbasedEndend2024, mandiDecisionFocusedLearningFoundations2024, sadanaSurveyContextualOptimization2025, elmachtoubSmartPredictThen2022}. Since regret functions are often non-differentiable, surrogate losses such as SPO+ and perturbations like PFYL are employed in practice. The normalized regret defined in Eq.~\eqref{df_dfl::eq::normregret} provides a scale-invariant measure also used by~\citet{mandiFeasibilityAwareDecisionFocusedLearning2025}:
\begin{subequations} \label{df_dfl::eq::regrets}
\begin{align}
\text{Regret}(\mathbf{z}^\star(\mathbf{\hat{y}}), \mathbf{y})&=f(\mathbf{z}^\star(\mathbf{\hat{y}}), \mathbf{y})-f(\mathbf{z}^\star(\mathbf{y}), \mathbf{y}) \label{df_dfl::eq::regret} \\ 
\text{Norm Regret}(\mathbf{z}^\star(\mathbf{\hat{y}}), \mathbf{y})&=\frac{1}{N}\sum_{i=1}^N \frac{\text{Regret}(\mathbf{z}^\star(\mathbf{\hat{y}}_i), \mathbf{y}_i)}{f(\mathbf{z}^\star(\mathbf{y}_i), \mathbf{y}_i)} .\label{df_dfl::eq::normregret}
\end{align}
\end{subequations}

As discussed, when uncertain parameters appear in constraints, predicted parameters may lead to infeasible solutions. That is, the constraint set $\{\mathbf{z} : g_i(\mathbf{z}, \mathbf{y})-b_i(\mathbf{z}, \mathbf{y}) \leq 0 \quad \forall \ i \in \mathcal{C}\}$ may be empty or may not contain any solution that would be optimal under the true parameters $\mathbf{y}$. Following \cite{ huBranchLearnPosthoc2023,huPredict+OptimizePackingCovering2023, hu2023twostage, huTwoStagePredict+OptimizeMixed2023}, we denote $\mathbf{z}^*_{corr}(\hat{\mathbf{y}}, \mathbf{y})$ as the corrected solution that restores feasibility under true parameters $\mathbf{y}$. The correction may involve a penalty $\text{Pen}(\mathbf{z}^*(\hat{\mathbf{y}}) \to \mathbf{z}^*_{corr}(\hat{\mathbf{y}}, \mathbf{y}))$ representing the cost of feasibility restoration. This motivates the post-hoc regret measure defined in Eq.~\eqref{df_dfl::eq::posthocregret}, which accounts for both the suboptimality of the corrected solution and the correction penalty. As with standard regret, we use the normalized post-hoc regret given in Eq.~\eqref{df_dfl::eq::normposthocregret} as an empirical surrogate for tractability:
\begin{subequations} \label{df_dfl::eq::posthocregrets}
\begin{align}
&\text{Post-hoc Regret}(\mathbf{z}^\star(\mathbf{\hat{y}}), \mathbf{y}) = f(\mathbf{z}^*_{corr}(\hat{\mathbf{y}}, \mathbf{y}), \mathbf{y}) - f(\mathbf{z}^*(\mathbf{y}), \mathbf{y}) + \text{Pen}(\mathbf{z}^*(\hat{\mathbf{y}}) \to \mathbf{z}^*_{corr}(\hat{\mathbf{y}}, \mathbf{y})) \label{df_dfl::eq::posthocregret} \\ 
&\text{Norm Post-hoc Regret}(\mathbf{z}^\star(\mathbf{\hat{y}}), \mathbf{y})=\frac{1}{N}\sum_{i=1}^N \frac{\text{Post-hoc Regret}(\mathbf{z}^\star(\mathbf{\hat{y}}_i), \mathbf{y}_i)}{f(\mathbf{z}^\star(\mathbf{y}_i), \mathbf{y}_i)} \label{df_dfl::eq::normposthocregret}.
\end{align}
\end{subequations}

Finally, we define the infeasibility rate as 
\begin{align}
    \text{Infeasibility Rate}(\mathbf{z}^\star(\hat{\mathbf{y}}), \mathbf{y}) = \frac{1}{N}\sum_{i=1}^{N} \text{Infeasible}(\mathbf{z}^\star(\hat{\mathbf{y}}_i), \mathbf{y}_i), \label{df_dfl::eq::infeasibilityrate}
\end{align}
where $\text{Infeasible}(\mathbf{z}^\star(\hat{\mathbf{y}}), \mathbf{y})$ is an indicator function that equals $1$ if $\mathbf{z}^\star(\hat{\mathbf{y}})$ violates at least one constraint under the true parameters $\mathbf{y}$, and $0$ otherwise.

\subsection{Uncertainty in the Constraints: A Chance-Constraint Programming Perspective}\label{qslo::chance_constraint_view}
When the uncertainty affects the constraints rather than the objective, the two prediction errors of overestimation and underestimation are no longer symmetric. Depending on the problem, one of them can render the committed decision infeasible, while the opposite error leaves it feasible, but overly conservative. A point forecast of the conditional mean ignores this asymmetry. It weights both errors equally and gives the decision-maker no way to bias predictions toward feasibility or toward objective performance \citep{wangCostSensitiveOperationalManagement2026, wangQuantileRegressionAnalyzing2012, dressResidualValueForecasting2018}. Distributional forecasts avoid this by providing access to the full conditional distribution of the uncertain parameters, so that this trade-off can be made explicit rather than fixed implicitly by a mean prediction \citep{fernandesUncertaintyDisplaysUsing2018, paulusCombOptNetFitRight2022}. Quantile regression provides a principled foundation for extracting such distributional information \citep{koenkerQuantileRegression402017}, and the resulting conditional quantile is a statistic of the predictive distribution that can enter the optimization model~\citep{cameron2022perils}.

 Replacing uncertain parameters by quantiles is common practice in CCP~\citep{CharnesCooper1959, Charnes1963}. A well-known result in CCP is that, when the RHS of a single constraint is affected by uncertainty, replacing it with a quantile yields an exact and tractable reformulation of a chance constraint. Such chance constraints arise, for example, in inventory service levels and power system capacities~\citep{Prekopa1995}. We formalize how the quantile level $\tau$ of the predicted uncertain RHS relates to the tolerated violation probability of the corresponding chance constraint, which gives $\tau$ a precise probabilistic meaning, and explain why this equivalence no longer holds once several uncertain parameters or several constraints are involved.

Formally, consider such a constraint $i\in\C$, in the form $g_i(\z,\yv)\le b_i(\z,\yv)$ introduced in Section~\ref{df_dfl::problem_formualtion}, whose right-hand side does not depend on the decision, $b_i(\z,\yv)=b_i(\yv)$, and whose activity $g_i(\z)$ is deterministic. 
A decision $\z$ is feasible for constraint $i$ if and only if $g_i(\z)\le b_i(\yv)$. Let $F$ denote the conditional cumulative distribution function (CDF) of the RHS $b_i(\yv)$ given the context $\x$, that is, $F(t)=\Prob\bigl(b_i(\yv)\le t \mid \x\bigr)$. We assume throughout that $F$ is continuous and strictly increasing, so that its inverse $F^{-1}(\tau)$ is the conditional $\tau$-quantile.

\begin{proposition}[{\citealp{Charnes1963}}]\label{qslo:prop:rhs}
For any tolerated violation probability $\tau\in(0,1)$,
\[
  \Prob\bigl(g_i(\z)\le b_i(\yv)\bigr)\;\ge\;1-\tau
  \qquad\Longleftrightarrow\qquad
  g_i(\z)\;\le\;F^{-1}(\tau).
\]
Predicting the RHS at its conditional $\tau$-quantile, $\bhat_i=F^{-1}(\tau)$, and solving the deterministic surrogate $g_i(\z)\le\bhat_i$ reproduces the individual chance-constrained feasible set exactly, and $\bhat_i=F^{-1}(\tau)$ is the least conservative prediction that guarantees the desired probability of feasibility $1-\tau$.
\end{proposition}
This is the deterministic equivalent of an individual chance constraint with a random right-hand side under a zero-order decision rule, i.e., a decision fixed before the right-hand side is observed~\citep{Charnes1963}, written here for the conditional distribution of $b_i(\yv)$ given the context and for a general activity $g_i(\z)$, which is deterministic once $\z$ is fixed. A prediction below $F^{-1}(\tau)$ only shrinks the feasible region, while a prediction above it admits decisions that violate the reliability target, so $F^{-1}(\tau)$ is the least conservative choice.

\textbf{Beyond the exact case.}
The quantile-based reformulation is exact because the feasibility of constraint $i$ depends on $\yv$ only through the single value $b_i(\yv)$. We discuss three cases in which this assumption does not hold.
(i)~In the exact case, feasibility depends on $b_i(\yv)$ monotonically by construction, since $b_i$ enters as an additive threshold. If the uncertainty instead enters the activity $g_i(\z,\yv)$ through a single parameter that likewise leaves the activity monotone in it, the picture is unchanged in spirit. The prediction still moves the decision up to, and across, the feasibility threshold, so the same quantile reasoning applies.
(ii)~If several uncertain coefficients enter one constraint, the errors interact. Overestimation and underestimation offset within the decision-weighted aggregate, so only the \emph{aggregate}'s quantile controls feasibility, and that aggregate depends both on the joint law of $\yv$ and on $\z$ itself. 
(iii)~With several constraints carrying uncertainty, say $k$ of them, as in the MDKP-C benchmark, a joint guarantee requires the cross-constraint dependence structure. A Bonferroni choice of per-constraint level $\tau/k$ remains valid but conservative \citep{NemirovskiShapiro2006}. \citet{imSmartSurrogateLosses2025} similarly obtain a safe, conservative approximation of a contextual chance-constrained problem through robust constraints. A closed-form quantile level optimal for the full feasibility versus (post-hoc) regret trade-off would therefore require assumptions on the cross-constraint dependence structure that we do not impose, which is why $\tau$ is treated as an operating point selected on validation data.

Based on the above-mentioned results and observations, we put quantiles at the center of our approach for CSO with  uncertain constraint parameters. Specifically, we propose a Quantile-based Sequential Learning and Optimization (QSLO) approach that predicts a conditional quantile of the uncertain parameters and uses those as input to the optimizer. In practice, QSLO can be implemented in two ways: training a quantile regression for a certain $\tau$ (QSLO-P) or reading a quantile $\tau$ from a learned distribution (QSLO-D). The choice between these two depends on the available models and the desired flexibility in quantile selection. When referring to a specific quantile level $\tau$ of a concrete model, we abbreviate QSLO-D/QSLO-P \textit{model}($\tau$) to \textit{model}($\tau$) for readability.

\subsection{Quantile-based Sequential Learning and Optimization}\label{qslo::methods::qslo}

\begin{figure}[ht!]
    \centering    \includegraphics[width=.8\linewidth]{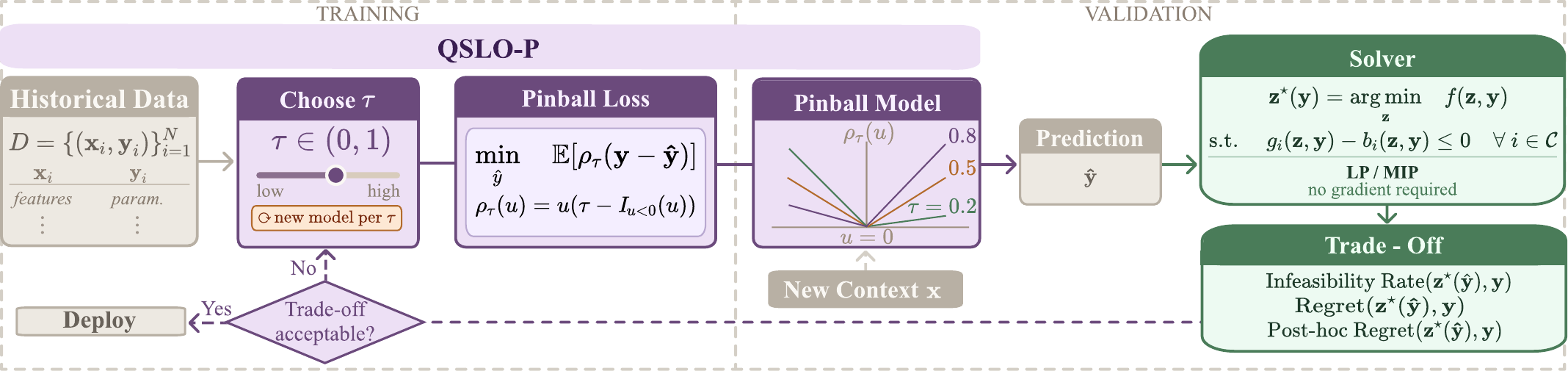}
        \caption{High-level overview of the QSLO-P approach that minimizes the pinball loss $\rho_{\tau}$ for a fixed $\tau$ chosen before training. If a decision maker is not satisfied with the trade-off between infeasibility and (post-hoc) regret, the training as well as the testing has to be repeated for a new quantile $\tau$.}
    \label{qslo::fig:qslop-overview}
\end{figure}

QSLO-P, as shown in Figure~\ref{qslo::fig:qslop-overview}, trains a quantile regression model by minimizing the expected pinball loss in Eq.~\eqref{df_dfl::eq::quantile_loss} for a target quantile level $\tau \in (0,1)$ \citep{Steinwart2011Estimating, jadonComprehensiveSurveyRegression2022}, rather than minimizing the mean squared error (MSE) or mean absolute error (MAE):
\begin{subequations} \label{df_dfl::eq::quantile_loss}
\begin{align}
\min_{\hat{y}} \quad &\mathbb{E}[\rho_{\tau}(\mathbf{y}-\mathbf{\hat{y}})]\\
&\rho_{\tau}(u)= u(\tau - \mathit{I}_{u<0}(u)),
\end{align}
\end{subequations}
where $\mathit{I}_{u<0}$ indicates whether the residual $u = \mathbf{y}-\hat{\mathbf{y}}$ is negative. The resulting estimator corresponds to the conditional $\tau$-quantile of $\mathbf{y}$ given $\mathbf{x}$. When a fixed quantile is trained directly via the pinball loss, the level $\tau$ must be chosen before training, and exploring different operating points requires separate models. 

\begin{figure}[ht!]
    \centering
\includegraphics[width=.8\linewidth]{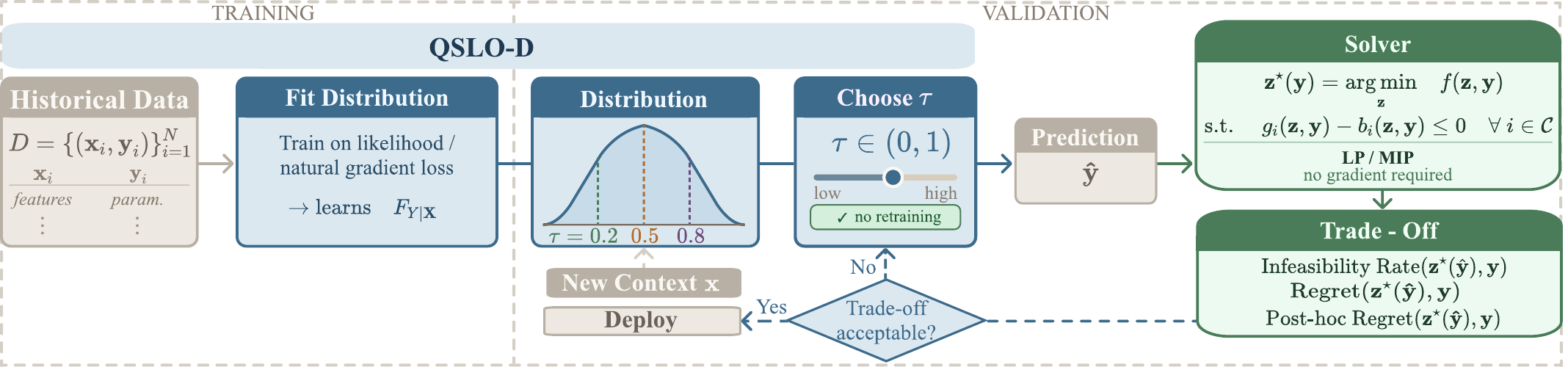}
    \caption{High-level overview of the QSLO-D approach that fits the full conditional
distribution $F_{Y|\mathbf{X}}$. If a decision maker is not satisfied with the trade-off between infeasibility and (post-hoc) regret, only the testing part has to be repeated for a new quantile $\tau$.}
    \label{qslo::fig:qslod-overview}
\end{figure}

In contrast, QSLO-D, as shown in Figure~\ref{qslo::fig:qslod-overview}, fits a model representing the full conditional distribution $F_{Y|\mathbf{X}}$, from which any quantile can be extracted at test time as $\hat{\mathbf{y}}_\tau = F^{-1}_{Y|\mathbf{X}}(\tau)$ without retraining. \citep{koenkerQuantileRegression402017, Steinwart2011Estimating}. When the uncertain parameter is the right-hand side itself, this is exactly the conditional quantile of Proposition~\ref{qslo:prop:rhs}.

In many practical decision-making settings, some degree of constraint violation is acceptable, if it yields substantially improved objective values, and this trade-off  between infeasibility and regret should therefore be explicitly modeled rather than implicitly enforced during training \citep{Kononova2022The, Diz2018A, Meselhi2022Modified, Jatty2023Diagnosis, mandiFeasibilityAwareDecisionFocusedLearning2025}. Our quantile-based approach leverages this property directly. By adjusting $\tau$, the decision-maker can continuously shift between conservative predictions that favor feasibility and optimistic predictions that favor objective performance. 

A second advantage of QSLO is its simplicity. Regardless of whether the quantile is obtained via QSLO-P or QSLO-D, the predicted parameters $\hat{\mathbf{y}}$  are passed directly to any standard solver, whether exact, heuristic, or metaheuristic, without requiring end-to-end differentiation through the optimization problem. Both variants are therefore lightweight and agnostic to the structure of the downstream optimization problem. Decoupling prediction from optimization in this way offers an advantage in terms of modularity. Any change to the objective function, the regret definition, or the post-hoc regret calculation requires only re-running the evaluation, not retraining the model. If new constraints or objective components are added, only the corresponding predictive models need to be updated. In contrast, ILO models must be retrained whenever the problem changes. 

\begin{figure}[ht!]
    \centering
\includegraphics[width=.8\linewidth]{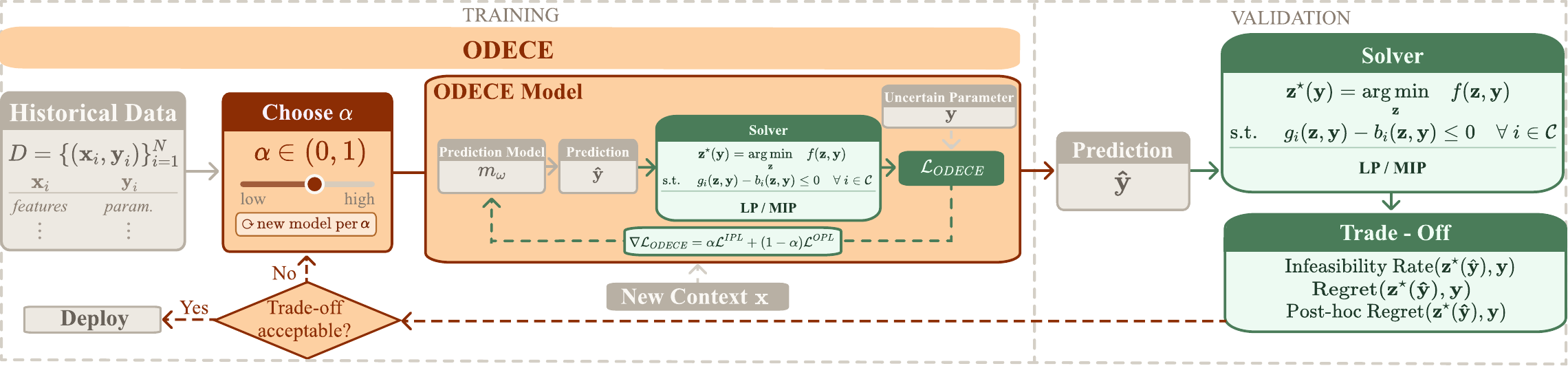}
        \caption{High-level overview ODECE, which optimizes a weighted combination of the Infeasibility Penalty Loss (IPL) and the Optimality-Preserving Loss (OPL) \citep{mandiFeasibilityAwareDecisionFocusedLearning2025}. If a decision maker is not satisfied with the trade-off between infeasibility and (post-hoc) regret, the training as well as the testing has to be repeated for a new infeasibility aversion coefficient $\alpha$.}
    \label{qslo::fig:odece-overview}
\end{figure}

The quantile level, however, does not reveal the trade-off it will produce. It biases predictions toward feasibility or toward objective performance. With the exception of very simple settings (see Section~\ref{qslo::chance_constraint_view}) it does not tell the decision-maker in advance which infeasibility rate or post-hoc regret will result. The same is true for ODECE \citep{mandiFeasibilityAwareDecisionFocusedLearning2025},  that balances infeasibility and (post-hoc) regret through a tunable infeasibility aversion coefficient $\alpha$ (see Figure~\ref{qslo::fig:odece-overview}). Both parameters must therefore be tuned on validation data whenever a specific infeasibility rate or post-hoc regret is targeted. Regret is a poor tuning target, as it is computed only over feasible instances. For QSLO-P and ODECE, tuning means retraining a model for every inspected $\tau$ or $\alpha$ (Figures~\ref{qslo::fig:qslop-overview} and~\ref{qslo::fig:odece-overview}), but QSLO-D avoids this. Any quantile can be read from a single fitted distribution, so a practitioner can sweep the full frontier and select the operating point that matches their cost structure \emph{a posteriori}, revising it later as repair costs or risk preferences change. Evaluating each inspected $\tau$ or $\alpha$ still requires solving the optimization problem once per validation instance, but, unlike ILO training, no repeated solves or differentiation through the solver.

We stress that solving and evaluating the optimization problem \emph{after} obtaining $\hat{\mathbf{y}}$, whether at decision time or when evaluating operating points during tuning, does not constitute integrating the optimization problem into training in the ILO sense. What distinguishes the two is not whether the optimizer is invoked, but whether its solution feeds back into the model parameters $\omega$. In QSLO it never does. The model is trained purely on the pinball loss in Eq.~\eqref{df_dfl::eq::quantile_loss} or fit as a distribution, and the optimizer is invoked only at decision time, with no backpropagation through the solver and no re-solving during learning.

\section{Experiments} \label{QSLO::experiments}
This section presents the experimental setup and results of our empirical evaluation. The goal is to assess whether quantile-based SLO can match or outperform state-of-the-art ILO methods, in particular with respect to the feasibility vs.  (post-hoc) regret trade-off and computational training cost. We first describe the benchmark problems and datasets, followed by the experimental settings and model selection. We then present and discuss the results, comparing QSLO against two baselines across three benchmark problems, partially replicating the experimental setup of \citet{mandiFeasibilityAwareDecisionFocusedLearning2025} to ensure a fair and direct comparison.

\subsection{Problems and datasets}
\textbf{Multi-dimensional knapsack problem (MDKP).}
We evaluate our approach on a 0–1 MDKP with two prediction settings: (i) forecasting the item weight vectors when the capacity constraints are known (MDKP-W), and (ii) forecasting the capacity vectors when item weights are known (MDKP-C).
For data generation, we follow the synthetic setup introduced by \citet{elmachtoubSmartPredictThen2022} and subsequently adopted by \citet{mandiFeasibilityAwareDecisionFocusedLearning2025}. Each instance consists of 50 items and three knapsack constraints, where each uncertain parameter depends on 10 observable features. For each experimental run, we generate $1{,}500$ instances, partitioned into $900$ training, $100$ validation, and $500$ test instances.
To avoid degenerate optimization problems in which the optimal solution selects either all or none of the items, we apply the same safeguards as \citet{mandiFeasibilityAwareDecisionFocusedLearning2025}, which rescale the capacities in both settings. 
A full, reproducible description of the instance-generation procedure, is given in Appendix~\ref{qslo::app::datagen}.

\textbf{Brass Alloy Production problem (BAPP).}
We evaluate our method on the Brass alloy production problem, replicating the setup from \cite{hu2023twostage}, which is formulated as a covering linear program (LP). We use the publicly available Brass alloy dataset from \cite{mandiFeasibilityAwareDecisionFocusedLearning2025} repository.
The problem involves procuring ore from 10 suppliers to produce Brass alloy, which requires at least 627.54 units of Copper (Cu) and 369.72 units of Zinc (Zn). Each supplier offers ore at a distinct unit price, but the precise Cu and Zn content per unit of ore is uncertain and must be predicted. The objective is to select a cost-minimizing combination of ores that meets these minimum metal requirements. Cu and Zn contents for each supplier's ore are predicted from separate feature sets of 4,096 dimensions each. From the 500 available instances, we allocate 350 for training, 50 for validation, and 100 for testing.

\textbf{Post-Hoc Correction.}
For both problems, feasibility is restored through a uniform scaling correction. In the MDKP, where constraints take the form $\mathbf{w}_d^\top \mathbf{z} \leq \mathbf{c}_d$, with $\mathbf{w}_d$ denoting the weights and $\mathbf{c}_d$ denoting the capacity for dimension $d = 1, \ldots, D$, an infeasible solution $\mathbf{z}$ is scaled down by $\eta = \max\left\{\mathbf{w}_d^\top \mathbf{z} / \mathbf{c}_d \mid d\right\}$, $\eta \geq 1$, yielding the corrected solution $\mathbf{z}_{corr} = \mathbf{z} / \eta$, which effectively reduces the number of selected items. In the BAPP, where constraints take the form $\mathbf{w}_d^\top \mathbf{z} \geq \mathbf{o}_d$, with $\mathbf{w}_d$ denoting the metal content of the ore and $\mathbf{o}_d$ denoting the minimum required amount of metal $d = 1, \ldots, D$, the solution is scaled up by $\eta = \max\left\{\mathbf{o}_d / \mathbf{w}_d^\top \mathbf{z} \mid d \right\}$, $\eta \geq 1$, yielding $\mathbf{z}_{corr} = \eta \cdot \mathbf{z}$, corresponding to procuring additional 
ore. In both cases, the penalty term is defined as the normalized absolute change in objective value induced by the correction:
\begin{equation}
    \text{Pen}(\mathbf{z}^*(\hat{\mathbf{y}}) \to \mathbf{z}^*_{corr}(\hat{\mathbf{y}}, \mathbf{y})) = \lambda \cdot 
    \frac{|f(\mathbf{z}^*(\mathbf{\hat{y}}),\mathbf{y}) - f(\mathbf{z}^*_{corr}(\mathbf{\hat{y}},
\mathbf{y}),
\mathbf{y})|}{f(\mathbf{z}^*(\mathbf{y}),\mathbf{y})}, 
    \quad \lambda \geq 0,
    \label{eq:penalty}
\end{equation}
with the penalty coefficient $\lambda$ being set to one in all experiments unless stated otherwise.

\subsection{Experimental Settings}\label{QSLO::sec::experimentalsettings}
We compare against the baselines ODECE \citep{mandiFeasibilityAwareDecisionFocusedLearning2025} and an MSE-trained model, the latter representing the standard SLO approach. We do not compare against \citet{hu2023twostage} directly, as \citet{mandiFeasibilityAwareDecisionFocusedLearning2025} demonstrate that ODECE dominates and outperforms it in our setting, making it the appropriate state-of-the-art baseline. We run ODECE using their released implementation and published model choices per problem.
On MDKP that is a linear model, which ODECE’s authors chose deliberately to show that DFL works despite misspecification, and a one hidden-layer network for the BAPP. Details are in Table~\ref{qslo::app::odece_hyperparameters_table} inside Appendix~\ref{qslo::app::hyperparameters}. We vary the infeasibility-aversion coefficient $\alpha \in \{0.0, 0.1, \ldots, 0.9, 1.0\}$, and include the extremes $\alpha = 0.0$ and $\alpha = 1.0$ for completeness, but regard them as unrealistic operating points. When $\alpha$ is set to $0.0$, the weight is placed on minimizing regret, completely ignoring feasibility, which usually leads to highly infeasible solutions. Setting $\alpha$ to  $1.0$ focuses only on feasibility at the expense of decision quality, typically generating degenerate solutions with high regret \citep{mandiFeasibilityAwareDecisionFocusedLearning2025}. In addition to these two baselines, we evaluate five regression model families, divided into point forecast models and distributional models. For each family, we also include a version yielding mean predictions, which we refer to as SLO.

\textbf{Point Forecast Models.}
Linear Regression models the target as a weighted linear combination of input features, estimated by minimizing either the MSE or the quantile loss \citep{hastie2009elements, koenkerQuantileRegression402017}. Random Forests aggregate predictions from an ensemble of decorrelated decision trees trained on bootstrapped subsamples, reducing variance through averaging \citep{breiman2001random}. For quantile estimation we use Quantile Regression Forests, which retain the full empirical distribution of training responses in each leaf and estimate quantiles directly \citep{meinshausen2006quantile}. Gradient Boosting sequentially fits shallow trees to the negative gradient of the loss, a procedure that generalizes naturally to the quantile loss for direct conditional quantile estimation \citep{friedman2001greedy}.

\textbf{Distributional Models.}
NGBoost extends gradient boosting to probabilistic prediction by fitting the parameters of a parametric output distribution using natural gradients, yielding a full conditional distribution from which means and quantiles can be extracted \citep{duan2020ngboost}. TabPFN is a Prior‑Data Fitted Network, i.e., a transformer‑based tabular foundation model pre‑trained offline on a large number of synthetically generated datasets to approximate Bayesian posterior predictive inference on tabular data \citep{hollmann2023tabpfn,hollmann2025tabpfn, grinsztajn2025tabpfn}. At inference time, the entire training set is provided as context and predictions for all test inputs are produced in a single forward pass, without any gradient‑based fine‑tuning on the new dataset. Due to its transformer architecture, TabPFN benefits strongly from GPU acceleration, with inference on standard benchmarks being reported to be up to several orders of magnitude faster on the GPU than on the CPU \citep{hollmann2025tabpfn}. Both mean and quantile predictions are extracted directly from the predictive distributions of NGBoost and TabPFN.

\textbf{Hyperparameters. }
We use the standard hyperparameter settings of each model, state every deviation from the library default together with its reason, and perform no hyperparameter tuning, since models such as RF already work well with their default values~\citep{probstHyperparametersTuningStrategies2019}. For comparability, we fix the ensemble size of all three tree-based models to $500$ trees. This is already the default for NGBoost, while for RF and GBM we raise it from the default of $100$. For RF, performance typically stabilizes well before $500$ trees~\citep{probst2018tune}, and additional trees mainly increase computation time without harming accuracy in general. The tree depth, in contrast, differs by design. RF grows full trees, as forests average deep, low-bias trees~\citep{breiman2001random, meinshausen2006quantile}, whereas GBM and NGBoost are boosting methods that combine many shallow trees, so we keep their library default depth of $3$. For LR, the SLO variant uses ordinary least squares. For QSLO-P on the MDKP, we remove the L1 penalty to obtain standard unpenalized quantile regression~\citep{koenkerQuantileRegression402017}, which matches the SLO variant. The BAPP, however, has more features than training instances, so ordinary quantile regression is inconsistent and requires a penalty~\citep{belloniL1penalizedQuantileRegression2011}. We therefore keep the default L1 penalty level for the BAPP, which on this dataset sets every coefficient to zero, so that LR predicts the same $\tau$-quantile for every instance. TabPFN requires no hyperparameter tuning by design, as it performs Bayesian posterior predictive inference at test time~\citep{hollmann2025tabpfn}. We use the \texttt{fit\_preprocessors} mode, in which the training data is preprocessed and cached once during training and then reused at inference instead of being recomputed, which reduces memory overhead during testing. Full hyperparameter specifications are reported in Appendix~\ref{qslo::app::hyperparameters}. Any regression model that supports quantile loss minimization or distributional prediction can be used, which makes our approach agnostic to the specific predictive architecture.

\textbf{Code Adaptations.}
We build on the public implementation of \citet{mandiFeasibilityAwareDecisionFocusedLearning2025}\footnote{\href{https://github.com/JayMan91/OdeceDFLforConstraintsNeurips25}{https://github.com/JayMan91/OdeceDFLforConstraintsNeurips25}} and extend it in three ways. First, during regression model training, we add a small Gaussian noise ($\texttt{noise\_scale} = 10^{-6}$) to any feature columns that are constant across all training instances, as several regression models do not accept constant-valued inputs. Second, we adapt the treatment of unsolvable instances. In settings like the MDKP, where a trivial feasible solution always exists (e.g., selecting no items), predicted constraint parameters such as negative capacities can still induce empty or unrealistic feasible regions. To prevent this while preserving the possibility of genuine unsolvability, we clip predicted capacities, weights, and metal amounts to zero during validation and testing, in line with standard non-negativity constraints. In practice, tighter domain-informed bounds, for instance physical capacity limits, can be applied in addition to zero-clipping to further reduce degenerate solutions. Even with clipping, truly unsolvable instances can still occur when predicted parameters drive the feasible region to be empty, e.g., zero metal content in BAPP. In these cases, we record the instance as unsolvable, set the post-hoc regret to the true optimal value as a penalty, and exclude them from all other reported averages. Third, we correct a misalignment between features and uncertain parameters in the MDKP-C. In the released code, the train and test split shuffles the features and weights, but not the capacities and costs, so that each instance is paired with the capacities of another instance. We split the capacities and costs together with the features and weights. The code and experimental scripts will be made publicly available.

\textbf{Computational Setup.} All models except TabPFN are run on large-memory CPU nodes equipped with $2 \times$ AMD EPYC Milan 7763 processors (2.45\,GHz, up to 3.5\,GHz, 64 cores each) and 1,024\,GiB main memory, using 4 CPU cores and 32\,GB RAM per job. TabPFN was run on GPU nodes equipped with NVIDIA A100 GPUs with NVLink and 40\,GB HBM2 memory, using 4 CPU cores and 32\,GB RAM per job.

\textbf{Reported KPIs. }For each problem class (MDKP with uncertain weights, MDKP with uncertain capacities, BAPP with uncertain ore quantities), we report the mean and standard deviation over five independent runs of the following test-set metrics: infeasibility rate, regret, post-hoc regret,  and training wall and CPU time. The infeasibility rate is defined as the fraction of test instances for which the decision obtained from the predicted parameters violates at least one constraint. Regret is the normalized regret and is only calculated for feasible solutions, measuring the relative loss of using decisions based on predictions instead of the true optimal decision (Eq.~\eqref{df_dfl::eq::normregret}). The post-hoc regret is the normalized post-hoc regret, quantifying the loss after applying a minimal correction procedure to restore feasibility (Eq.~\eqref{df_dfl::eq::normposthocregret}), which reduces to the regret for feasible solutions.

Detailed results for all three problem classes and all settings (quantile $\tau$ and infeasibility aversion coefficient $\alpha$), including the proportion of unsolvable instances where the solver finds no feasible solution under the predicted parameters even after clipping, appear in Appendix Tables \ref{QSLO::app_WeightsTable_1}–\ref{QSLO::app_AlloyTable_2}.

\subsection{Regression Model Selection}
\label{qslo:sec:ablation_reg}

QSLO turns the feasibility versus (post-hoc) regret trade-off into the choice of a
single quantile level $\tau$. Any model that predicts conditional quantiles can be
steered by $\tau$, but the control is only as useful as the quantiles it acts on. We
therefore assess five candidate models 
by the hypervolume (HV), which condenses each model's whole $\tau$-sweep into a single number, so that the Pareto fronts of Section~\ref{qslo::sec::paretorfronts} can be compared directly. The HV is not the training objective of any candidate, so it does not favor a particular model family. As a check on the quantiles themselves, the empirical coverage in Appendix~\ref{qslo::app::coverage} shows that moving $\tau$ shifts the predictions of every model across the whole range, although the predicted quantiles are not calibrated.

\textbf{Hypervolume} treats each operating point as a pair of objectives to be
minimized, and places a fixed reference point $\mathbf{r}$ at the worst corner of the
objective plane, where both are highest. For a set $S$ of operating points, the HV is
the area of the region dominated by at least one point of $S$ and bounded above by
$\mathbf{r}$ \citep{zitzlerMultiobjectiveEvolutionaryAlgorithms1999},
\begin{equation}
\mathrm{HV}(S;\mathbf{r}) = \mathrm{area}\!\left( \bigcup_{\mathbf{s}\in S}[s_1, r_1]\times[s_2, r_2]\right).
\label{eq:hypervolume}
\end{equation}
A frontier closer to the ideal corner dominates more of this region, so a larger HV
is better.

\begin{figure}[b!]
  \centering
  \includegraphics[width=.9\linewidth]{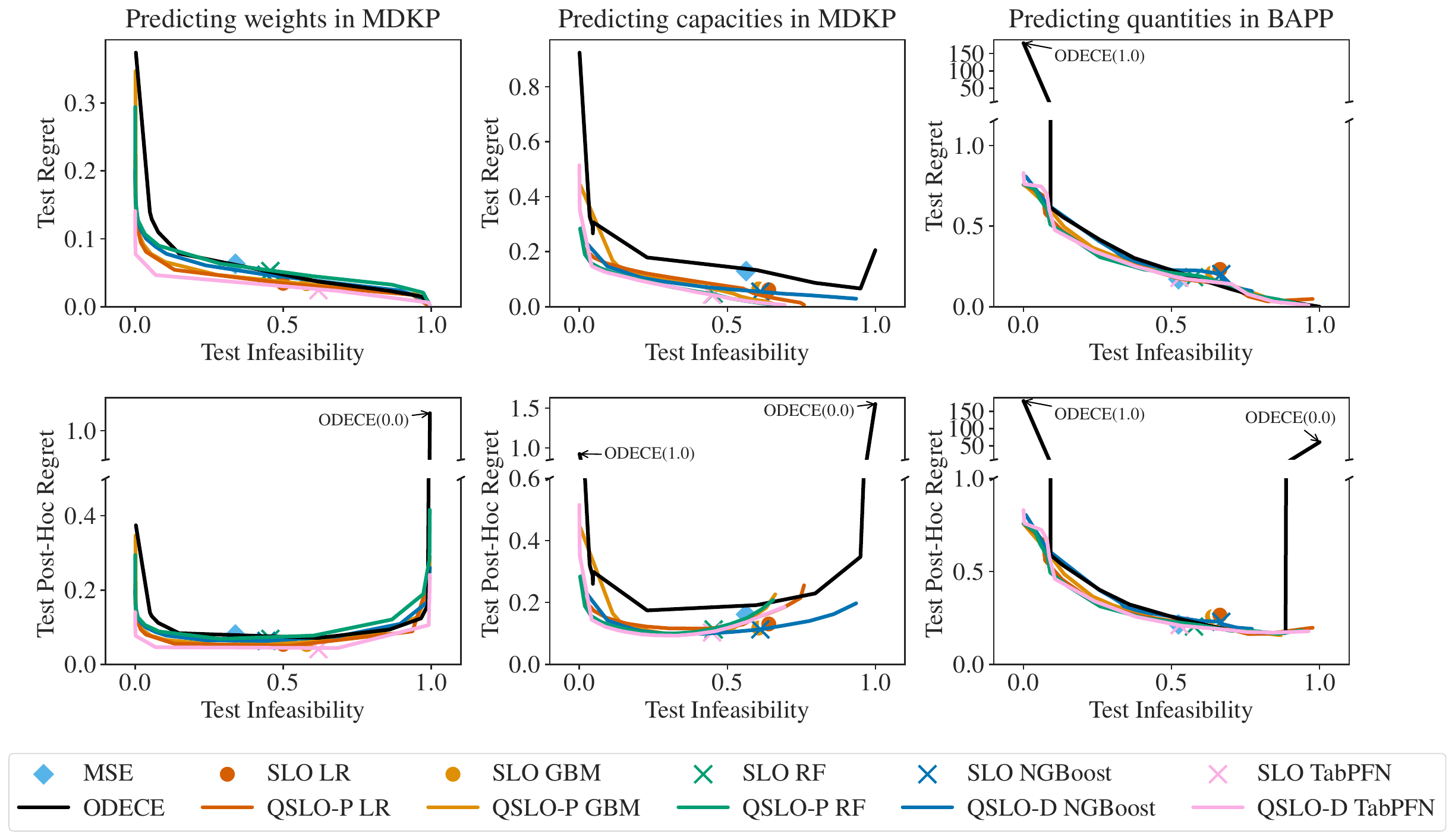}
\caption{Pareto frontiers showing the trade-off between the test infeasibility rate and test regret (top row) and test post-hoc regret (bottom row) for MSE, ODECE, and SLO and QSLO with LR, GBM, RF, NGBoost, and TabPFN across all three problem classes. A method is preferred if it achieves lower infeasibility at a given level of (post-hoc) regret, or lower (post-hoc) regret at a given infeasibility rate. The infeasibility rate accounts for both constraint violations and unsolvable instances. QSLO methods achieve better trade-offs than ODECE across all problems.}
\label{qslo::fig::odece_quantile_slo_ngboost}
\end{figure}

We report the HV for two objective pairs, (infeasibility, regret) and (infeasibility, post-hoc regret), matching the two Pareto fronts in Figure~\ref{qslo::fig::odece_quantile_slo_ngboost} of Section~\ref{qslo::sec::paretorfronts}. Both use the worst-case reference infeasibility $r_1 = 1$, the rate reached when every test instance is infeasible, and set $r_2$ to the worst-case (post-hoc) regret observed for that problem across all models, rounded up. Before setting $r_2$, we exclude ODECE's degenerate operating points at $\alpha=0$ and $\alpha=1$, which are practically unusable (Section~\ref{qslo::sec::paretorfronts}), and TabPFN at $\tau=0.0001$ on the MDKP with uncertain capacities, whose predictions collapse below the range of the true values. Otherwise the worst-case regret of ODECE at $\alpha=1$ would set the BAPP reference to $190$, so that the HV would mostly reflect the shared area far from any usable operating point. These give reference points of $(1.00, 0.35)$, $(1.00, 0.45)$ and $(1.00, 0.85)$ for the regret sweep on MDKP-W, MDKP-C and the BAPP, and $(1.00, 0.42)$, $(1.00, 0.45)$ and $(1.00, 0.85)$ for the post-hoc regret sweep. Because the reference points differ by problem, HV values are comparable within a problem but not across problems. Since ODECE is our baseline, we also report HV relative to it as $\mathrm{HV}/\mathrm{HV}_{\text{ODECE}} - 1$, so that $0$ denotes parity and a positive value means the model dominates more of the objective space than ODECE.

\begin{table}[t]
\centering
\footnotesize %
\caption{Regression-model ablation reporting hypervolume (HV) per problem, with ODECE as
the baseline row. Higher HV is better. The \emph{vs ODECE} columns give the relative gap
HV/HV$_\text{ODECE}-1$ for the corresponding metric. ODECE's degenerate operating points ($\alpha\in\{0,1\}$) and TabPFN at $\tau=0.0001$ on MDKP-C are excluded before the reference points are set, which are given in
Section~\ref{qslo:sec:ablation_reg}. \emph{Runs} counts the runs, out of five, in which the HV of a model's per-run front exceeds that of ODECE at the same reference point.}
\label{tab:regression-ablation}
\begin{tabular}{llrrrrrr}
\toprule
 & & \multicolumn{3}{c}{\textbf{Regret}} & \multicolumn{3}{c}{\textbf{Post-hoc Regret}} \\
\cmidrule(lr){3-5} \cmidrule(lr){6-8}
 & \textbf{Model} & \textbf{HV} & \textbf{vs ODECE} & \textbf{Runs} & \textbf{HV} & \textbf{vs ODECE} & \textbf{Runs} \\
\midrule
\multirow{6}{*}{\rotatebox[origin=c]{90}{\textbf{MDKP-W}}} & ODECE & 0.272 & --- & --- & 0.321 & --- & --- \\
 & LR & 0.302 & +11\% & 5/5 & 0.362 & +13\% & 5/5 \\
 & GBM & 0.302 & +11\% & 5/5 & 0.360 & +12\% & 5/5 \\
 & RF & 0.286 & +5\% & 5/5 & 0.343 & +7\% & 5/5 \\
 & NGBoost & 0.295 & +8\% & 5/5 & 0.350 & +9\% & 5/5 \\
 & TabPFN & \textbf{0.309} & +13\% & 5/5 & \textbf{0.372} & +16\% & 5/5 \\
\midrule
\multirow{6}{*}{\rotatebox[origin=c]{90}{\textbf{MDKP-C}}} & ODECE & 0.273 & --- & --- & 0.249 & --- & --- \\
 & LR & 0.365 & +34\% & 5/5 & 0.319 & +28\% & 5/5 \\
 & GBM & 0.351 & +29\% & 5/5 & 0.307 & +24\% & 5/5 \\
 & RF & \textbf{0.391} & +43\% & 5/5 & 0.339 & +36\% & 5/5 \\
 & NGBoost & 0.360 & +32\% & 5/5 & 0.331 & +33\% & 5/5 \\
 & TabPFN & 0.388 & +42\% & 5/5 & \textbf{0.341} & +37\% & 5/5 \\
\midrule
\multirow{6}{*}{\rotatebox[origin=c]{90}{\textbf{BAPP}}} & ODECE & 0.528 & --- & --- & 0.493 & --- & --- \\
 & LR & \textbf{0.583} & +10\% & 5/5 & 0.541 & +10\% & 5/5 \\
 & GBM & 0.564 & +7\% & 5/5 & 0.530 & +8\% & 5/5 \\
 & RF & 0.578 & +9\% & 5/5 & \textbf{0.544} & +10\% & 5/5 \\
 & NGBoost & 0.524 & $-$1\% & 3/5 & 0.492 & $-$0\% & 3/5 \\
 & TabPFN & 0.576 & +9\% & 5/5 & 0.539 & +9\% & 5/5 \\
\bottomrule
\end{tabular}
\end{table}

Table~\ref{tab:regression-ablation} shows that every regression model improves on ODECE on both MDKP problems and on both sweeps, and that LR, GBM, RF, and TabPFN also improve on it on the BAPP. On the MDKP with uncertain weights, TabPFN attains the highest HV on both sweeps ($+13\%$ and $+16\%$), while RF improves the least ($+5\%$ and $+7\%$). The gains are largest on the MDKP with uncertain capacities, where RF and TabPFN lead with $+43\%$ and $+42\%$ on the regret sweep and $+36\%$ and $+37\%$ on the post-hoc regret sweep, and GBM improves the least ($+29\%$ and $+24\%$). On the BAPP, LR, RF, and TabPFN improve by $+9\%$ to $+10\%$ and GBM by $+7\%$ to $+8\%$, while NGBoost is on par with ODECE ($-1\%$ and $0\%$). No model is uniformly best, since the leading model changes from problem to problem, and none is uniformly worse, so the comparison does not support eliminating any of them.

These gains are not an artifact of averaging over runs. Every regression model attains a higher HV than ODECE in each of the five runs on both MDKP problems, and so do LR, GBM, RF, and TabPFN on the BAPP, while NGBoost does so in three of five runs (column \emph{Runs} in Table~\ref{tab:regression-ablation}). Section~\ref{qslo::sec::stattests} tests the differences at selected operating points.

The ablation also separates two possible sources of QSLO's advantage, the quantile mechanism and the choice of predictor. On the MDKP, ODECE uses a linear predictor (Table~\ref{qslo::app::odece_hyperparameters_table}), the same model class as LR. QSLO with LR still improves the post-hoc regret HV over ODECE by $13\%$ on the weights and $28\%$ on the capacities (Table~\ref{tab:regression-ablation}), and lowers the best post-hoc regret from $0.071$ to $0.050$ and from $0.174$ to $0.116$. With the model class held fixed, the gain therefore comes from the quantile control. The same holds when we compare SLO and QSLO with the same LR, where the best post-hoc regret drops from $0.054$ to $0.050$, from $0.131$ to $0.116$, and from $0.266$ to $0.164$ on the three problems. On the BAPP, LR is effectively context-free (Appendix~\ref{qslo::app::hyperparameters}), yet it matches ODECE with its 512-unit MLP ($0.164$ against $0.169$). The predictor still matters once the context is informative. On the capacities, the learned models reach a best post-hoc regret between $0.093$ and $0.099$, below the $0.116$ of LR.  As every model improves on ODECE and none is uniformly worse, we carry all five forward. A stronger predictor adds to the gain, but the quantile control is what produces it.
This is not a choice ODECE could replicate. ILO's requirement of end-to-end differentiability excludes the tree-based and distributional models used here~\citep{mandiDecisionFocusedLearningFoundations2024, kotaryEndtoEndConstrainedOptimization2021}, one of the limitations of the integrated paradigm that motivates our approach.

\subsection{Pareto front: Infeasibility vs. (Post-Hoc) Regret}\label{qslo::sec::paretorfronts}

Figure~\ref{qslo::fig::odece_quantile_slo_ngboost} shows the trade-off between feasibility and decision quality as Pareto fronts, plotting test infeasibility rate against test regret and post-hoc regret for all methods. Although the post-hoc regret already accounts for infeasibility through the correction cost, plotting it against the infeasibility rate shows where along the trade-off the corrected cost is lowest. For ODECE, varying $\alpha$ traces out a feasibility–suboptimality frontier, as in \cite{mandiDecisionFocusedLearningFoundations2024}. For LR, GBM, RF, NGBoost, and TabPFN, we obtain analogous frontiers by varying the prediction quantile $\tau$. SLO and MSE have no such control parameter and yield a single operating point each, shown as markers, so the decision-maker cannot move them along the trade-off.

The Pareto fronts for LR, GBM, RF, NGBoost, and TabPFN lie below or coincide with ODECE's frontier across all problems, indicating a strictly better or comparable trade-off between infeasibility and (post-hoc) regret. In the MDKP with uncertain weights, TabPFN strictly dominates ODECE, reaching near-zero infeasibility at $\tau=0.7$ with a regret and post-hoc regret of $0.077$, while NGBoost and RF match or exceed ODECE across the frontier, with best post-hoc regrets of $0.064$ and $0.069$ against ODECE's $0.071$. LR and GBM reach best post-hoc regrets of $0.050$ ($\tau=0.5$) and $0.053$ ($\tau=0.6$), below those of NGBoost and RF. In the MDKP with uncertain capacities, all five regression models clearly outperform ODECE, reaching best post-hoc regrets between $0.093$ (TabPFN) and $0.116$ (LR) against ODECE's $0.174$. In the BAPP, the five are comparable to ODECE across the usable frontier, with best post-hoc regrets between $0.158$ (GBM) and $0.192$ (NGBoost) against ODECE's $0.169$, while avoiding the degenerate configurations ODECE produces at its extremes. TabPFN attains an infeasibility rate of $0.532$ with a regret of $0.176$ and post-hoc regret of $0.209$ at $\tau=0.5$, and NGBoost reaches a post-hoc regret of $0.192$ at $\tau=0.9$ (infeasibility $0.772$).

The extreme cases of ODECE reveal degenerate solutions, configurations that satisfy the method's training objective but are practically unusable, either because all solutions are infeasible or because decision quality collapses entirely. This is clearest in the BAPP, where ODECE(0.0) achieves $100\%$ infeasibility and a post-hoc regret of $61.2$, as some instances are entirely unsolvable, and ODECE(1.0) reaches near-zero infeasibility but a mean regret of $180.6$, the result of substantial over-ordering when ore quantities are underestimated.

Across all problems, regret decreases monotonically as the infeasibility rate increases. This is a trivial artifact of computing regret only over feasible solutions, which vanish as infeasibility approaches one. Post-hoc regret behaves differently, forming a U-shaped curve. At low infeasibility rates, conservative predictions incur high regret on feasible solutions, and at high infeasibility rates, correction costs accumulate across many instances. The minimum typically lies at intermediate infeasibility rates, particularly evident in the MDKP with uncertain weights.

\begin{figure}[ht!]
    \centering
    \includegraphics[width=\linewidth]{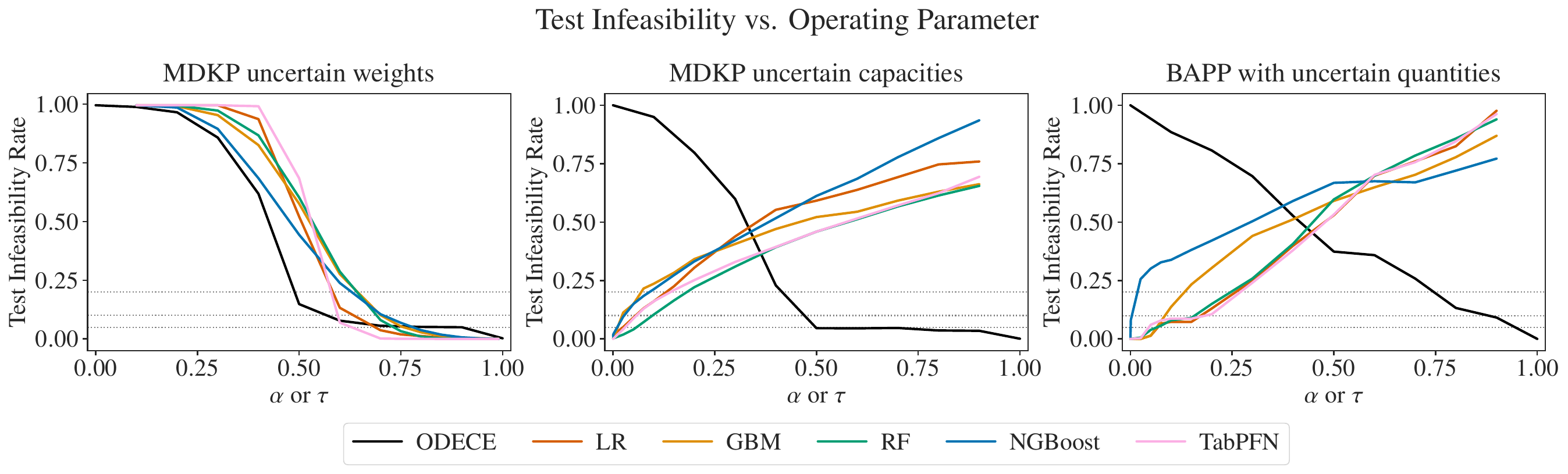}
\caption{Test infeasibility rate as a function of the operating parameter, the quantile level $\tau$ for LR, GBM, RF, NGBoost, and TabPFN and the infeasibility aversion coefficient $\alpha$ for ODECE, per problem. The horizontal lines mark the target infeasibility rates $t\in\{0.05,0.1,0.2\}$ used for tuning in Section~\ref{qslo::sec::trainingtimes}. Validation curves, which are nearly identical, are shown in Section~\ref{qslo::sec::trainingtimes} in Figure~\ref{qslo::fig::val_infeasibility_vs_knob}.}
\label{qslo::fig::infeasibility_vs_knob}
\end{figure}

Both $\tau$ and $\alpha$ are directional control parameters, but each comes with a caveat. 
In QSLO, $\tau$ moves infeasibility monotonically within every problem, though the direction is problem-dependent as shown in Figure \ref{qslo::fig::infeasibility_vs_knob}. A higher quantiles lower infeasibility on the MDKP with  uncertain weights and raise it on the capacities problem and the BAPP, so the decision-maker must know which way to push for a given problem. 
In ODECE, higher $\alpha$ moves infeasibility monotonically toward feasibility on all three problems, but the response is degenerate at the extremes. 
More importantly, neither $\tau$ nor $\alpha$ forecasts the resulting infeasibility rate or (post-hoc) regret. For $\alpha$ this is because it simply weights two loss terms and carries no probabilistic interpretation. For $\tau$ it follows from Section~\ref{qslo::chance_constraint_view}. The exact chance-constraint reading of Proposition~\ref{qslo:prop:rhs} holds for a single constraint with an uncertain RHS. The MDKP with uncertain capacities instead has several such constraints (case~(iii)), and the MDKP with uncertain weights has several uncertain coefficients within each constraint (case~(ii)). Feasibility is then a joint event that depends on how the uncertain parameters interact, so $\tau$ controls the direction of the infeasibility rate but not its value.
The operating point must therefore be chosen empirically by traversing the frontier, which QSLO-D does at test time for free, whereas QSLO-P and ODECE require a full retrain per $\tau$ and $\alpha$.

By decoupling prediction from optimization, QSLO removes the restriction to differentiable architectures that ILO imposes, so any model producing quantiles or a predictive distribution can be used. Across all three problems, all five regression models match or dominate ODECE on the infeasibility versus (post-hoc) regret frontier, with stable behavior even at extreme quantile levels, whereas ODECE degenerates at its own extremes. None of these models uses a decision-focused loss, which runs against the assumption that such a loss is necessary for good decisions in constrained settings. As
the ablation of Section~\ref{qslo:sec:ablation_reg} showed, the gain over ODECE remains when the predictor is held to the same linear model class, and a more expressive predictor adds to it where the context is informative. What matters most is choosing a quantile suited to the problem and pairing it with a model that fits the data, not the alignment of the loss with decision quality.

\subsection{Statistical Tests}\label{qslo::sec::stattests}
We test whether these differences are significant at operating points selected on validation data. For each problem, run, and method, we select the quantile level $\tau$ or the infeasibility aversion coefficient $\alpha$ with the lowest validation post-hoc regret, and compare the test post-hoc regrets of QSLO and the reference method instance by instance with a two-sided paired $t$-test on the mean difference and the Holm correction~\citep{holm1979simple}. Appendix~\ref{qslo::app::stattests} describes the procedure in detail and lists the full results in Table~\ref{tab:stat-tests}.

Table~\ref{tab:stat-tests} shows that on both MDKP problems every regression model attains a significantly lower test post-hoc regret than ODECE, with mean differences of $-0.003$ to $-0.027$ on the uncertain weights and $-0.034$ to $-0.057$ on the uncertain capacities. On the BAPP, no model differs significantly from ODECE at the selected operating point, although LR, GBM, RF, and TabPFN attain a higher HV over the whole frontier (Table~\ref{tab:regression-ablation}). QSLO and ODECE are therefore comparable in decision quality on the BAPP, but TabPFN and LR reach this quality at a lower cost, training one model in $85$\,s and $11$\,s against ODECE's $141$\,s and building the whole frontier for about a third and a tenth of ODECE's cost (Section~\ref{qslo::sec::trainingtimes}). QSLO is also significantly better than SLO with the same regression model on the capacities problem and, except for TabPFN, on the BAPP (Section~\ref{QSLO::sec:ecds_regret_phregret}).

\subsection{Training and Tuning Times}\label{qslo::sec::trainingtimes}

We report three cost figures: the cost of training a single model, the cost of tuning a method to a target infeasibility rate, and the cost of building the full feasibility versus (post-hoc) regret frontier. Table~\ref{tab:tuning-times} covers the first two, and the frontier cost is given in the text.

To keep the comparison with ILO fair, we separate three components: (i) Model training, the cost of fitting a predictive model, (ii) Operating-point selection, choosing $\tau$ or $\alpha$ on the validation set, which applies equally to QSLO and ODECE, (iii) Test-time solving, the per-instance optimization at decision time. 
TabPFN runs on a GPU, so its CPU and GPU times are not directly comparable to the CPU-only methods, and the reported CPU times additionally undercount the models that use multiprocessing.
We therefore compare in (i) the methods on wall time, while still reporting wall, CPU, and GPU time for model training.
In (ii) Operating-point selection and (iii) test-time solving we report wall time only. All values are averaged over five seeds and the tuning targets. For QSLO-D, fitting the distributional model in~(i) already learns the full conditional distribution, so extracting any quantile in~(ii) and~(iii) needs no further training and no re-solving during learning. The MSE baseline trains a point predictor with no tunable operating point, so it cannot trade feasibility against regret and appears only in the training column, dashed elsewhere.

\textbf{Training one model.} In the MDKP settings, training one QSLO model is one to almost three orders of magnitude cheaper than ODECE (Table~\ref{tab:tuning-times}). TabPFN trains in $36$\,s (weights) and $4.6$\,s (capacities) against ODECE's $5{,}303$\,s and $3{,}468$\,s, and NGBoost in $321$\,s and $23$\,s and RF in $348$\,s and $4.3$\,s. LR is the cheapest on the weights at $9.6$\,s and GBM takes $86$\,s, while on the capacities they need $5.4$\,s and $13$\,s. The BAPP reverses this, and the reason is instructive, because what dominates training time differs by method. ODECE's cost is driven by the optimization it solves inside the training loop rather than by the data, so the BAPP's small $20$-variable problem keeps ODECE cheap ($141$\,s) despite its roughly four thousand features. NGBoost and RF instead fit $500$-tree ensembles whose cost grows with the feature count, making them more expensive ($687$ and $698$\,s), and GBM, whose $500$ trees are shallow, takes $241$\,s. TabPFN handles the high dimensionality with far less penalty ($85$\,s), even though the BAPP exceeds its nominal feature limit, which we override to use all features. LR trains in $11$\,s.
Which method is cheapest, therefore depends on the problem. Where the inner solve is expensive, decoupling prediction from optimization lets QSLO win by orders of magnitude, and where the solve is cheap but the feature space is large, TabPFN scales better than the tree ensembles, while ODECEs training time stays low.

\begin{table*}[t]
\centering
\footnotesize
\caption{Tuning runtime and accuracy, in wall-clock seconds, averaged over seeds and the three target infeasibility rates $t\in\{0.05,0.1,0.2\}$. The three column groups, single-model training, the bisection tuning search, and the infeasibility error of the tuned operating point, are defined in the text. MSE has no operating-point tuning, so its tuning and error columns are dashed. Italicized CPU times undercount pooled child-process work (see text).}
\label{tab:tuning-times}
\resizebox{\textwidth}{!}{%
\begin{tabular}{llr@{\hspace{4pt}}r@{\hspace{4pt}}rr@{\hspace{4pt}}r@{\hspace{4pt}}r@{\hspace{4pt}}r@{\hspace{4pt}}rr@{\hspace{4pt}}r@{\hspace{4pt}}r}
\toprule
 & & \multicolumn{3}{c}{\textbf{Train one model}} & \multicolumn{5}{c}{\textbf{Tuning}} & \multicolumn{3}{c}{\textbf{Mean infeasibility error}} \\
\cmidrule(lr){3-5} \cmidrule(lr){6-10} \cmidrule(lr){11-13}
 & \textbf{Model} & \textbf{Wall (s)} & \textbf{CPU (s)} & \textbf{GPU (s)} & \textbf{Evals} & \textbf{Fits} & \textbf{Val (s)} & \textbf{Test (s)} & \textbf{Total (s)} & \textbf{Val} & \textbf{Test} & \textbf{Val$\to$Test} \\
\midrule
\multirow{7}{*}{\rotatebox[origin=c]{90}{\textbf{MDKP-W}}} & MSE & 616.7 & 1,092.0 & --- & --- & --- & --- & --- & --- & --- & --- & --- \\
 & ODECE & 5,302.7 & 20,841.7 & --- & 2 & 2 & \textbf{73.6} & 176.3 & 11,306.7 & 0.029 & \textbf{0.035} & $-$0.017 \\
 & LR & \textbf{9.6} & \textit{0.2} & --- & 3 & 3 & 96.1 & \textbf{166.4} & \textbf{287.4} & 0.046 & 0.048 & \textbf{$-$0.001} \\
 & GBM & 85.6 & \textit{2.2} & --- & 3 & 3 & 101.6 & 189.0 & 523.0 & 0.037 & 0.055 & $-$0.021 \\
 & RF & 347.6 & \textit{346.2} & --- & 3 & 3 & 159.6 & 247.1 & 1,573.1 & 0.046 & 0.058 & $-$0.011 \\
 & NGBoost & 320.9 & \textit{7.7} & --- & 3 & 1 & 289.2 & 452.8 & 1,062.9 & \textbf{0.026} & 0.036 & $-$0.011 \\
 & TabPFN & 35.6 & \textbf{49.5} & 35.6 & 3 & 1 & 337.0 & 540.4 & 913.0 & 0.064 & 0.070 & $-$0.012 \\
\midrule
\multirow{7}{*}{\rotatebox[origin=c]{90}{\textbf{MDKP-C}}} & MSE & 522.0 & 985.3 & --- & --- & --- & --- & --- & --- & --- & --- & --- \\
 & ODECE & 3,467.7 & 13,619.3 & --- & 2 & 2 & \textbf{56.1} & \textbf{107.5} & 8,622.4 & 0.045 & 0.048 & +0.002 \\
 & LR & 5.4 & \textit{0.1} & --- & 3 & 3 & 108.9 & 132.5 & 259.2 & 0.036 & 0.040 & +0.017 \\
 & GBM & 13.4 & \textit{0.2} & --- & 3 & 3 & 101.4 & 113.9 & 261.8 & 0.039 & 0.038 & \textbf{+0.001} \\
 & RF & \textbf{4.3} & \textbf{4.1} & --- & 3 & 3 & 89.1 & 130.5 & \textbf{233.1} & 0.034 & 0.056 & +0.015 \\
 & NGBoost & 23.0 & \textit{0.2} & --- & 3 & 1 & 97.7 & 137.2 & 257.9 & \textbf{0.031} & \textbf{0.037} & +0.009 \\
 & TabPFN & 4.6 & 6.2 & 4.6 & 3 & 1 & 106.9 & 152.3 & 263.8 & 0.038 & 0.047 & +0.015 \\
\midrule
\multirow{7}{*}{\rotatebox[origin=c]{90}{\textbf{BAPP}}} & MSE & 55.2 & \textbf{37.4} & --- & --- & --- & --- & --- & --- & --- & --- & --- \\
 & ODECE & 140.7 & 521.4 & --- & 4 & 4 & 4.3 & 1.6 & 521.4 & 0.082 & \textbf{0.049} & $-$0.113 \\
 & LR & \textbf{10.6} & \textit{2.5} & --- & 4 & 4 & \textbf{0.8} & \textbf{0.2} & \textbf{39.7} & 0.050 & 0.068 & $-$0.045 \\
 & GBM & 240.7 & \textit{2.3} & --- & 4 & 4 & 2.4 & 0.6 & 892.2 & 0.067 & 0.072 & $-$0.039 \\
 & RF & 697.6 & \textit{2.7} & --- & 3 & 3 & 3.0 & 1.7 & 2,097.6 & 0.053 & \textbf{0.049} & $-$0.055 \\
 & NGBoost & 687.4 & \textit{3.5} & --- & 3 & 1 & 26.6 & 18.5 & 732.5 & 0.091 & 0.080 & \textbf{$-$0.019} \\
 & TabPFN & 85.1 & 201.5 & 85.1 & 3 & 1 & 72.1 & 46.0 & 203.3 & \textbf{0.042} & 0.052 & $-$0.061 \\
\bottomrule
\end{tabular}%
}
\end{table*}

\textbf{Tuning to a target infeasibility rate.} We tune to a target infeasibility rate $t\in\{0.05,0.1,0.2\}$ and accept a deviation of $0.05$. We target this metric rather than regret or post-hoc regret, since regret is computed only over feasible instances, while post-hoc regret is problem-specific. When the infeasibility rate is monotone in the operating parameter, the operating point can be found by bisection over the grid rather than by an exhaustive sweep, reducing the number of evaluated levels from $L$ to on the order of $\log_2 L$. We use the grid $\{0.01, 0.1, 0.2, \dots, 0.9, 0.99\}$ of $L=11$ values for both $\tau$ and $\alpha$. Figure~\ref{qslo::fig::infeasibility_vs_knob} shows that infeasibility is monotone in the operating parameter for every method and problem, so bisection is valid for both QSLO and ODECE.  Its infeasibility rates are taken from the full Pareto-front sweep and therefore differ slightly from the tuned operating points in Table~\ref{tab:tuning-times}, which are selected by bisection. The figure establishes monotonicity, while the table measures the wall-clock cost and error of the search.

The \emph{Evals} column is the number of evaluated operating parameters $\tau$ or $\alpha$, and \emph{Fits} the number of models fitted. QSLO-D with NGBoost and TabPFN needs a single fit regardless of how many parameters are evaluated ($\text{Fits}=1$), since any quantile is read from the one distributional model, whereas ODECE and RF refit at every evaluated level. \emph{Val} and \emph{Test} are the total validation time across the search and the final test-set evaluation.
\emph{Total} is the measured end-to-end wall time of the tuning pipeline and exceeds $\emph{Train} + \emph{Val} +\emph{Test}$ by the overhead not attributed to any single stage, such as writing and reloading fitted models between steps.
The mean-infeasibility-error columns report, averaged over the targets and seeds, the mean validation error $= |\text{val infeasibility}-t|$ (\emph{Val}), the mean test error $= |\text{test infeasibility}-t|$ (\emph{Test}), and the signed gap $= \text{test}-\text{val}$ (\emph{Val}$\to$\emph{Test}), where a positive value means the model was more infeasible on test than validation predicted.

\begin{figure}[ht!]
    \centering
    \includegraphics[width=\linewidth]{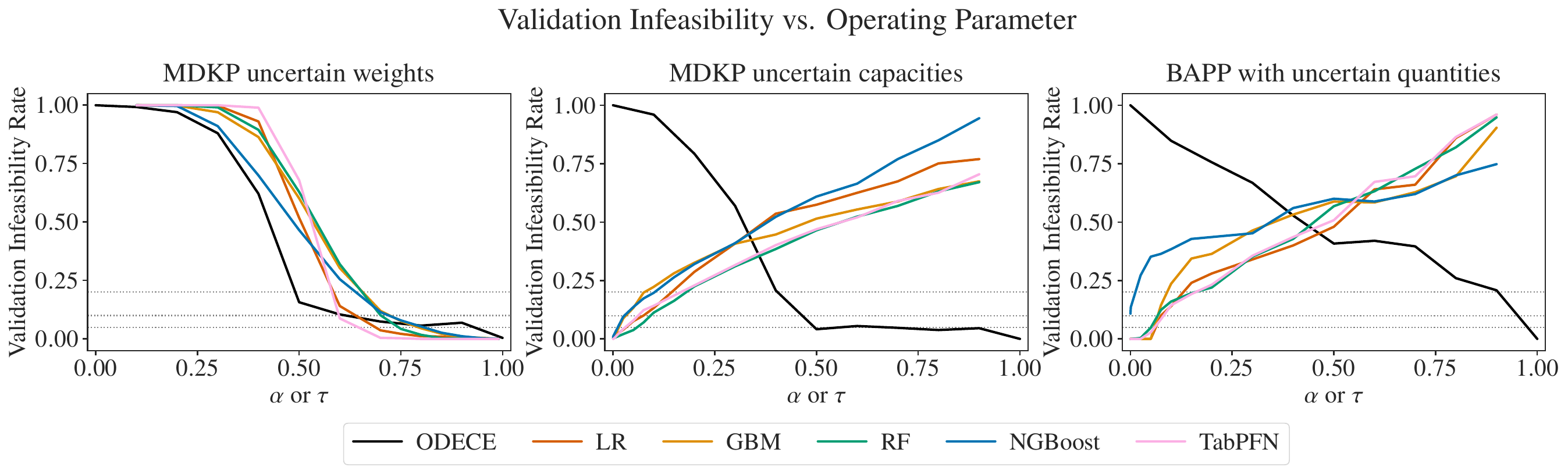}
\caption{Validation infeasibility rate as a function of the operating parameter, $\tau$ for LR, GBM, RF, NGBoost, and TabPFN and $\alpha$ for ODECE, per problem. On the MDKP problems these curves nearly coincide with the test-set curves of Figure~\ref{qslo::fig::infeasibility_vs_knob}, while on the BAPP they diverge, most clearly for ODECE, which is the source of the larger Val$\to$Test tuning gaps in Table~\ref{tab:tuning-times}. The horizontal lines mark the target infeasibility rates $t\in\{0.05,0.1,0.2\}$.}
\label{qslo::fig::val_infeasibility_vs_knob}
\end{figure}

The number of evaluations is comparable across methods (2 to 4), so the tuning cost is driven by the fits, not the search. In the MDKP settings, QSLO tunes far faster than ODECE. Total tuning time is $1{,}063$\,s (weights) and $258$\,s (capacities) for NGBoost and $913$\,s and $264$\,s for TabPFN, against ODECE's $11{,}307$\,s and $8{,}622$\,s, an order of magnitude or more. LR and GBM also stay far below ODECE, with $287$\,s and $523$\,s on the weights and $259$\,s and $262$\,s on the capacities.
On the weights, RF, which refits at every level, is cheaper than ODECE but more expensive than the distributional models ($1{,}573$\,s), illustrating the value of the single fit, since NGBoost and TabPFN pay $\text{Fits}=1$ while ODECE, LR, GBM, and RF pay $\text{Fits}=\text{Evals}$. On the capacities every regressor trains in seconds, so all QSLO models tune in $233$ to $264$\,s.
The BAPP reverses the ranking for the tree ensembles, as in the single-model case, with RF, GBM, and NGBoost the most expensive ($2{,}098$, $892$, and $733$\,s), while TabPFN ($203$\,s) and LR ($40$\,s) stay below ODECE's $521$\,s. LR is this cheap on the BAPP because it is effectively context-free there (Appendix~\ref{qslo::app::hyperparameters}), and as Section~\ref{qslo:sec:ablation_reg} shows, it still matches the learned models, so on this problem the quantile level rather than the predictor carries most of the gain.

The mean infeasibility error columns show that tuning to a target is reliable on the MDKP and looser on the BAPP. On the MDKP problems, every method lands within roughly $0.03$ to $0.07$ of the target on the test set, close to the $0.05$ tolerance. The small signed Val$\to$Test gaps ($|{\cdot}|\le 0.021$) confirm that operating points chosen on validation carry over, with the validation and test infeasibility curves being nearly identical on both MDKP problems (Figures~\ref{qslo::fig::infeasibility_vs_knob} and~\ref{qslo::fig::val_infeasibility_vs_knob}). 
On the BAPP the validation curve tends to lie above the test curve, so the selected operating point, which fits the target on validation, has a lower infeasibility on test. This produces the consistently negative Val$\to$Test gaps, largest for ODECE at $-0.113$, because the wider the gap between the two curves, the larger the resulting error. The test errors nonetheless stay near tolerance ($0.049$ to $0.080$), so tuning to a target remains usable on the BAPP, if less precise than on the synthetic MDKP data.

\textbf{Building the full frontier.} Drawing the complete frontier in Figure~\ref{qslo::fig::odece_quantile_slo_ngboost} sweeps every operating point, $11$ values of $\alpha$ for ODECE and $13$ to $15$ quantiles for the regressors, and has no validation phase. The cost follows a simple pattern. ODECE, LR, GBM, and RF retrain at every level, so their frontier cost is $L$ trainings plus $L$ test evaluations, whereas NGBoost and TabPFN fit once and only the $L$ test evaluations scale, giving one training plus $L$ tests. In the MDKP with uncertain weights this puts ODECE's frontier at about $58{,}000$\,s against roughly $3{,}400$ to $8{,}000$\,s for the QSLO models, and in the MDKP with uncertain capacities about $45{,}000$\,s against $2{,}100$ to $3{,}900$\,s, so QSLO builds the whole frontier for a seventh to a twenty-first of ODECE's cost. The BAPP again reverses for the tree ensembles, where RF and GBM cost about $10{,}500$\,s and $4{,}100$\,s against ODECE's $1{,}900$\,s, while the single fit keeps NGBoost ($990$\,s) and TabPFN ($620$\,s) below ODECE, and LR is cheapest at $180$\,s.

These results show that QSLO matches or exceeds ODECE's decision quality at a fraction of its training and tuning cost in the MDKP settings, where the inner solve is expensive. The distributional models sharpen this further, since any quantile can be read from a single fit. On the BAPP, where the solve is cheap, this cost advantage largely disappears and ODECE is competitive, but TabPFN and LR remain faster to tune than ODECE. The advantage therefore grows with the cost of the inner solve, since that is the factor QSLO removes from the training loop.

\subsection{Empirical CDFs: Effects of Stakeholder Decisions on Prediction Behavior}

\begin{figure}[p!]
    \centering
    \includegraphics[width=.82\linewidth]{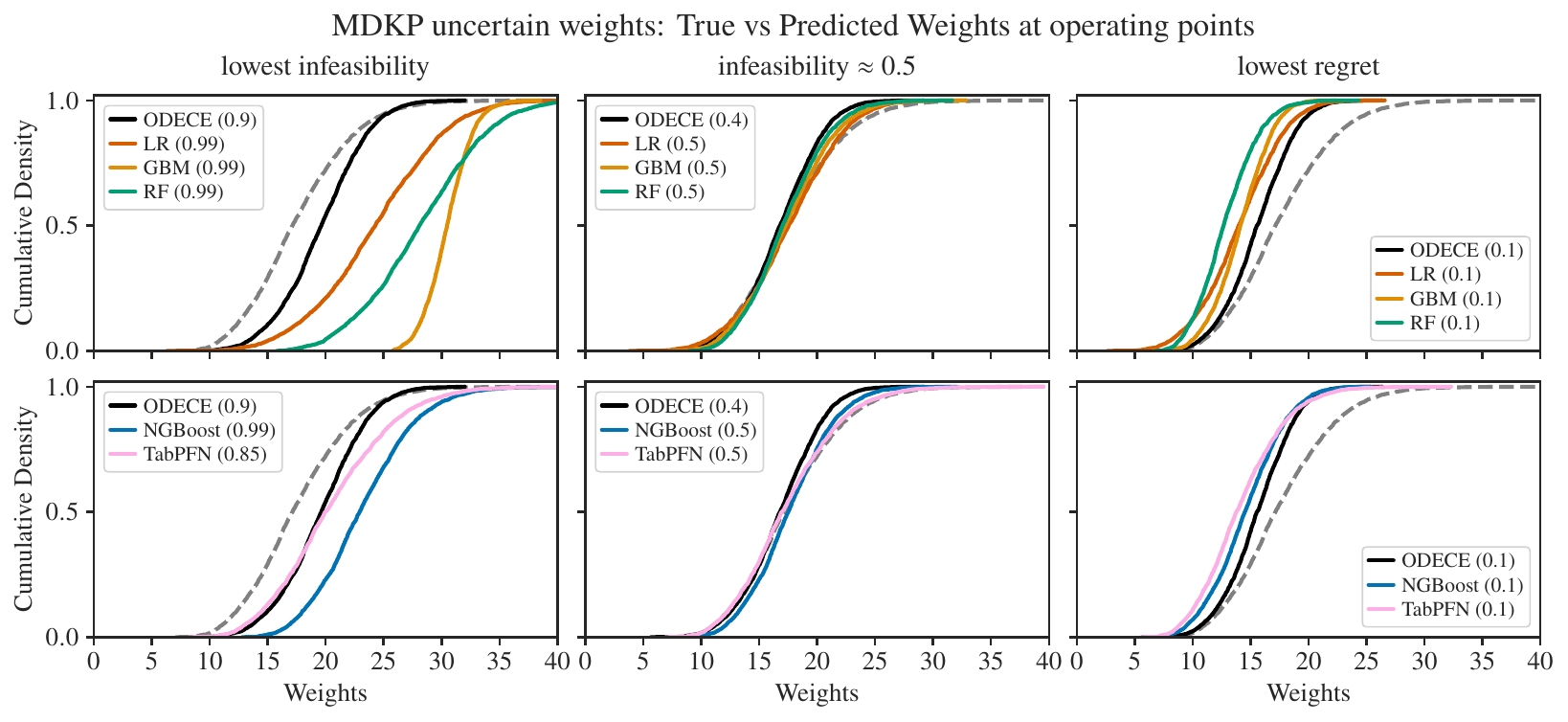}\\
    \vspace{.5cm}
    \centering
    \includegraphics[width=.82\linewidth]{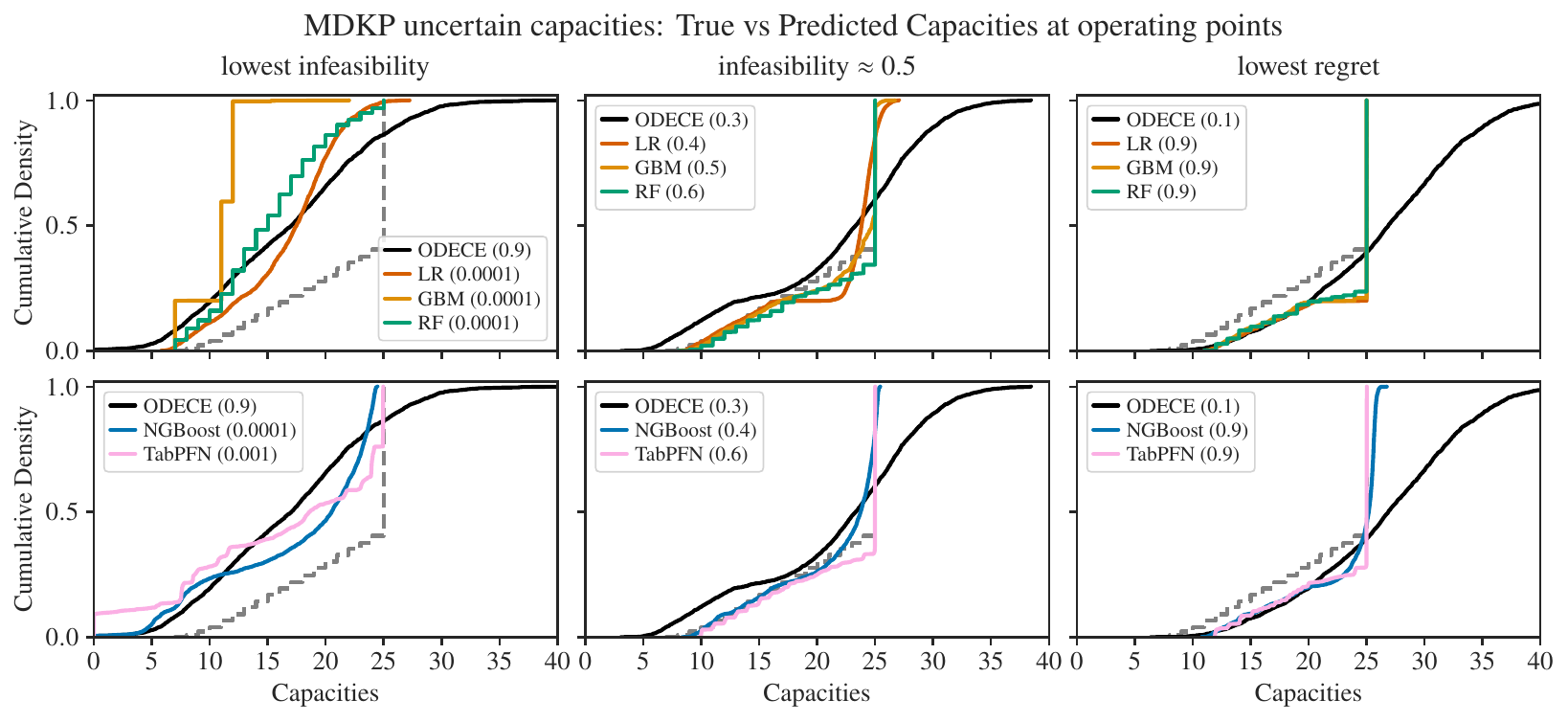}\\
    \vspace{.5cm}
    \centering
    \includegraphics[width=.82\linewidth]{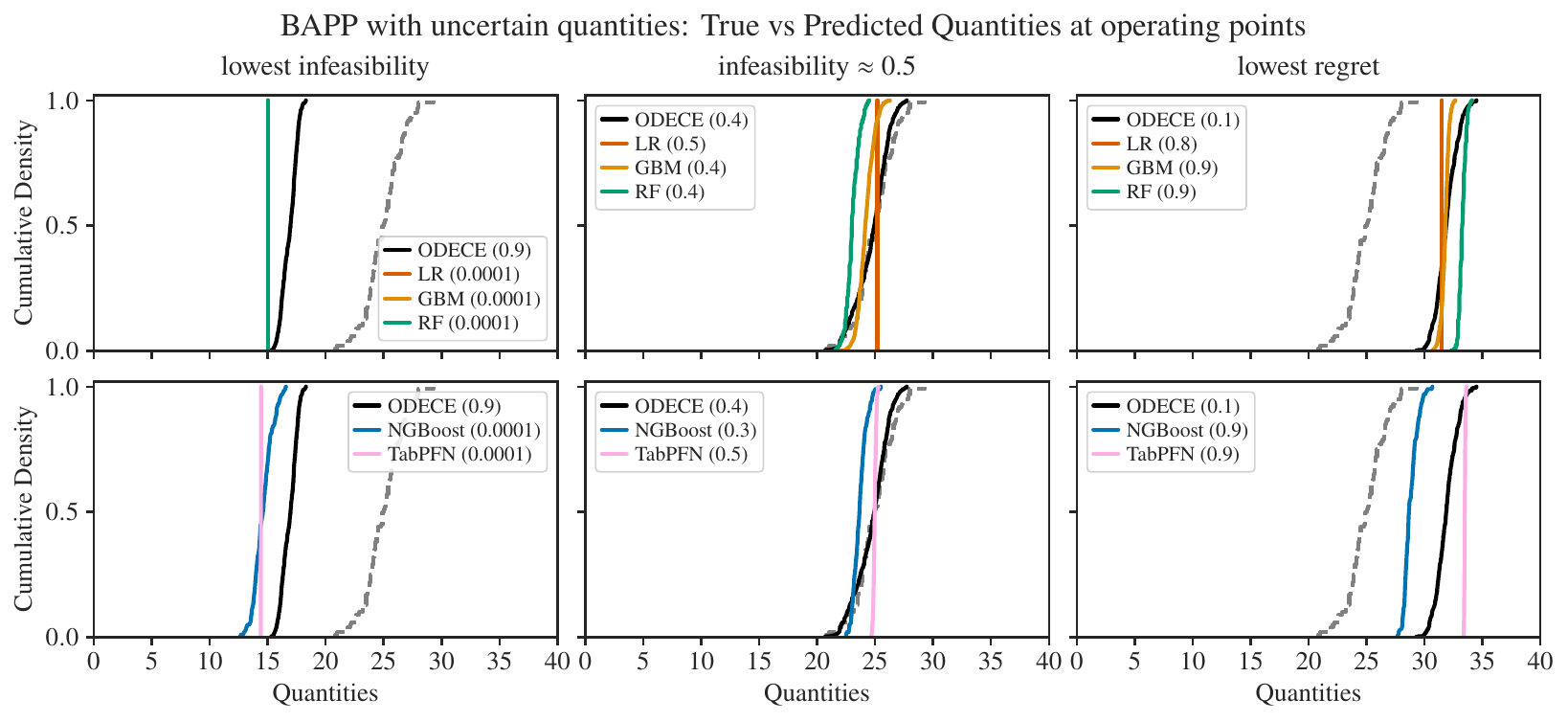} \\
    \vspace{.5cm}
    \caption{ECDFs of the true parameter distribution versus predictions from ODECE and the five regression models at three representative operating points: lowest infeasibility (left), infeasibility $\approx 0.5$ (center), and lowest regret (right). Top: MDKP with uncertain weights. Middle: MDKP with uncertain capacities. Bottom: BAPP with uncertain ore quantities. Within each problem, the upper row shows the QSLO-P models (LR, GBM, RF) and the lower row the QSLO-D models (NGBoost, TabPFN), with ODECE in both rows as the reference. The true distribution is the grey dashed line and is not listed in the legend. The capacity ECDF is less smooth because it has fewer data points ($500$ versus $25{,}000$ for weights and $1{,}000$ for quantities).}
    \label{qslo::fig::ecdf_params}
\end{figure}

Figure~\ref{qslo::fig::ecdf_params} shows the ECDFs of the true uncertain parameters and the predicted distributions of ODECE and the five regression models at three representative operating points per problem. This lets us see how the choice of $\tau$ or $\alpha$ shapes the feasibility vs.\ (post-hoc) regret trade-off. Moving from the lowest-infeasibility operating point to the lowest-regret one shifts the predicted ECDFs across almost the whole range of the true parameter, from one side of the true distribution to the other. The intermediate operating point, near infeasibility $0.5$, lies closest to the true distribution. This is the same traversal seen in the coverage curves of Appendix~\ref{qslo::app::coverage}, now visible in the predicted distributions themselves. Models targeting low infeasibility produce conservative predictions, overestimating weights or underestimating capacities and quantities, while models minimizing regret do the opposite. Intermediate configurations align most closely with the true distribution and achieve competitive post-hoc regret. 

\begin{wrapfigure}{b!}{0.4\columnwidth}
    \centering
    \includegraphics[width=\linewidth]{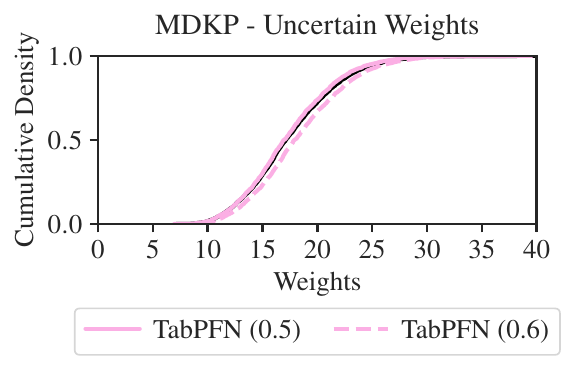}
    \caption{TabPFN at $\tau=0.5$ (underestimate) and $\tau=0.6$ (overestimate) against the true weight distribution (black). Infeasibility falls from $0.68$ to $0.07$.}
    \label{qslo::fig::infeasjump_tabpfn}
    \vspace{-1em}
\end{wrapfigure}

In the MDKP with uncertain weights, TabPFN at $\tau=0.5$ slightly underestimates the true weights, leaving many solutions infeasible, while at $\tau=0.6$ it slightly overestimates them (Figure~\ref{qslo::fig::infeasjump_tabpfn}). This small, deliberate overestimation cuts the infeasibility rate from $0.68$ to $0.07$ while regret rises by only $0.024$, buying most of the feasibility at little cost to decision quality.

A structural difference emerges for the MDKP with uncertain capacities and the BAPP. The regression models implicitly constrain predictions to plausible ranges, while ODECE's predictions are unconstrained and spread well beyond the support of the true distribution. This is visible in the capacity panel where ODECE's ECDF extends far to the right of the others. This is clearest at the degenerate extremes of ODECE in the BAPP, already mentioned in Section~\ref{qslo::sec::paretorfronts}. 
These cases suggest that in constrained settings, respecting the predictive distribution is not merely a statistical nicety but a practical necessity.

A central premise of integrated learning and optimization is that aligning the training objective with downstream decision quality leads to better decisions than optimizing prediction accuracy alone~\citep{elmachtoubSmartPredictThen2022, mandiDecisionFocusedLearningFoundations2024}. Our results challenge this premise in the constrained setting. Across all three problem classes, predicting a suitable quantile of the uncertain parameters matches or outperforms ODECE without integrating the optimization problem into the training process. We attribute this in part to a distributional grounding effect. ODECE's training objective does not keep predictions within plausible ranges, and at its extremes it produces predictions that bear little resemblance to the true parameter distribution. GBM, RF, NGBoost, and TabPFN, by contrast, are implicitly regularized by the data and cannot stray far from the support of the training distribution. 
Unlike in the objective-uncertain setting, where predictions need only preserve the ranking of solutions and the scaling may not matter~\citep{elmachtoubSmartPredictThen2022}, constraint-uncertain problems require predicted parameters to be close to their true values for feasibility to hold.

The operating point that minimizes post-hoc regret depends on how expensive corrections are. We take this up in the next subsection, varying the correction penalty and tracing how the best operating point shifts as repair becomes more costly.

\subsection{Sensitivity of the Optimal Operating Point to the Correction Cost}\label{QSLO::sec:ecds_regret_phregret}

\begin{figure}[ht!]
    \centering
    \includegraphics[width=\linewidth]{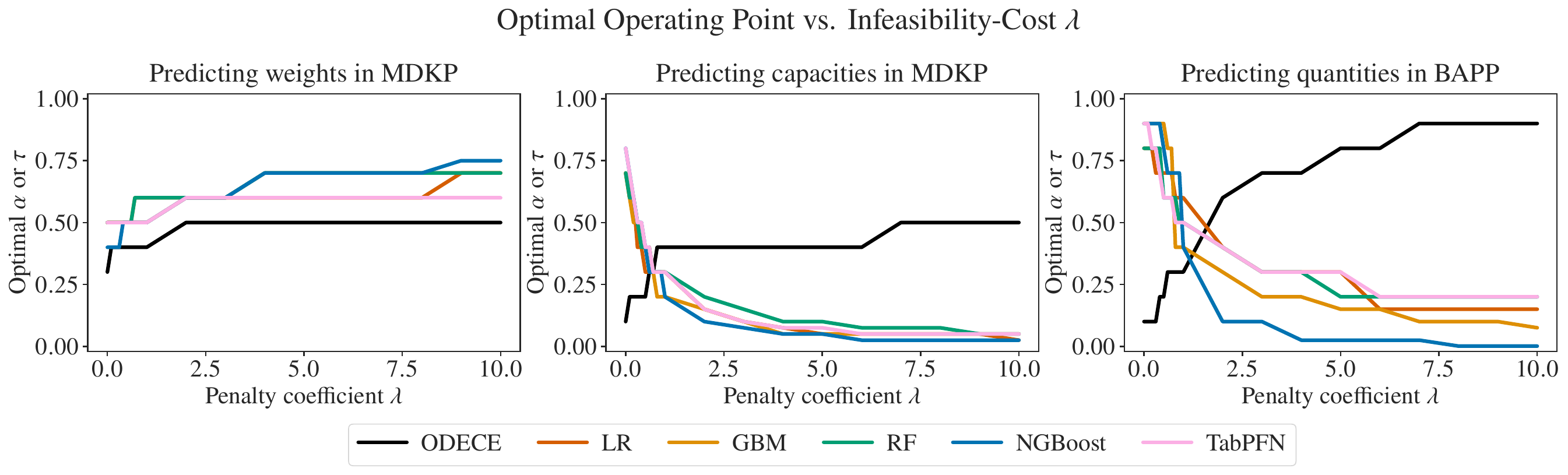}
    \caption{Optimal operating point as a function of the correction penalty $\lambda$, per problem. For each $\lambda$ we recompute the post-hoc regret from the saved test decisions and report the operating point that minimizes it, the quantile $\tau$ for LR, GBM, RF, NGBoost, and TabPFN and the infeasibility aversion coefficient $\alpha$ for ODECE. Left: MDKP with uncertain weights. Center: MDKP with uncertain capacities. Right: BAPP with uncertain ore quantities.}
    \label{qslo::fig::optimal_op_lambda}
\end{figure}

The penalty factor $\lambda$ in Eq.~\eqref{eq:penalty} sets the cost of restoring feasibility, and all preceding results used $\lambda = 1$. Where Section~\ref{qslo::sec::trainingtimes} tuned the operating point to a target infeasibility rate, here we select it to minimize post-hoc regret for a given correction cost, the other criterion a decision-maker might have. Because we store the test decisions across the frontier, the post-hoc regret at any $\lambda$ can be recomputed offline, without re-solving the optimization or retraining any model. The best operating point at a new $\lambda$ can be read off directly, rather than searched for. Figure~\ref{qslo::fig::optimal_op_lambda} plots, for each problem and method, the operating point that minimizes the mean post-hoc regret as $\lambda$ grows.

As $\lambda$ rises, restoring feasibility becomes more expensive, so in every problem the optimal operating point shifts toward the low-infeasibility end of the curve. Which direction this means for $\tau$ is problem-dependent, as established in Section~\ref{qslo::sec::paretorfronts} and visible in Figure~\ref{qslo::fig::infeasibility_vs_knob}: the optimal $\tau$ rises on the MDKP with uncertain weights and falls on the capacities problem and the BAPP, while ODECE always moves toward higher $\alpha$.

The optimal operating point largely settles by about $\lambda = 5$ in every problem, with only small shifts afterwards, but where it settles depends on how sharply infeasibility responds to the operating parameter as shown in Figure~\ref{qslo::fig::infeasibility_vs_knob}. On the MDKP with
uncertain weights, infeasibility falls steeply over a narrow band for every method, so the optimum settles in the interior and does not approach the extremes: ODECE at $\alpha \approx 0.5$, TabPFN at $\tau \approx 0.6$, LR at $\tau = 0.6$ to $0.7$, and RF, GBM, and NGBoost around $\tau \approx 0.7$ to $0.75$. Once the operating point sits just past the steep band, only a few instances remain infeasible, and pushing further buys little feasibility while paying steadily rising regret, which keeps the optimum away from the conservative extreme.

On the capacities problem only ODECE's infeasibility curve is steep (Figure~\ref{qslo::fig::infeasibility_vs_knob}), while the QSLO curves are gradual. ODECE therefore stays in the interior at $\alpha = 0.4$ to $0.5$, and its optimum barely moves with $\lambda$, while the QSLO models are driven steadily to a small buffered quantile of $\tau = 0.05$ to $0.1$ at $\lambda = 4$ and $0.025$ to $0.05$ at $\lambda = 10$.
In contrast, in the BAPP, the infeasibility curves of all methods are gradual, all lying close to the diagonal, so the operating parameters spread widely even though the methods target a
similar low-infeasibility region. NGBoost settles near $\tau = 0$ (infeasibility $0.081$, regret $0.634$) and ODECE near $\alpha \approx 0.9$ (infeasibility $0.091$, regret $0.616$). RF and TabPFN reach the same region from the interior, at $\tau \approx 0.2$, and GBM and LR settle in between, at $\tau = 0.075$ to $0.1$ and $\tau = 0.15$.

The same analysis shows when QSLO outperforms SLO. SLO yields a single operating point, the one its mean prediction happens to produce, while QSLO can move along the frontier. On the MDKP with uncertain weights at $\lambda=1$, this point lies close to the post-hoc optimum, so the two perform alike, with best post-hoc regrets within $0.004$ of each other for every model. As corrections become more expensive, SLO stays at its infeasibility rate of $0.44$ to $0.62$ and its post-hoc regret grows linearly with $\lambda$, whereas QSLO shifts to a more conservative quantile. At $\lambda=10$, QSLO's post-hoc regret is two to almost five times lower than SLO's for every model on both MDKP problems. On the capacities problem and the BAPP, QSLO is better for every model already at $\lambda=1$, since SLO's single operating point lies away from the post-hoc optimum there.

The paired tests confirm this at $\lambda=1$ (Table~\ref{tab:stat-tests}). QSLO is significantly better than SLO with the same model for every model on the capacities problem and for all models except TabPFN on the BAPP, where QSLO lowers the mean post-hoc regret from $0.207$ to $0.175$ without reaching significance. On the uncertain weights, the two differ significantly only for LR, in favor of QSLO, and for TabPFN, in favor of SLO.

Two things follow for a decision-maker. First, the looser the infeasibility response to $\tau$ or $\alpha$, the less the chosen parameter pins down the realized cost. This is clearest on the BAPP, where the shallow curves mean small differences between methods translate into large differences in the chosen parameter and the optima spread across the range (Figure~\ref{qslo::fig::infeasibility_vs_knob})

Second, sensitivity to the correction cost can be assessed after the fact. Because neither method trains on the post-hoc regret, its value at any $\lambda$ is recomputed offline from the stored test decisions, without re-solving. For a fixed operating point this is immediate. Comparing across operating points instead needs the post-hoc regret at every $\tau$ or $\alpha$, that is, the full frontier. A decision-maker who has built it, as we do, can re-select the best operating point for any $\lambda$ offline, while one with a single tuned point faces the per-level cost of Section~\ref{qslo::sec::trainingtimes}. Figure~\ref{qslo::fig::optimal_op_lambda} is this analysis.

\section{Conclusion}\label{QSLO::conclusion}

This paper has investigated a quantile-based SLO method for constrained optimization problems under parameter uncertainty. Across three benchmark problems, the MDKP with uncertain weights and uncertain capacities and the BAPP with uncertain ore quantities, QSLO achieves a significantly lower post-hoc regret than ODECE, a state-of-the-art ILO method, on both MDKP variants and a comparable one on the BAPP, while training one to almost three orders of magnitude faster when the optimization solved inside ILO's training loop is expensive. 
QSLO requires no assumptions about the optimization or problem structure, uses the solver only after training, so it can be exchanged without retraining, and gives decision-makers a single control parameter for the feasibility vs.\ (post-hoc) regret trade-off, which can be tuned to a target infeasibility rate and re-selected offline as correction costs change. Unlike SLO with a mean prediction, QSLO can move to a more conservative operating point when correcting infeasible decisions becomes more expensive. These results rest on three benchmarks, two of them synthetic, so further problem classes are needed to establish how far they generalize. Within this scope, QSLO is a lightweight alternative to ILO and a natural first method to try when the inner optimization is costly to solve.

Our analysis locates this advantage in the quantile control itself, not in a particular predictive model or a decision-focused training loss. We recall that predicting a conditional quantile of an uncertain right-hand side reproduces the corresponding chance constraint exactly, a known result from chance-constrained programming that links the quantile level to the tolerated violation probability, while in our benchmarks, with several uncertain constraints or coefficients, the quantile level serves as a control parameter selected on validation data. Empirically, no single regression model is best across all problems, but every one of them outperforms ODECE, including linear quantile regression, which uses the same model class as ODECE on the MDKP. As the MDKP data is generated by a nonlinear mapping, both are misspecified there, which is the case in which \citet{elmachtoubEstimateThenOptimizeIntegratedEstimationOptimizationSample2023} show that integrated approaches can outperform sequential ones for uncertain objectives. In the constrained setting, the quantile control more than compensates. More expressive models add to the gain where the context is informative, so the decisive factor is choosing a suitable quantile and pairing it with a model that fits the data. This challenges the common assumption that aligning the training loss with decision quality is necessary in constrained settings, and suggests that ILO's added complexity is not always warranted.

\textbf{Limitations.} We do not perform hyperparameter tuning for the regression models, and use the library defaults apart from the deviations stated in Section~\ref{QSLO::experiments}. Likewise, we run ODECE with the predictors of its original setup, a linear model on the MDKP and an MLP on the BAPP, and do not test whether a larger predictor would change its results. TabPFN requires a GPU for efficient inference and is limited in the number of features and training instances it can accommodate. NGBoost requires the user to specify a parametric output distribution, and a misspecified distribution can degrade predictive quality. We evaluate only settings where uncertainty appears in the constraints. Extending the approach to objective-function uncertainty is supported by the implementation but not studied empirically here. The three benchmarks differ in how much of the parameter variation is recoverable from the contextual features, which is strong for the MDKP with uncertain weights, moderate for the uncertain capacities, and weak for the BAPP, where a context-free quantile performs on par with the learned models. We adopt these instances from prior work for comparability, correcting only the MDKP-C train and test split (Section~\ref{QSLO::experiments}). The benchmarks are small and two of them are synthetic, so a larger suite with consistently strong context-dependence is needed to test the predictive component more sharply and to show how far our findings generalize. Finally, unsolvable instances remain a practical challenge for any predict-then-optimize approach, and we recommend that practitioners apply domain-informed prediction bounds and establish a fallback strategy, such as a heuristic solution or human escalation.

\textbf{Open questions.} All three benchmarks use artificially set correction costs, and two rely on synthetic data generated from known distributions. Extending the comparison to larger, real-world problems, where uncertainty must be estimated from historical data, correction costs reflect actual operational penalties, and distributions may shift over time, would provide stronger validation. A further direction is to learn instance-specific quantiles from contextual features, rather than applying a fixed $\tau$ across all instances, or to tune separate $\tau$ values for different groups of uncertain constraints. Care is needed here when multiple uncertain parameters must be predicted jointly, since the number of quantile combinations grows exponentially with the number of parameters.

\section*{Declaration of Conflict of Interest}
The authors declare that they have no known competing financial interests or personal relationships that could have appeared to influence the work reported in this paper.

\section*{Declaration of Use of generative AI and AI-assisted technologies}
During the preparation of this work the authors used Claude (Anthropic) in order to improve the language and readability. After using this tool, the authors reviewed and edited the content as needed and take full responsibility for the content of the published article.

\section*{Acknowledgments}
All experiments were conducted on the Noctua 2 cluster at the Paderborn Center for Parallel Computing (PC2). This work was funded by the Deutsche Forschungsgemeinschaft (DFG) under grant 540478491. We thank Jayanta Mandi and colleagues for providing the corrected datasets used in the ODECE benchmark experiments.

\bibliographystyle{plainnat}
\bibliography{cas-refs}

@article{sadanaSurveyContextualOptimization2025,
  title = {A Survey of Contextual Optimization Methods for Decision-Making under Uncertainty},
  author = {Sadana, Utsav and Chenreddy, Abhilash and Delage, Erick and Forel, Alexandre and Frejinger, Emma and Vidal, Thibaut},
  year = 2025,
  month = jan,
  journal = {European Journal of Operational Research},
  volume = {320},
  number = {2},
  pages = {271--289},
  issn = {0377-2217},
  doi = {https://doi.org/10.1016/j.ejor.2024.03.020},
  urldate = {2024-10-31}
}

@article{Charnes1963,
author = {Charnes, A. and Cooper, W. W.},
title = {Deterministic Equivalents for Optimizing and Satisficing under Chance Constraints},
journal = {Operations Research},
volume = {11},
number = {1},
pages = {18-39},
year = {1963},
doi = {10.1287/opre.11.1.18},
}

@article{holm1979simple,
 ISSN = {03036898, 14679469},
 URL = {http://www.jstor.org/stable/4615733},
 author = {Sture Holm},
 journal = {Scandinavian Journal of Statistics},
 number = {2},
 pages = {65--70},
 publisher = {[Board of the Foundation of the Scandinavian Journal of Statistics, Wiley]},
 title = {A Simple Sequentially Rejective Multiple Test Procedure},
 urldate = {2026-11-09},
 volume = {6},
 year = {1979}
}

@article{mandiDecisionFocusedLearningFoundations2024,
  title={Decision-focused learning: Foundations, state of the art, benchmark and future opportunities},
  author={Mandi, Jayanta and Kotary, James and Berden, Senne and Mulamba, Maxime and Bucarey, Victor and Guns, Tias and Fioretto, Ferdinando},
  journal={Journal of Artificial Intelligence Research},
  volume={80},
  pages={1623--1701},
  year={2024},
  doi = { https://doi.org/10.1613/jair.1.15320}
}

@inproceedings{berthetLearningDifferentiablePerturbed2020,
 author = {Berthet, Quentin and Blondel, Mathieu and Teboul, Olivier and Cuturi, Marco and Vert, Jean-Philippe and Bach, Francis},
 booktitle = {Advances in Neural Information Processing Systems},
 editor = {H. Larochelle and M. Ranzato and R. Hadsell and M.F. Balcan and H. Lin},
 pages = {9508--9519},
 publisher = {Curran Associates, Inc.},
 title = {Learning with Differentiable Pertubed Optimizers},
  url = {https://proceedings.neurips.cc/paper_files/paper/2020/file/6bb56208f672af0dd65451f869fedfd9-Paper.pdf},
 volume = {33},
 year = {2020},

}

@article{grinsztajn2022tree,
  title={Why do tree-based models still outperform deep learning on typical tabular data?},
  author={Grinsztajn, L{\'e}o and Oyallon, Edouard and Varoquaux, Ga{\"e}l},
  journal={Advances in neural information processing systems},
  volume={35},
  pages={507--520},
  year={2022}
}

@article{koenkerQuantileRegression402017,
  title={Quantile regression: 40 years on},
  author={Koenker, Roger},
  journal={Annual review of economics},
  volume={9},
  pages={155--176},
  year={2017},
  publisher={Annual Reviews},
 doi = {10.1146/annurev-economics-063016-103651},

}

@article{elmachtoubSmartPredictThen2022,
  title={Smart “predict, then optimize”},
  author={Elmachtoub, Adam N and Grigas, Paul},
  journal={Management Science},
  volume={68},
  number={1},
  pages={9--26},
  year={2022},
  publisher={INFORMS},
doi = {https://doi.org/10.1287/mnsc.2020.3922},

}

@article{tangPyEPOPyTorchbasedEndend2024,
  title = {{{PyEPO}}: A {{PyTorch}}-Based End-to-End Predict-Then-Optimize Library for Linear and Integer Programming},
  author = {Tang, Bo and Khalil, Elias B.},
  year = 2024,
  month = sep,
  journal = {Mathematical Programming Computation},
  volume = {16},
  number = {3},
  pages = {297--335},
  issn = {1867-2957},
  doi = {10.1007/s12532-024-00255-x}
}

@article{vlastelicaDifferentiationBlackboxCombinatorial2020,
  title={Differentiation of blackbox combinatorial solvers},
  author={Vlastelica, Marin and Paulus, Anselm and Musil, V{\'\i}t and Martius, Georg and Rol{\'\i}nek, Michal},
  journal={arXiv preprint arXiv:1912.02175},
  year={2019},
  doi={https://doi.org/10.48550/arXiv.1912.02175}
}

@article{wangPredictThenOptimize2022,
title={``Predict, then optimize'' with quantile regression: A global method from predictive to prescriptive analytics and applications to multimodal transportation},
journal = {Multimodal Transportation},
volume = {1},
number = {4},
pages = {100035},
year = {2022},
issn = {2772-5863},
doi = {https://doi.org/10.1016/j.multra.2022.100035},
author = {Shuaian Wang and Ran Yan}
}

@article{mahmutogullariLeveragingContextualInformation2023,
  title={Leveraging Contextual Information for Robustness in Vehicle Routing Problems},
  author={Mahmuto{\u{g}}ullar{\i}, Ali {\.I}rfan and Guns, Tias},
  journal={arXiv preprint arXiv:2310.17368},
  year={2023},
  doi={https://doi.org/10.48550/arXiv.2310.17368}
}

@article{gengBenchmarkingPtOPnO,
  title={Benchmarking pto and pno methods in the predictive combinatorial optimization regime},
  author={Geng, Haoyu and Ruan, Hang and Wang, Runzhong and Li, Yang and Wang, Yang and Chen, Lei and Yan, Junchi},
  journal={Advances in Neural Information Processing Systems},
  volume={37},
  pages={65944--65971},
  year={2024},
  doi={https://doi.org/10.52202/079017-2108}
  
}

@inproceedings{mandiFeasibilityAwareDecisionFocusedLearning2025,
 author = {Mandi, Jayanta and Defresne, Marianne and Berden, Senne and Guns, Tias},
 booktitle = {Advances in Neural Information Processing Systems},
 doi = {10.52202/085713-5451},
 editor = {D. Belgrave and C. Zhang and H. Lin and R. Pascanu and P. Koniusz and M. Ghassemi and N. Chen},
 pages = {163285--163309},
 publisher = {Curran Associates, Inc.},
 title = {Feasibility-Aware Decision-Focused Learning for Predicting Parameters in the Constraints},
 url = {https://proceedings.neurips.cc/paper_files/paper/2025/file/ef158a0d4f5364741a571b8d1d44fb1b-Paper-Conference.pdf},
 volume = {38, Main Conference},
 year = {2025}
}

@inproceedings{huBranchLearnRecursively2022,
 author = {Hu, Xinyi and Lee, Jasper and Lee, Jimmy and Zhong, Allen Z.},
 booktitle = {Advances in Neural Information Processing Systems},
 pages = {25807--25817},
 publisher = {Curran Associates, Inc.},
 title = {Branch \& Learn for Recursively and Iteratively Solvable Problems in Predict+Optimize},
 volume = {35},
 year = {2022}
}

@inproceedings{huBranchLearnPosthoc2023,
  title={Branch \& learn with post-hoc correction for predict+ optimize with unknown parameters in constraints},
  author={Hu, Xinyi and Lee, Jasper CH and Lee, Jimmy HM},
  booktitle={Integration of Constraint Programming, Artificial Intelligence, and Operations Research},
  pages={264--280},
  year={2023},
  organization={Springer},
  publisher={Springer Nature Switzerland},
 doi={https://doi.org/10.1007/978-3-031-33271-5_18}
}

@article{huPredict+OptimizePackingCovering2023, 
title={Predict+Optimize for Packing and Covering LPs with Unknown Parameters in Constraints}, 
volume={37}, 
DOI={https://doi.org/10.1609/aaai.v37i4.25513}, 
number={4}, 
journal={Proceedings of the AAAI Conference on Artificial Intelligence}, 
author={Hu, Xinyi and Lee, Jasper C.H. and Lee, Jimmy H.M.}, 
year={2023}, 
month={Jun.}, 
pages={3987-3995} }

@article{huTwoStagePredict+OptimizeMixed2023,
  title={Multi-stage predict+ optimize for (mixed integer) linear programs},
  author={Hu, Xinyi and Lee, Jasper C and Lee, Jimmy H and Stuckey, Peter J},
  journal={Advances in Neural Information Processing Systems},
  volume={37},
  pages={64794--64827},
  year={2024},
  doi = {https://doi.org/10.52202/079017-2068}
}

@inproceedings{berdenSolverFreeDecisionFocusedLearning2025,
author = {Berden, Senne and Mahmuto{\u{g}}ullar{\i}, Ali {\.I}rfan and Tsouros, Dimos and Guns, Tias},
title = {Solver-free decision-focused learning for linear optimization problems},
year = {2026},
publisher = {Curran Associates Inc.},
address = {Red Hook, NY, USA},
url = {https://doi.org/10.52202/085713-5284},
booktitle = {Proceedings of the 39th International Conference on Neural Information Processing Systems},
articleno = {4783},
numpages = {19}
}

@article{garciaApplicationDrivenLearningClosedLoop2024,
author = {Dias Garcia, Joaquim and Street, Alexandre and Homem-de-Mello, Tito and Mu\~{n}oz, Francisco D.},
title = {Application-Driven Learning: A Closed-Loop Prediction and Optimization Approach Applied to Dynamic Reserves and Demand Forecasting},
journal = {Operations Research},
volume = {73},
number = {1},
pages = {22-39},
year = {2025},
doi = {https://doi.org/10.1287/opre.2023.0565},
publisher={INFORMS}
}

@inproceedings{paulusCombOptNetFitRight2022,
  title = 	 {{{CombOptNet}}: Fit the Right NP-Hard Problem by Learning Integer Programming Constraints},
  author =       {Paulus, Anselm and Rolinek, Michal and Musil, Vit and Amos, Brandon and Martius, Georg},
  booktitle = 	 {Proceedings of the 38th International Conference on Machine Learning},
  pages = 	 {8443--8453},
  year = 	 {2021},
  editor = 	 {Meila, Marina and Zhang, Tong},
  volume = 	 {139},
  series = 	 {Proceedings of Machine Learning Research},
  month = 	 {18--24 Jul},
  publisher =    {PMLR},
}

@article{shahDecisionFocusedLearningDifferentiable,
  title={Decision-focused learning without decision-making: Learning locally optimized decision losses},
  author={Shah, Sanket and Wang, Kai and Wilder, Bryan and Perrault, Andrew and Tambe, Milind},
  journal={Advances in Neural Information Processing Systems},
  volume={35},
  pages={1320--1332},
  year={2022}
}

@inproceedings{beichterDecisionFocusedRetrainingForecast2024,
author = {Beichter, Maximilian and Werling, Dorina and Heidrich, Benedikt and Phipps, Kaleb and Neumann, Oliver and Friederich, Nils and Mikut, Ralf and Hagenmeyer, Veit},
title = {Decision-Focused Retraining of Forecast Models for Optimization Problems in Smart Energy Systems},
year = {2024},
isbn = {9798400704802},
publisher = {Association for Computing Machinery},
address = {New York, NY, USA},
doi = {https://doi.org/10.1145/3632775.3661952},
booktitle = {Proceedings of the 15th ACM International Conference on Future and Sustainable Energy Systems},
pages = {170–181},
numpages = {12},
location = {Singapore, Singapore},
series = {e-Energy '24}
}

@article{yangDFFDecisionFocusedFinetuning2025, 
title={DFF: Decision-Focused Fine-Tuning for Smarter Predict-Then-Optimize with Limited Data}, 
volume={39}, 
DOI={https://doi.org/10.1609/aaai.v39i25.34891}, 
number={25}, 
journal={Proceedings of the AAAI Conference on Artificial Intelligence}, 
author={Yang, Jiaqi and Liang, Enming and Su, Zicheng and Zou, Zhichao and Zhen, Peng and Guo, Jiecheng and Ma, Wanjing and An, Kun}, 
year={2025}, 
month={Apr.}, 
pages={26868-26876} }

@article{demiroviDynamicProgrammingPredictOptimise2020, 
title={Dynamic Programming for Predict+Optimise}, volume={34}, 
DOI={https://doi.org/10.1609/aaai.v34i02.5502}, 
number={02}, journal={Proceedings of the AAAI Conference on Artificial Intelligence}, 
author={Demirovic, Emir and Stuckey, Peter J. and Guns, Tias and Bailey, James and Leckie, Christopher and Ramamohanarao, Kotagiri and Chan, Jeffrey}, 
year={2020}, month={Apr.}, pages={1444-1451} }

@inproceedings{kotaryEndtoEndConstrainedOptimization2021,
  title     = {End-to-End Constrained Optimization Learning: A Survey},
  author    = {Kotary, James and Fioretto, Ferdinando and Van Hentenryck, Pascal and Wilder, Bryan},
  booktitle = {Proceedings of the Thirtieth International Joint Conference on
               Artificial Intelligence, {IJCAI-21}},
  publisher = {International Joint Conferences on Artificial Intelligence Organization},
  editor    = {Zhi-Hua Zhou},
  pages     = {4475--4482},
  year      = {2021},
  month     = {8},
  note      = {Survey Track},
  doi       = {10.24963/ijcai.2021/610},
  url       = {https://doi.org/10.24963/ijcai.2021/610},
}

@article{elmachtoubEstimateThenOptimizeIntegratedEstimationOptimizationSample2023,
      title={Estimate-Then-Optimize versus Integrated-Estimation-Optimization versus Sample Average Approximation: A Stochastic Dominance Perspective}, 
      author={Adam N. Elmachtoub and Henry Lam and Haofeng Zhang and Yunfan Zhao},
      year={2025},
  journal={arXiv preprint arXiv:2304.06833},
      doi={
https://doi.org/10.48550/arXiv.2304.06833
}
}

@inproceedings{mandiInteriorPointSolving2020,
 author = {Mandi, Jayanta and Guns, Tias},
 booktitle = {Advances in Neural Information Processing Systems},
 editor = {H. Larochelle and M. Ranzato and R. Hadsell and M.F. Balcan and H. Lin},
 pages = {7272--7282},
 publisher = {Curran Associates, Inc.},
 title = {Interior Point Solving for LP-based prediction+optimisation},
 volume = {33},
 year = {2020}
}

@article{wilderMeldingDataDecisionsPipeline2018, 
title={Melding the Data-Decisions Pipeline: Decision-Focused Learning for Combinatorial Optimization}, 
volume={33}, 
DOI={https://doi.org/10.1609/aaai.v33i01.33011658}, 
number={01}, 
journal={Proceedings of the AAAI Conference on Artificial Intelligence}, 
author={Wilder, Bryan and Dilkina, Bistra and Tambe, Milind}, 
year={2019}, month={Jul.}, pages={1658-1665} }

@article{vanderschuerenPredictthenoptimizePredictandoptimizeEmpirical2022,
author = {Vanderschueren, Toon and Verdonck, Tim and Baesens, Bart and Verbeke, Wouter},
title = {Predict-then-optimize or predict-and-optimize? An empirical evaluation of cost-sensitive learning strategies},
year = {2022},
issue_date = {May 2022},
publisher = {Elsevier Science Inc.},
address = {USA},
volume = {594},
number = {C},
issn = {0020-0255},
doi = {https://doi.org/10.1016/j.ins.2022.02.021},
journal = {Inf. Sci.},
month = may,
pages = {400–415},
numpages = {16}
}

@inproceedings{schutteRobustLossesDecisionFocused2024,
  title = {Robust Losses for Decision-Focused Learning},
  booktitle = {Proceedings of the {{Thirty-Third International Joint Conference}} on {{Artificial Intelligence}}},
  author = {Schutte, Noah and Postek, Krzysztof and Yorke-Smith, Neil},
  date = {2024-08},
  year = {2024},
  pages = {4868--4875},
  publisher = {International Joint Conferences on Artificial Intelligence Organization},
  location = {Jeju, South Korea},
  doi = {https://doi.org/10.24963/ijcai.2024/538},
  eventtitle = {Thirty-{{Third International Joint Conference}} on {{Artificial Intelligence}} \{{{IJCAI-24}}\}},
}

@article{ScenarioPredictthenOptimizeDataDriven,
author = {Jia, Menglei and Schrotenboer, Albert H. and Chen, Feng},
title = {Scenario Predict-then-Optimize for Data-Driven Online Inventory Routing},
journal = {Transportation Science},
publisher={INFORMS},
volume = {59},
number = {5},
pages = {1032-1056},
year = {2025},
doi = {https://doi.org/10.1287/trsc.2024.0613}
}

@article{silvestriScoreFunctionGradient2024,
  title={Score function gradient estimation to widen the applicability of decision-focused learning},
  author={Silvestri, Mattia and Berden, Senne and Signorelli, Gaetano and Mahmuto{\u{g}}ullar{\i}, Ali {\.I}rfan and Mandi, Jayanta and Amos, Brandon and Guns, Tias and Lombardi, Michele},
  year = 2026,
  month = feb,
  journal = {Journal of Artificial Intelligence Research},
  volume = {85},
  issn = {1076-9757},
  doi = {10.1613/jair.1.19498},
}

@inproceedings{schutteSufficientDecisionProxies2026, series={IJCAI-2026},
   title={Sufficient Decision Proxies for Decision-Focused Learning},
   url={http://dx.doi.org/10.24963/ijcai.2026/260},
   DOI={10.24963/ijcai.2026/260},
   booktitle={Proceedings of the Thirty-Fifth International Joint Conference on Artificial Intelligence},
   publisher={International Joint Conferences on Artificial Intelligence Organization},
   author={Schutte, Noah and Veviurko, Grigorii and Postek, Krzysztof and Yorke-Smith, Neil},
   year={2026},
   month=sep, pages={2337--2345},
   collection={IJCAI-2026} }

@inproceedings{brandoUsingQuantileRegression2022,
  author    = {Brando, A. and Serra, I. and Mezzetti, E. and Abella, J. and Cazorla, F.J.},
  title     = {Using Quantile Regression in Neural Networks for Contention Prediction in Multicore Processors},
  booktitle = {34th Euromicro Conference on Real-Time Systems (ECRTS 2022)},
  series    = {Leibniz International Proceedings in Informatics (LIPIcs)},
  volume    = {231},
  pages     = {4:1--4:25},
  publisher = {Schloss Dagstuhl -- Leibniz-Zentrum f{\"u}r Informatik},
  address   = {Dagstuhl, Germany},
  year      = {2022},
  doi       = {https://doi.org/10.4230/LIPIcs.ECRTS.2022.4},
}

@inproceedings{fernandesUncertaintyDisplaysUsing2018,
author = {Fernandes, Michael and Walls, Logan and Munson, Sean and Hullman, Jessica and Kay, Matthew},
title = {Uncertainty Displays Using Quantile Dotplots or CDFs Improve Transit Decision-Making},
year = {2018},
isbn = {9781450356206},
publisher = {Association for Computing Machinery},
address = {New York, NY, USA},
url = {https://doi.org/10.1145/3173574.3173718},
doi = {10.1145/3173574.3173718},
booktitle = {Proceedings of the 2018 CHI Conference on Human Factors in Computing Systems},
pages = {1–12},
numpages = {12},
location = {Montreal QC, Canada},
series = {CHI '18}
}

@inproceedings{sunPredictthenCalibrateNewPerspective,
 author = {Sun, Chunlin and Liu, Linyu and Li, Xiaocheng},
 booktitle = {Advances in Neural Information Processing Systems},
 editor = {A. Oh and T. Naumann and A. Globerson and K. Saenko and M. Hardt and S. Levine},
 pages = {17713--17741},
 publisher = {Curran Associates, Inc.},
 title = {Predict-then-Calibrate: A New Perspective of Robust Contextual LP},
 volume = {36},
 year = {2023}
}

@InProceedings{jadonComprehensiveSurveyRegression2022,
author="Jadon, Aryan
and Patil, Avinash
and Jadon, Shruti",
editor="Sharma, Neha
and Goje, Amol C.
and Chakrabarti, Amlan
and Bruckstein, Alfred M.",
title="A Comprehensive Survey of Regression-Based Loss Functions for Time Series Forecasting",
booktitle="Data Management, Analytics and Innovation",
year="2024",
publisher="Springer Nature Singapore",
address="Singapore",
pages="117--147",
doi="https://doi.org/10.1007/978-981-97-3245-6_9"
}

@inproceedings{wangCostSensitiveOperationalManagement2026,
author = {Wang, Xinyu},
title = {Cost-Sensitive Operational Management Demand Forecasting: A Minimalist Framework Based on Naive Seasonality and Residual GRU},
year = {2025},
isbn = {9798400719516},
publisher = {Association for Computing Machinery},
address = {New York, NY, USA},
doi = {https://doi.org/10.1145/3785706.3785813},
booktitle = {Proceedings of the 2025 2nd International Conference on Digital Economy and Computer Science},
pages = {684–689},
numpages = {6},
series = {DECS '25}
}

@article{dressResidualValueForecasting2018,
title = {Residual value forecasting using asymmetric cost functions},
journal = {International Journal of Forecasting},
volume = {34},
number = {4},
pages = {551-565},
year = {2018},
issn = {0169-2070},
doi = {https://doi.org/10.1016/j.ijforecast.2018.01.008},
author = {Korbinian Dress and Stefan Lessmann and Hans-Jörg {von Mettenheim}},
}

@article{wangQuantileRegressionAnalyzing2012,
  author       = {Wang, Lan and Wu, Yichao and Li, Runze},
  title        = {Quantile Regression for Analyzing Heterogeneity in Ultra-high Dimension},
  journal      = {Journal of the American Statistical Association},
  year         = {2012},
  volume       = {107},
  number       = {497},
  pages        = {214--222},
  month        = {March},
  doi          = {https://doi.org/10.1080/01621459.2012.656014},
  issn         = {0162-1459},
  publisher    = {American Statistical Association}
}

@book{hastie2009elements,
  title={The elements of statistical learning: data mining, inference, and prediction},
  author={Hastie, Trevor and Tibshirani, Robert and Friedman, Jerome H and Friedman, Jerome H},
  volume={2},
  year={2009},
  publisher={Springer},
  doi={https://doi.org/10.1007/978-0-387-84858-7}
}

@article{breiman2001random,
  title   = {Random Forests},
  author  = {Breiman, Leo},
  journal = {Machine Learning},
  volume  = {45},
  number  = {1},
  pages   = {5--32},
  year    = {2001},
  publisher={Springer},
  doi={https://doi.org/10.1023/A:1010933404324}

}

@article{meinshausen2006quantile,
  author  = {Nicolai Meinshausen},
  title   = {Quantile Regression Forests},
  journal = {Journal of Machine Learning Research},
  year    = {2006},
  volume  = {7},
  number  = {35},
  pages   = {983--999},
  url     = {http://jmlr.org/papers/v7/meinshausen06a.html}
}

@article{belloniL1penalizedQuantileRegression2011,
  title = {{$\ell$}1-Penalized Quantile Regression in High-Dimensional Sparse Models},
  author = {Belloni, Alexandre and Chernozhukov, Victor},
  year = 2011,
  month = feb,
  journal = {The Annals of Statistics},
  volume = {39},
  number = {1},
  pages = {82--130},
  publisher = {Institute of Mathematical Statistics},
  issn = {0090-5364, 2168-8966},
  doi = {10.1214/10-AOS827},
  urldate = {2026-10-28},
  langid = {english}
}

@article{friedman2001greedy,
  title={Greedy function approximation: a gradient boosting machine},
  author={Friedman, Jerome H},
  journal={Annals of statistics},
  pages={1189--1232},
  year={2001},
  publisher={JSTOR},
  doi={ https://doi.org/10.1214/aos/1013203451}
}

@article{Jatty2023Diagnosis,
title={Diagnosis of linear programming supply chain optimization models: Detecting infeasibilities and minimizing changes for new parameter values},
author={Sitoshna Jatty and Niharika Singh and I. Grossmann and Leonardo Salsano de Assis and C. Galanopoulos and Pablo García-Herreros and B. Springub and Nga Tran},
journal={Comput. Chem. Eng.},
year={2023},volume={171},pages={108139},
doi={https://doi.org/10.1016/j.compchemeng.2023.108139}}

@article{Kononova2022The,
title={The Importance of Being Constrained: Dealing with Infeasible Solutions in Differential Evolution and Beyond},
author={Anna V. Kononova and Diederick Vermetten and Fabio Caraffini and Madalina Mitran and D. Zaharie},
journal={Evolutionary Computation},year={2022},volume={32},pages={3-48},
doi={https://doi.org/10.1162/evco_a_00333}}

@article{Meselhi2022Modified,
  title = {Modified Differential Evolution Algorithm for Solving Dynamic Optimization with Existence of Infeasible Environments},
  author = {Meselhi, Mohamed A. and Elsayed, Saber M. and Essam, Daryl L. and Sarker, Ruhul A.},
  year = 2022,
  journal = {Computers, Materials and Continua},
  volume = {74},
  number = {1},
  pages = {1--17},
  issn = {1546-2218},
  doi = {10.32604/cmc.2023.027448},
}

@Article{Diz2018A,
journal={Flexible Services and Manufacturing Journal},
author={Gustavo Souto dos Santos Diz and Silvio Hamacher and Fabricio Oliveira},
title={A robust optimization model for the maritime inventory routing problem},
year={2019},
month={September},
pages={675-701},
volume={31},
number={3},
doi={10.1007/s10696-018-9327-9},
}

@article{Steinwart2011Estimating,
title={Estimating conditional quantiles with the help of the pinball loss},
author={Ingo Steinwart and A. Christmann},journal={Bernoulli},year={2011},volume={17},pages={211-225},
doi={https://doi.org/10.3150/10-bej267}}

@inproceedings{duan2020ngboost,
  title =  {{NGB}oost: Natural Gradient Boosting for Probabilistic Prediction},
  author =  {Duan, Tony and Anand, Avati and Ding, Daisy Yi and Thai, Khanh K. and Basu, Sanjay and Ng, Andrew and Schuler, Alejandro},
  booktitle = {Proceedings of the 37th International Conference on Machine Learning},
  pages = {2690--2700},
  year = {2020},
  editor = {III, Hal Daumé and Singh, Aarti},
  volume = {119},
  series = {Proceedings of Machine Learning Research},
  month =  {13--18 Jul},
  publisher =  {PMLR},
  doi={https://doi.org/10.48550/arXiv.1910.03225}
}

@article{grinsztajn2025tabpfn,
  title={Tabpfn-2.5: Advancing the state of the art in tabular foundation models},
  author={Grinsztajn, L{\'e}o and Fl{\"o}ge, Klemens and Key, Oscar and Birkel, Felix and Jund, Philipp and Roof, Brendan and J{\"a}ger, Benjamin and Safaric, Dominik and Alessi, Simone and Hayler, Adrian and Manium, Mihir and Yu, Rosen and Jablonski, Felix and Hoo, Shi Bin and Garg, Anurag and Robertson, Jake and B{\"u}hler, Magnus and Moroshan, Vladyslav and Purucker, Lennart and Cornu, Clara and Wehrhahn, Lilly Charlotte and Bonetto, Alessandro and Sch{\"o}lkopf, Bernhard and Gambhir, Sauraj and Hollmann, Noah and Hutter, Frank},
  journal={arXiv preprint arXiv:2511.08667},
  year={2025},
  doi={https://doi.org/10.48550/arXiv.2511.08667}

}

@article{hollmann2025tabpfn,
 title={Accurate predictions on small data with a tabular foundation model},
 author={Hollmann, Noah and M{\"u}ller, Samuel and Purucker, Lennart and
         Krishnakumar, Arjun and K{\"o}rfer, Max and Hoo, Shi Bin and
         Schirrmeister, Robin Tibor and Hutter, Frank},
 journal={Nature},
 year={2025},
 month={01},
 day={09},
 doi={https://doi.org/10.1038/s41586-024-08328-6},
 publisher={Springer Nature},
}

@inproceedings{hollmann2023tabpfn,
  title={TabPFN: A transformer that solves small tabular classification problems in a second},
  author={Hollmann, Noah and M{\"u}ller, Samuel and Eggensperger, Katharina and Hutter, Frank},
  booktitle={International Conference on Learning Representations 2023},
  year={2023},
  pages={1--21},
}

@inproceedings{hu2023twostage,
 author = {Hu, Xinyi and Lee, Jasper and Lee, Jimmy},
 booktitle = {Advances in Neural Information Processing Systems},
 editor = {A. Oh and T. Naumann and A. Globerson and K. Saenko and M. Hardt and S. Levine},
 pages = {14247--14272},
 publisher = {Curran Associates, Inc.},
 title = {Two-Stage Predict+Optimize for MILPs with Unknown Parameters in Constraints},
 volume = {36},
 year = {2023}
}

@article{probst2018tune,
  author  = {Philipp Probst and Anne-Laure Boulesteix},
  title   = {To Tune or Not to Tune the Number of Trees in Random Forest},
  journal = {Journal of Machine Learning Research},
  year    = {2018},
  volume  = {18},
  number  = {181},
  pages   = {1--18},
  url     = {http://jmlr.org/papers/v18/17-269.html}
}

@article{probstHyperparametersTuningStrategies2019,
  title = {Hyperparameters and {{Tuning Strategies}} for {{Random Forest}}},
  author = {Probst, Philipp and Wright, Marvin and Boulesteix, Anne-Laure},
  year = 2019,
  month = may,
  journal = {WIREs Data Mining and Knowledge Discovery},
  volume = {9},
  number = {3},
  primaryclass = {stat.ML},
  pages = {e1301},
  issn = {1942-4787, 1942-4795},
  doi = {10.1002/widm.1301},
  urldate = {2026-10-28},
}

@book{birge1997introduction,
  title={Introduction to stochastic programming},
  author={Birge, John R and Louveaux, Francois},
  year={1997},
  publisher={Springer},
  doi={https://doi.org/10.1007/978-1-4614-0237-4},
  isbn={978-1-4614-0236-7}
}

@book{Shapiro2014Lectures ,
author = {Shapiro, Alexander and Dentcheva, Darinka and Ruszczyński, Andrzej},
title = {Lectures on Stochastic Programming: Modeling and Theory, Second Edition},
publisher = {Society for Industrial and Applied Mathematics},
year = {2014},
doi = {https://doi.org/10.1137/1.9781611973433},
address = {Philadelphia, PA},
edition   = {}
}

@article{Kleywegt2002The,
author = {Kleywegt, Anton J. and Shapiro, Alexander and Homem-de-Mello, Tito},
title = {The Sample Average Approximation Method for Stochastic Discrete Optimization},
journal = {SIAM Journal on Optimization},
volume = {12},
number = {2},
pages = {479-502},
year = {2002},
doi = {https://doi.org/10.1137/S1052623499363220}
}

@inproceedings{romer2023predict,
author = {R{\"o}mer, Michael and Hagemann, Felix and Porrmann, Till Frederik},
title = {{Predict, Tune and Optimize for Data-Driven Shift Scheduling with Uncertain Demands}},
year = {2023},
isbn = {978-3-031-44504-0},
publisher = {Springer-Verlag},
address = {Berlin, Heidelberg},
url = {https://doi.org/10.1007/978-3-031-44505-7_18},
doi = {10.1007/978-3-031-44505-7_18},
booktitle = {Learning and Intelligent Optimization: 17th International Conference, LION 17, Nice, France, June 4–8, 2023, Revised Selected Papers},
pages = {254--269},
numpages = {16}
}

@article{li2025ridehailing,
  author  = {Li, Xiang and Jin, Kun and Ma, Hongguang and Feng, Ziyan and Xia, Yusen},
  title   = {Data-Driven Capacity Management of Ride-Hailing Service with Prescriptive Analytics and Deep Learning},
  journal = {INFORMS Journal on Computing},
  year    = {2026},
  note    = {Articles in Advance},
  doi     = {10.1287/ijoc.2024.0595},
}

@article{CharnesCooper1959,
  author  = {Charnes, A. and Cooper, W. W.},
  title   = {Chance-Constrained Programming},
  journal = {Management Science},
  year    = {1959},
  volume  = {6},
  number  = {1},
  pages   = {73--79},
  doi     = {10.1287/mnsc.6.1.73}
}

@book{Prekopa1995,
  author    = {Pr\'ekopa, A.},
  title     = {Stochastic Programming},
  series    = {Mathematics and Its Applications},
  volume    = {324},
  publisher = {Springer},
  address   = {Dordrecht},
  year      = {1995},
  doi       = {10.1007/978-94-017-3087-7}
}

@ARTICLE{zitzlerMultiobjectiveEvolutionaryAlgorithms1999,
  author={Zitzler, Eckart and Thiele, Lothar},
  journal={IEEE Transactions on Evolutionary Computation}, 
  title={Multiobjective evolutionary algorithms: a comparative case study and the strength Pareto approach}, 
  year={1999},
  volume={3},
  number={4},
  pages={257-271},
  doi={10.1109/4235.797969},
  publisher={IEEE}
  }

@article{NemirovskiShapiro2006,
  author  = {Nemirovski, A. and Shapiro, A.},
  title   = {Convex Approximations of Chance Constrained Programs},
  journal = {SIAM Journal on Optimization},
  year    = {2006},
  volume  = {17},
  number  = {4},
  pages   = {969--996},
  doi     = {10.1137/050622328}
}

@article{cameron2022perils,
  title = {The Perils of Learning before Optimizing},
  author = {Cameron, Chris and Hartford, Jason and Lundy, Taylor and {Leyton-Brown}, Kevin},
  year = 2022,
  month = jun,
  journal = {Proceedings of the AAAI Conference on Artificial Intelligence},
  volume = {36},
  number = {4},
  pages = {3708--3715},
  doi = {10.1609/aaai.v36i4.20284},
}

@article{homem-de-melloForecastingOutsideBox2026,
  title = {Forecasting Outside the {{Box}}: Application-Driven Optimal Pointwise Forecasts for a Class of Two-Stage Stochastic Programs},
  shorttitle = {Forecasting Outside the {{Box}}},
  author = {{Homem-de-Mello}, Tito and Valencia, Juan and Lagos, Felipe and Lagos, Guido},
  year = 2026,
  month = may,
  journal = {Mathematical Programming},
  issn = {1436-4646},
  doi = {10.1007/s10107-026-02359-9},
  urldate = {2026-10-21},
  langid = {english},
}

@inproceedings{imSmartSurrogateLosses2025,
  title = {Smart {{Surrogate Losses}} for {{Contextual Stochastic Linear Optimization}} with {{Robust Constraints}}},
  booktitle = {Advances in {{Neural Information Processing Systems}} 38},
  author = {Im, Hyungki and Benslimane, Wyame and Grigas, Paul},
  year = 2025,
  pages = {83686--83711},
  publisher = {Neural Information Processing Systems Foundation, Inc. (NeurIPS)},
  address = {San Diego, California, USA and Mexico City, Mexico},
  doi = {10.52202/085713-2523},
  urldate = {2026-10-21},
  isbn = {979-8-3313-3827-5},
  langid = {english},
}

@inproceedings{zhaoDiffusionDFLDecisionfocusedDiffusion2025,
  title={Diffusion-DFL: decision-focused diffusion models for stochastic optimization},
  author={Zhao, Zihao and Yeh, Christopher and Kong, Lingkai and Wang, Kai},
  booktitle={International Conference on Learning Representations},
  volume={2026},
  pages={36834--36859},
  year={2026}
}
\appendix
\newpage

\section{Appendix}
This appendix provides supplementary material supporting the experiments in Section~\ref{QSLO::experiments}. Appendix~\ref{qslo::app::datagen} documents the instance-generation procedure for both problems as it is implemented in the code we build on. Appendix~\ref{qslo::app::hyperparameters} lists the exact package versions and hyperparameter settings used for all regression models (Table~\ref{qslo::app::hyperparameters_table}). Appendix~\ref{qslo:app:fullresults} reports the full per-configuration results (Tables~\ref{QSLO::app_WeightsTable_1} to~\ref{QSLO::app_AlloyTable_2}), including infeasibility, regret, post-hoc regret, and wall, CPU, and GPU time, for every quantile level $\tau$ and infeasibility aversion coefficient $\alpha$ evaluated.

\subsection{Instance Generation}\label{qslo::app::datagen}

For the MDKP we use the instance-generation code released by \citet{mandiFeasibilityAwareDecisionFocusedLearning2025}, who adapted the procedure of \citet{elmachtoubSmartPredictThen2022} and \citet{tangPyEPOPyTorchbasedEndend2024} from objective to constraint parameters. We describe it here as it is \emph{implemented}, since the code differs in several places from the description given in \citet{mandiFeasibilityAwareDecisionFocusedLearning2025}'s appendix. We describe each deviation below, so that a reader consulting the code is not surprised by a mismatch. For the BAPP we use the instances of \citet{huTwoStagePredict+OptimizeMixed2023} as published, without generating data ourselves.

\textbf{MDKP.} Each instance is a $0/1$ multi-dimensional knapsack that minimizes the total value $\sum_n (-q_n) x_n$ subject to $M$ capacity constraints $\sum_n \rho_{ni} x_n \le \rho_i$ for N items.
It's feature vector is drawn as $\phi_{\kappa} \sim \mathrm{Uniform}(0,I_{10})$, rather than from the standard normal distribution as stated in the appendix of \cite{mandiFeasibilityAwareDecisionFocusedLearning2025}. A fixed matrix $B$ with independent $\mathrm{Bernoulli}(0.5)$ entries plays the role of the true, unknown data-generating model.

In the uncertain-weights setting, the weight of item $n$ in dimension $i$ of instance $\kappa$ is 
\[
\rho_{ni}^{(\kappa)} =  \frac{1}{3.5^{\mathrm{Deg}}}\, \left[ \left(\tfrac{1}{\sqrt{10}}\, \phi_{\kappa}^{\top} B[i,n] + 3\right)^{\mathrm{Deg}} + 1 \right]  \xi_{ni}^{(\kappa)},
\]
where $\mathrm{Deg}$ controls how nonlinear the true mapping is, so that a larger $\mathrm{Deg}$ makes a linear predictor increasingly misspecified, $\xi_{ni}^{(\kappa)} \sim U[1-w, 1+w]$ is multiplicative noise, and 10 is the number of features. We use $\mathrm{Deg} = 6$ and $w = 0.25$. Note that the constant $1$ enters before the division by $3.5^{\mathrm{Deg}}$, which is different to the description in the appendix of \cite{mandiFeasibilityAwareDecisionFocusedLearning2025}. 
The capacity of dimension $i$ is
\[
\rho_i^{(\kappa)} = \left\lceil 10\, r_i \left( \sum_{n,p} B[i,n,p] \right) \varepsilon_i^{(\kappa)} \right\rceil, \qquad r = (0.18,\, 0.2,\, 0.22),
\]
with $\varepsilon_i^{(\kappa)} \sim U[1-w, 1+w]$ drawn per instance and dimension. Unlike the weights, the capacities do not depend on the features: their base value is identical across instances and varies only through the noise term. The capacities are therefore a fraction of the sum of the raw entries of $B$, rather than of the maximal achievable weight as described, and the fraction is dimension-specific rather than a single value $0.2$.

In the uncertain-capacities setting, the capacities are generated from the features by the same functional form, except that $B$ has one dimension fewer, of shape $(M, 10)$ rather than $(M, N, 10)$, so that it yields a single value per dimension rather than one per item. The features are rescaled by $0.5\mathcal{I}$ after the capacities have been generated from them, with the optimization model receiving the rescaled features, whereas the capacities themselves are rescaled by $5\mathcal{I}$ and also clipped at $5\mathcal{I}$. The written procedure in \citep{mandiFeasibilityAwareDecisionFocusedLearning2025} applies the single factor $0.5\mathcal{I}$ to both.

\subsection{Hyperparameters}\label{qslo::app::hyperparameters}
As described in Section~\ref{QSLO::sec::experimentalsettings}, we fix the ensemble size of all tree-based models for comparability and otherwise keep the library defaults, rather than tuning each method individually. The exact settings and package versions are listed in Table~\ref{qslo::app::hyperparameters_table}. The settings of ODECE and the MSE baseline follow \citet{mandiFeasibilityAwareDecisionFocusedLearning2025} and are listed in Table~\ref{qslo::app::odece_hyperparameters_table}.

\begin{table}[ht!]
\centering
\small
\begin{tabularx}{\textwidth}{@{}>{\raggedright\arraybackslash}p{2.4cm}>{\raggedright\arraybackslash}X>{\raggedright\arraybackslash}p{3.4cm}>{\raggedright\arraybackslash}p{4.4cm}@{}}
\toprule
\textbf{Model} & \textbf{Package (version)} & \textbf{Hyperparameter} & \textbf{Value} \\
\midrule
SLO LR & \texttt{sklearn.linear\_model.} \texttt{LinearRegression} (1.6.1) & all & default \\
\midrule
QSLO-P LR & \texttt{sklearn.linear\_model.} \texttt{QuantileRegressor} (1.6.1) & \texttt{alpha} (L1 penalty) & 0 (MDKP-W, MDKP-C), 1.0 (BAPP) \\
 & & \texttt{solver} & \texttt{highs} \\
\midrule
SLO \& QSLO-P GBM & \texttt{sklearn.ensemble.} \texttt{GradientBoosting}\allowbreak\texttt{Regressor} (1.6.1) & \texttt{n\_estimators} & 500 \\
 & & \texttt{max\_depth} & 3 (default) \\
 & & \texttt{learning\_rate} & 0.1 (default) \\
 & & \texttt{loss} & \texttt{squared\_error} (SLO), \texttt{quantile} (QSLO-P) \\
\midrule
SLO RF & \texttt{sklearn.ensemble.} \texttt{RandomForest}\allowbreak\texttt{Regressor} (1.6.1) & \texttt{n\_estimators} & 500 \\
 & & \texttt{max\_depth} & 100 \\
\midrule
QSLO-P RF & \texttt{sklearn\_quantile.} \texttt{RandomForest}\allowbreak\texttt{Quantile}\allowbreak\texttt{Regressor} (0.1.1) & \texttt{n\_estimators} & 500 \\
 & & \texttt{max\_depth} & 100 \\
\midrule
SLO \& QSLO-D NGBoost & \texttt{ngboost.} \texttt{NGBRegressor} (0.5.8) & \texttt{n\_estimators} & 500 \\
 & & \texttt{learning\_rate} & 0.01 (default) \\
 & & \texttt{Base} & \texttt{DecisionTree}\allowbreak\texttt{Regressor}, \texttt{max\_depth} 3 (default) \\
 & & \texttt{dist} & Log-Normal \\
\midrule
SLO \& QSLO-D TabPFN & \texttt{tabpfn.} \texttt{TabPFNRegressor} (2.5) & \texttt{checkpoint} & \texttt{tabpfn-v2.5-}\allowbreak\texttt{regressor-}\allowbreak\texttt{v2.5\_default} \\
 & & \texttt{fit\_mode} & \texttt{fit\_}\allowbreak\texttt{preprocessors} \\
 & & \texttt{memory\_saving\_}\allowbreak\texttt{mode} & \texttt{auto} \\
 & & \texttt{ignore\_}\allowbreak\texttt{pretraining\_}\allowbreak\texttt{limits} & \texttt{True} \\
\bottomrule
\end{tabularx}
\caption{Hyperparameter settings and package information for all regression models.}
\label{qslo::app::hyperparameters_table}
\end{table}

\begin{table}[ht!]
\centering
\small
\begin{tabularx}{\textwidth}{@{}>{\raggedright\arraybackslash}p{3.6cm}>{\raggedright\arraybackslash}X>{\raggedright\arraybackslash}X>{\raggedright\arraybackslash}X@{}}
\toprule
\textbf{Hyperparameter} & \textbf{MDKP-W} & \textbf{MDKP-C} & \textbf{BAPP} \\
\midrule
Predictor & linear, \texttt{nn.Linear} (10 features $\to$ $50 \times 3$ weights) & linear, \texttt{nn.Linear} (10 features $\to$ 3 capacities) & MLP, one hidden layer with 512 ReLU units (4,096 features $\to$ 1) \\
\midrule
Optimizer & Adam & Adam & Adam \\
Learning rate & 0.05 & 0.005 & 0.001 \\
Epochs & 20 & 20 & 20 \\
Batch size & 32 & 32 & 32 \\
\midrule
Infeasibility aversion coefficient $\alpha$ (ODECE) & \multicolumn{3}{l}{$\{0, 0.1, \dots, 1.0\}$ (11 values)} \\
Margin threshold (ODECE) & 2.0 & 2.0 & 2.0 \\
\midrule
Packages & \multicolumn{3}{l}{\texttt{torch} 2.6.0, \texttt{lightning} 2.5.0} \\
\bottomrule
\end{tabularx}
\caption{Hyperparameter settings of ODECE and the MSE baseline per problem. The MSE baseline uses the same predictor and training settings with an MSE loss~\citep{mandiFeasibilityAwareDecisionFocusedLearning2025}.}
\label{qslo::app::odece_hyperparameters_table}
\end{table}

\subsection{Empirical Coverage}\label{qslo::app::coverage}

Empirical coverage at nominal level $\tau$ is the fraction of test instances whose true value lies at or
below the predicted $\tau$-quantile. A model predicting the true conditional quantile
places exactly that fraction of the true values at or below it, so plotting the observed
coverage against the requested level traces the diagonal for a well-calibrated model and
departs from it when the predicted quantiles are too high or too low. Coverage is a
marginal statistic, however, and says nothing about whether the predicted quantiles are
placed well for any individual instance. What it does show is whether moving $\tau$
shifts the predicted values at all, which is the minimum condition for using it to
traverse the trade-off.

\begin{figure}[ht!]
    \centering
    \includegraphics[width=\linewidth]{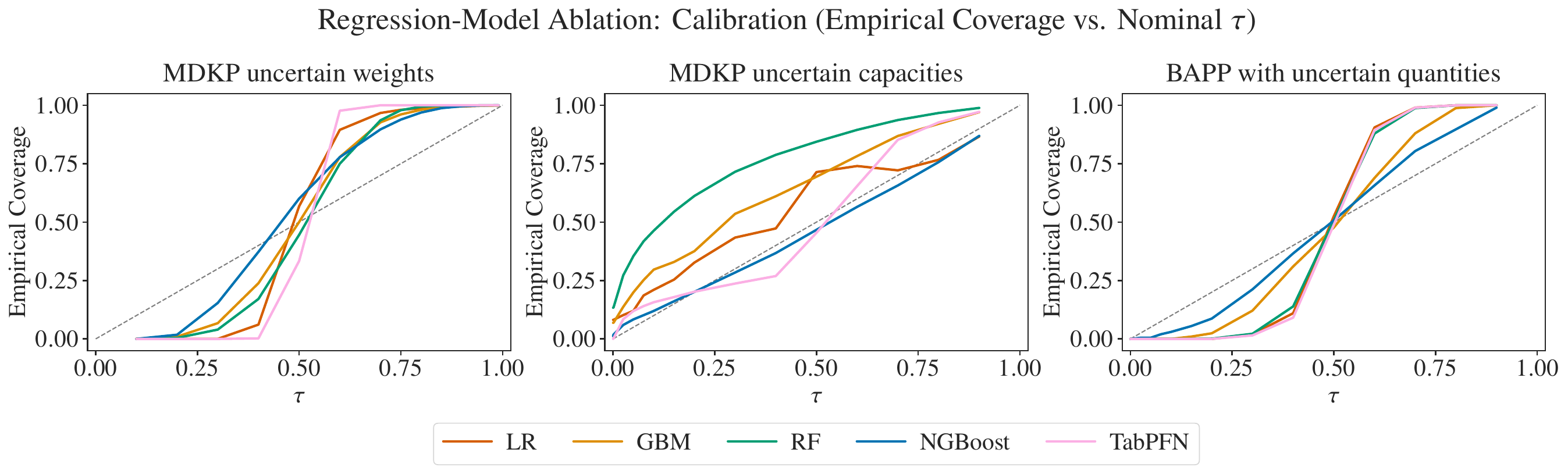}
    \caption{Empirical coverage of the candidate regressors against the nominal quantile level $\tau$, per problem, pooled over all uncertain parameters of a problem. A well-calibrated model follows the dashed diagonal.}\label{qslo::fig::ablation_calibration}
\end{figure}

Figure~\ref{qslo::fig::ablation_calibration} shows that for every model and problem the coverage rises from about $0$ to close to $1$ as $\tau$ increases, so $\tau$ shifts the predictions over the whole range and can be used to traverse the trade-off. The curves do not follow the diagonal, however. On the MDKP with uncertain weights and on the BAPP they are S-shaped and steeper than the diagonal around $\tau = 0.5$, meaning that the predicted quantiles lie closer together than the true ones. On the MDKP with uncertain capacities, NGBoost and TabPFN stay closest to the diagonal, while RF lies above it for every $\tau$, so its predicted quantiles are too high.

\subsection{Statistical Tests}\label{qslo::app::stattests}

We compare the methods at operating points selected on validation data. For each problem, run, and method, we select the quantile level $\tau$ (QSLO) or the infeasibility aversion coefficient $\alpha$ (ODECE) with the lowest validation post-hoc regret, and pair the test post-hoc regrets of the selected configurations instance by instance. On the MDKP, every run generates its own instances, so we pool the five runs into $2{,}500$ independent pairs per comparison. The BAPP uses one fixed data split for all runs, so its $100$ test instances recur in every run, and we first average each instance's post-hoc regret over the five runs and use the $100$ instances as units. We test the mean difference with a two-sided paired $t$-test, report a $95\%$ bootstrap confidence interval for it, and control the family-wise error rate with the Holm correction~\citep{holm1979simple}, applied separately to the comparisons with ODECE and with SLO. We test the mean rather than the median because the post-hoc regret is strongly skewed and the tables report means. Table~\ref{tab:stat-tests} lists the results, compared with ODECE and with SLO using the same regression model.

\begin{table}[ht]
\centering
\footnotesize
\caption{Paired tests at operating points selected on validation data. Difference of the mean test post-hoc regret, QSLO minus the reference method, with $95\%$ bootstrap confidence interval and Holm-corrected $p$-value of a two-sided paired $t$-test. Negative differences favor QSLO. Differences significant at the $5\%$ level after the Holm correction are set in bold. The confidence intervals are not adjusted for multiple comparisons. Pairs per test: $2{,}500$ on the MDKP (five runs pooled) and $100$ on the BAPP (one shared test set, each instance averaged over the five runs).}
\label{tab:stat-tests}
\resizebox{\textwidth}{!}{%
\begin{tabular}{llrrrr}
\toprule
 & & \multicolumn{2}{c}{QSLO vs ODECE} & \multicolumn{2}{c}{QSLO vs SLO (same model)} \\
\cmidrule(lr){3-4}\cmidrule(lr){5-6}
Problem & Model & $\Delta$ mean [95\% CI] & $p_{\text{Holm}}$ & $\Delta$ mean [95\% CI] & $p_{\text{Holm}}$ \\
\midrule
\multirow{5}{*}{MDKP-W} & LR & {\boldmath$-0.020$ $[-0.022, -0.018]$} & {\boldmath$<0.001$} & {\boldmath$-0.002$ $[-0.004, -0.001]$} & {\boldmath$0.002$} \\
 & GBM & {\boldmath$-0.019$ $[-0.021, -0.017]$} & {\boldmath$<0.001$} & $-0.001$ $[-0.002, +0.001]$ & $0.637$ \\
 & RF & {\boldmath$-0.003$ $[-0.005, -0.001]$} & {\boldmath$0.012$} & $0.000$ $[-0.001, +0.001]$ & $0.911$ \\
 & NGBoost & {\boldmath$-0.008$ $[-0.010, -0.006]$} & {\boldmath$<0.001$} & $+0.001$ $[0.000, +0.002]$ & $0.637$ \\
 & TabPFN & {\boldmath$-0.027$ $[-0.029, -0.026]$} & {\boldmath$<0.001$} & {\boldmath$+0.004$ $[+0.003, +0.005]$} & {\boldmath$<0.001$} \\
\midrule
\multirow{5}{*}{MDKP-C} & LR & {\boldmath$-0.034$ $[-0.039, -0.029]$} & {\boldmath$<0.001$} & {\boldmath$-0.015$ $[-0.018, -0.012]$} & {\boldmath$<0.001$} \\
 & GBM & {\boldmath$-0.048$ $[-0.053, -0.044]$} & {\boldmath$<0.001$} & {\boldmath$-0.016$ $[-0.020, -0.012]$} & {\boldmath$<0.001$} \\
 & RF & {\boldmath$-0.049$ $[-0.054, -0.044]$} & {\boldmath$<0.001$} & {\boldmath$-0.012$ $[-0.015, -0.010]$} & {\boldmath$<0.001$} \\
 & NGBoost & {\boldmath$-0.051$ $[-0.055, -0.047]$} & {\boldmath$<0.001$} & {\boldmath$-0.014$ $[-0.017, -0.012]$} & {\boldmath$<0.001$} \\
 & TabPFN & {\boldmath$-0.057$ $[-0.061, -0.053]$} & {\boldmath$<0.001$} & {\boldmath$-0.011$ $[-0.014, -0.008]$} & {\boldmath$<0.001$} \\
\midrule
\multirow{5}{*}{BAPP} & LR & $-0.006$ $[-0.019, +0.008]$ & $1.000$ & {\boldmath$-0.099$ $[-0.162, -0.043]$} & {\boldmath$0.014$} \\
 & GBM & $-0.015$ $[-0.032, -0.001]$ & $0.347$ & {\boldmath$-0.103$ $[-0.154, -0.059]$} & {\boldmath$<0.001$} \\
 & RF & $-0.004$ $[-0.019, +0.010]$ & $1.000$ & {\boldmath$-0.034$ $[-0.060, -0.011]$} & {\boldmath$0.038$} \\
 & NGBoost & $+0.023$ $[0.000, +0.049]$ & $0.347$ & {\boldmath$-0.034$ $[-0.058, -0.014]$} & {\boldmath$0.020$} \\
 & TabPFN & $+0.005$ $[-0.007, +0.017]$ & $1.000$ & $-0.032$ $[-0.060, -0.006]$ & $0.093$ \\
\bottomrule
\end{tabular}%
}
\end{table}

\subsection{Full Results} \label{qslo:app:fullresults}

Tables~\ref{QSLO::app_WeightsTable_1} to \ref{QSLO::app_AlloyTable_2} report the complete per-configuration results underlying the aggregate metrics and Pareto frontiers in Section~\ref{QSLO::experiments}. For each method, we list the mean and standard deviation over five independent runs of the test-set infeasibility rate, regret, post-hoc regret, and wall, CPU, and GPU time, across all evaluated infeasibility aversion coefficients $\alpha$ (ODECE) and quantile levels $\tau$ (the regression models). Each problem is split across two tables for readability. Tables~\ref{QSLO::app_WeightsTable_1} and \ref{QSLO::app_WeightsTable_2} cover the MDKP with uncertain weights, Tables~\ref{QSLO::app_CapacityTable_1} and \ref{QSLO::app_CapacityTable_2} the MDKP with uncertain capacities, and Tables~\ref{QSLO::app_AlloyTable_1} and \ref{QSLO::app_AlloyTable_2} the BAPP with uncertain ore quantities. For the BAPP, approximately $20\%$ of instances are unsolvable under ODECE(0.0) and are excluded from the regret average, as noted in Section~\ref{QSLO::sec::experimentalsettings}.

\begin{table}[b!]
    \centering
    \small
\footnotesize
\begin{tabular}{{l|cc|cc|cc|rr|rr|rr}}
\hline
Model&\multicolumn{2}{c}{Infeasibility} & \multicolumn{2}{c}{Regret} & \multicolumn{2}{c}{Post-hoc Regret} & \multicolumn{2}{c}{Wall Time (s)} & \multicolumn{2}{c}{CPU Time (s)} & \multicolumn{2}{c}{GPU Time (s)} \\
&mean & std & mean & std & mean & std & mean & std & mean & std & mean & std \\
\toprule
MSE & 0.338 & 0.018 & 0.063 & 0.003 & 0.080 & 0.004 & 616 & 57 & 1,091 & 697 & - & - \\
\midrule\midrule
ODECE(0.0) & 0.996 & 0.003 & 0.000 & 0.000 & 1.061 & 0.018 & 74 & 12 & 31 & 17 & - & - \\
ODECE(0.1) & 0.988 & 0.003 & 0.008 & 0.002 & 0.159 & 0.012 & 5,888 & 380 & 10,745 & 7,378 & - & - \\
ODECE(0.2) & 0.966 & 0.005 & 0.015 & 0.002 & 0.124 & 0.009 & 5,850 & 381 & 10,759 & 7,296 & - & - \\
ODECE(0.3) & 0.858 & 0.036 & 0.023 & 0.003 & 0.094 & 0.008 & 5,700 & 402 & 10,560 & 7,158 & - & - \\
ODECE(0.4) & 0.619 & 0.069 & 0.037 & 0.003 & 0.071 & 0.005 & 5,683 & 422 & 10,464 & 7,108 & - & - \\
ODECE(0.5) & 0.148 & 0.032 & 0.078 & 0.004 & 0.084 & 0.003 & 6,061 & 355 & 10,801 & 7,647 & - & - \\
ODECE(0.6) & 0.078 & 0.021 & 0.110 & 0.011 & 0.112 & 0.010 & 6,045 & 357 & 10,417 & 7,587 & - & - \\
ODECE(0.7) & 0.056 & 0.009 & 0.129 & 0.009 & 0.130 & 0.009 & 5,823 & 430 & 9,888 & 7,319 & - & - \\
ODECE(0.8) & 0.051 & 0.011 & 0.137 & 0.011 & 0.138 & 0.011 & 5,753 & 338 & 9,677 & 7,165 & - & - \\
ODECE(0.9) & 0.050 & 0.011 & 0.140 & 0.013 & 0.141 & 0.013 & 5,627 & 390 & 9,367 & 6,983 & - & - \\
ODECE(1.0) & 0.003 & 0.001 & 0.374 & 0.008 & 0.374 & 0.008 & 4,144 & 248 & 7,493 & 5,133 & - & - \\
\midrule\midrule
LR(0.1) & 0.996 & 0.003 & 0.000 & 0.000 & 0.314 & 0.004 & 71 & 29 & 105 & 6 & - & - \\
LR(0.2) & 0.996 & 0.003 & 0.000 & 0.000 & 0.231 & 0.004 & 102 & 23 & 116 & 7 & - & - \\
LR(0.3) & 0.996 & 0.003 & 0.000 & 0.000 & 0.156 & 0.005 & 102 & 9 & 112 & 3 & - & - \\
LR(0.4) & 0.936 & 0.008 & 0.018 & 0.003 & 0.089 & 0.005 & 88 & 32 & 120 & 14 & - & - \\
LR(0.5) & 0.521 & 0.033 & 0.033 & 0.001 & 0.050 & 0.002 & 104 & 7 & 118 & 6 & - & - \\
LR(0.6) & 0.133 & 0.013 & 0.054 & 0.001 & 0.056 & 0.001 & 95 & 21 & 134 & 8 & - & - \\
LR(0.7) & 0.037 & 0.005 & 0.081 & 0.002 & 0.080 & 0.002 & 80 & 36 & 127 & 9 & - & - \\
LR(0.75) & 0.019 & 0.004 & 0.094 & 0.001 & 0.094 & 0.002 & 50 & 4 & 123 & 16 & - & - \\
LR(0.8) & 0.012 & 0.002 & 0.108 & 0.002 & 0.107 & 0.002 & 68 & 3 & 123 & 12 & - & - \\
LR(0.85) & 0.006 & 0.003 & 0.121 & 0.002 & 0.121 & 0.002 & 177 & 247 & 129 & 7 & - & - \\
LR(0.9) & 0.003 & 0.003 & 0.137 & 0.002 & 0.136 & 0.002 & 128 & 40 & 132 & 14 & - & - \\
LR(0.95) & 0.001 & 0.001 & 0.162 & 0.003 & 0.162 & 0.003 & 173 & 73 & 131 & 11 & - & - \\
LR(0.99) & 0.000 & 0.000 & 0.216 & 0.004 & 0.216 & 0.004 & 163 & 127 & 131 & 5 & - & - \\
LR & 0.500 & 0.027 & 0.034 & 0.001 & 0.054 & 0.003 & 46 & 2 & 119 & 7 & - & - \\
\midrule\midrule
GBM(0.1) & 0.995 & 0.003 & 0.000 & 0.000 & 0.291 & 0.005 & 360 & 101 & 111 & 16 & - & - \\
GBM(0.2) & 0.991 & 0.002 & 0.006 & 0.003 & 0.195 & 0.004 & 385 & 43 & 117 & 13 & - & - \\
GBM(0.3) & 0.953 & 0.012 & 0.018 & 0.002 & 0.128 & 0.004 & 397 & 31 & 130 & 15 & - & - \\
GBM(0.4) & 0.826 & 0.006 & 0.024 & 0.002 & 0.084 & 0.004 & 405 & 23 & 132 & 15 & - & - \\
GBM(0.5) & 0.576 & 0.013 & 0.033 & 0.001 & 0.059 & 0.003 & 389 & 46 & 129 & 19 & - & - \\
GBM(0.6) & 0.276 & 0.013 & 0.047 & 0.001 & 0.053 & 0.002 & 364 & 21 & 136 & 13 & - & - \\
GBM(0.7) & 0.101 & 0.014 & 0.066 & 0.002 & 0.066 & 0.002 & 364 & 22 & 143 & 19 & - & - \\
GBM(0.75) & 0.053 & 0.009 & 0.078 & 0.002 & 0.078 & 0.002 & 352 & 49 & 140 & 24 & - & - \\
GBM(0.8) & 0.028 & 0.002 & 0.092 & 0.001 & 0.091 & 0.002 & 358 & 29 & 142 & 21 & - & - \\
GBM(0.85) & 0.013 & 0.003 & 0.107 & 0.002 & 0.106 & 0.002 & 361 & 24 & 137 & 20 & - & - \\
GBM(0.9) & 0.005 & 0.003 & 0.124 & 0.003 & 0.124 & 0.003 & 256 & 118 & 140 & 13 & - & - \\
GBM(0.95) & 0.002 & 0.002 & 0.158 & 0.003 & 0.158 & 0.003 & 132 & 8 & 140 & 20 & - & - \\
GBM(0.99) & 0.001 & 0.002 & 0.347 & 0.015 & 0.347 & 0.015 & 151 & 9 & 149 & 8 & - & - \\
GBM & 0.579 & 0.019 & 0.034 & 0.002 & 0.054 & 0.003 & 280 & 157 & 136 & 14 & - & - \\
\bottomrule
\end{tabular}

    \caption{Full results for the MDKP with uncertain weights (part 1): infeasibility, regret, post-hoc regret, and wall, CPU and GPU time for MSE, ODECE, LR, and GBM across all evaluated infeasibility aversion coefficients $\alpha$ and quantile levels $\tau$. Lower is better for all metrics.}
    \label{QSLO::app_WeightsTable_1}
\end{table}

\begin{table}[b!]
    \centering
    \small
\footnotesize
\begin{tabular}{{l|cc|cc|cc|rr|rr|rr}}
\hline
Model&\multicolumn{2}{c}{Infeasibility} & \multicolumn{2}{c}{Regret} & \multicolumn{2}{c}{Post-hoc Regret} & \multicolumn{2}{c}{Wall Time (s)} & \multicolumn{2}{c}{CPU Time (s)} & \multicolumn{2}{c}{GPU Time (s)} \\
&mean & std & mean & std & mean & std & mean & std & mean & std & mean & std \\
\toprule
RF(0.1) & 0.995 & 0.003 & 0.000 & 0.000 & 0.416 & 0.006 & 396 & 2 & 460 & 8 & - & - \\
RF(0.2) & 0.994 & 0.003 & 0.003 & 0.006 & 0.282 & 0.006 & 399 & 5 & 468 & 12 & - & - \\
RF(0.3) & 0.973 & 0.008 & 0.021 & 0.005 & 0.189 & 0.005 & 369 & 4 & 442 & 11 & - & - \\
RF(0.4) & 0.867 & 0.013 & 0.032 & 0.004 & 0.121 & 0.004 & 385 & 35 & 467 & 31 & - & - \\
RF(0.5) & 0.603 & 0.016 & 0.045 & 0.002 & 0.078 & 0.004 & 397 & 13 & 483 & 17 & - & - \\
RF(0.6) & 0.287 & 0.017 & 0.063 & 0.002 & 0.069 & 0.004 & 403 & 3 & 493 & 7 & - & - \\
RF(0.7) & 0.080 & 0.012 & 0.090 & 0.003 & 0.089 & 0.003 & 403 & 3 & 495 & 12 & - & - \\
RF(0.75) & 0.033 & 0.007 & 0.107 & 0.003 & 0.106 & 0.003 & 374 & 1 & 464 & 6 & - & - \\
RF(0.8) & 0.010 & 0.003 & 0.127 & 0.003 & 0.127 & 0.003 & 375 & 3 & 468 & 11 & - & - \\
RF(0.85) & 0.002 & 0.002 & 0.151 & 0.004 & 0.151 & 0.004 & 396 & 21 & 494 & 27 & - & - \\
RF(0.9) & 0.001 & 0.002 & 0.181 & 0.004 & 0.181 & 0.004 & 408 & 2 & 508 & 14 & - & - \\
RF(0.95) & 0.000 & 0.001 & 0.223 & 0.005 & 0.223 & 0.005 & 410 & 1 & 514 & 8 & - & - \\
RF(0.99) & 0.000 & 0.000 & 0.294 & 0.006 & 0.294 & 0.006 & 385 & 5 & 487 & 12 & - & - \\
RF & 0.457 & 0.022 & 0.053 & 0.002 & 0.069 & 0.004 & 371 & 3 & 459 & 12 & - & - \\
\midrule\midrule
NGBoost(0.1) & 0.996 & 0.003 & 0.000 & 0.000 & 0.260 & 0.004 & 314 & 14 & 187 & 9 & - & - \\
NGBoost(0.2) & 0.986 & 0.005 & 0.010 & 0.005 & 0.167 & 0.003 & 372 & 94 & 197 & 5 & - & - \\
NGBoost(0.3) & 0.895 & 0.013 & 0.024 & 0.003 & 0.109 & 0.003 & 408 & 69 & 193 & 10 & - & - \\
NGBoost(0.4) & 0.685 & 0.007 & 0.034 & 0.002 & 0.076 & 0.004 & 454 & 38 & 200 & 8 & - & - \\
NGBoost(0.5) & 0.444 & 0.017 & 0.047 & 0.002 & 0.064 & 0.003 & 413 & 70 & 198 & 11 & - & - \\
NGBoost(0.6) & 0.238 & 0.016 & 0.061 & 0.003 & 0.065 & 0.003 & 440 & 78 & 199 & 13 & - & - \\
NGBoost(0.7) & 0.105 & 0.005 & 0.078 & 0.002 & 0.078 & 0.003 & 391 & 59 & 209 & 12 & - & - \\
NGBoost(0.75) & 0.069 & 0.014 & 0.089 & 0.003 & 0.088 & 0.003 & 375 & 43 & 218 & 19 & - & - \\
NGBoost(0.8) & 0.037 & 0.005 & 0.100 & 0.003 & 0.099 & 0.003 & 351 & 54 & 209 & 13 & - & - \\
NGBoost(0.85) & 0.018 & 0.003 & 0.114 & 0.003 & 0.114 & 0.003 & 319 & 17 & 215 & 7 & - & - \\
NGBoost(0.9) & 0.007 & 0.005 & 0.132 & 0.003 & 0.131 & 0.003 & 342 & 66 & 215 & 12 & - & - \\
NGBoost(0.95) & 0.002 & 0.002 & 0.157 & 0.004 & 0.157 & 0.004 & 343 & 49 & 216 & 8 & - & - \\
NGBoost(0.99) & 0.000 & 0.000 & 0.202 & 0.004 & 0.202 & 0.004 & 395 & 29 & 222 & 11 & - & - \\
NGBoost & 0.443 & 0.017 & 0.047 & 0.002 & 0.064 & 0.003 & 306 & 8 & 202 & 15 & - & - \\
\midrule\midrule
TabPFN(0.1) & 0.996 & 0.003 & 0.000 & 0.000 & 0.305 & 0.003 & 156 & 9 & 240 & 5 & 156 & 9 \\
TabPFN(0.2) & 0.996 & 0.003 & 0.000 & 0.000 & 0.242 & 0.004 & 165 & 13 & 258 & 12 & 164 & 13 \\
TabPFN(0.3) & 0.996 & 0.003 & 0.000 & 0.000 & 0.177 & 0.003 & 163 & 10 & 262 & 20 & 163 & 10 \\
TabPFN(0.4) & 0.991 & 0.003 & 0.006 & 0.006 & 0.106 & 0.004 & 156 & 4 & 273 & 13 & 156 & 4 \\
TabPFN(0.5) & 0.684 & 0.042 & 0.023 & 0.001 & 0.044 & 0.002 & 159 & 7 & 276 & 6 & 158 & 7 \\
TabPFN(0.6) & 0.069 & 0.005 & 0.047 & 0.001 & 0.047 & 0.001 & 180 & 12 & 277 & 13 & 180 & 12 \\
TabPFN(0.7) & 0.002 & 0.003 & 0.077 & 0.001 & 0.077 & 0.001 & 184 & 24 & 274 & 8 & 184 & 24 \\
TabPFN(0.75) & 0.000 & 0.001 & 0.091 & 0.001 & 0.091 & 0.001 & 175 & 23 & 266 & 13 & 175 & 23 \\
TabPFN(0.8) & 0.000 & 0.001 & 0.105 & 0.001 & 0.105 & 0.001 & 158 & 11 & 278 & 15 & 158 & 11 \\
TabPFN(0.85) & 0.000 & 0.000 & 0.117 & 0.002 & 0.117 & 0.002 & 151 & 4 & 283 & 19 & 151 & 4 \\
TabPFN(0.9) & 0.000 & 0.000 & 0.130 & 0.002 & 0.130 & 0.002 & 176 & 8 & 278 & 15 & 175 & 8 \\
TabPFN(0.95) & 0.000 & 0.000 & 0.146 & 0.002 & 0.146 & 0.002 & 173 & 10 & 283 & 6 & 172 & 10 \\
TabPFN(0.99) & 0.000 & 0.000 & 0.170 & 0.003 & 0.170 & 0.003 & 171 & 17 & 279 & 13 & 171 & 17 \\
TabPFN & 0.619 & 0.029 & 0.024 & 0.001 & 0.041 & 0.001 & 167 & 11 & 266 & 10 & 167 & 11 \\
\bottomrule
\end{tabular}

    \caption{Full results for the MDKP with uncertain weights (part 2): infeasibility, regret, post-hoc regret, and wall, CPU and GPU time for RF, NGBoost, and TabPFN across all evaluated quantile levels $\tau$. Lower is better for all metrics.}
    \label{QSLO::app_WeightsTable_2}
\end{table}

\begin{table}[b!]
    \centering
    \small
\footnotesize
\begin{tabular}{{l|cc|cc|cc|rr|rr|rr}}
\hline
Model&\multicolumn{2}{c}{Infeasibility} & \multicolumn{2}{c}{Regret} & \multicolumn{2}{c}{Post-hoc Regret} & \multicolumn{2}{c}{Wall Time (s)} & \multicolumn{2}{c}{CPU Time (s)} & \multicolumn{2}{c}{GPU Time (s)} \\
&mean & std & mean & std & mean & std & mean & std & mean & std & mean & std \\
\toprule
MSE & 0.563 & 0.074 & 0.130 & 0.017 & 0.162 & 0.014 & 522 & 73 & 985 & 627 & - & - \\
\midrule\midrule
ODECE(0.0) & 1.000 & 0.001 & 0.206 & - & 1.554 & 0.267 & 4,505 & 332 & 8,813 & 5,513 & - & - \\
ODECE(0.1) & 0.950 & 0.030 & 0.066 & 0.019 & 0.347 & 0.021 & 5,735 & 771 & 10,832 & 7,157 & - & - \\
ODECE(0.2) & 0.797 & 0.172 & 0.087 & 0.026 & 0.229 & 0.039 & 4,944 & 927 & 9,327 & 6,311 & - & - \\
ODECE(0.3) & 0.599 & 0.318 & 0.133 & 0.078 & 0.192 & 0.043 & 5,204 & 1,190 & 9,751 & 6,829 & - & - \\
ODECE(0.4) & 0.229 & 0.155 & 0.179 & 0.069 & 0.174 & 0.067 & 4,546 & 1,047 & 8,268 & 5,922 & - & - \\
ODECE(0.5) & 0.046 & 0.013 & 0.267 & 0.030 & 0.260 & 0.030 & 3,840 & 712 & 7,028 & 4,932 & - & - \\
ODECE(0.6) & 0.045 & 0.028 & 0.293 & 0.045 & 0.286 & 0.047 & 3,730 & 722 & 6,759 & 4,807 & - & - \\
ODECE(0.7) & 0.047 & 0.021 & 0.307 & 0.037 & 0.299 & 0.037 & 3,584 & 677 & 6,462 & 4,612 & - & - \\
ODECE(0.8) & 0.036 & 0.019 & 0.323 & 0.038 & 0.316 & 0.038 & 3,546 & 714 & 6,376 & 4,592 & - & - \\
ODECE(0.9) & 0.035 & 0.019 & 0.332 & 0.043 & 0.325 & 0.043 & 3,219 & 572 & 5,791 & 4,129 & - & - \\
ODECE(1.0) & 0.001 & 0.002 & 0.924 & 0.040 & 0.924 & 0.040 & 605 & 214 & 1,071 & 825 & - & - \\
\midrule\midrule
LR(0.0001) & 0.018 & 0.005 & 0.210 & 0.020 & 0.207 & 0.019 & 47 & 6 & 102 & 16 & - & - \\
LR(0.001) & 0.018 & 0.005 & 0.210 & 0.020 & 0.207 & 0.019 & 45 & 3 & 99 & 15 & - & - \\
LR(0.025) & 0.051 & 0.012 & 0.178 & 0.021 & 0.172 & 0.020 & 43 & 2 & 97 & 11 & - & - \\
LR(0.05) & 0.091 & 0.016 & 0.158 & 0.026 & 0.151 & 0.023 & 43 & 3 & 94 & 17 & - & - \\
LR(0.075) & 0.131 & 0.020 & 0.147 & 0.027 & 0.140 & 0.023 & 43 & 4 & 95 & 15 & - & - \\
LR(0.1) & 0.161 & 0.029 & 0.137 & 0.028 & 0.132 & 0.023 & 45 & 4 & 96 & 18 & - & - \\
LR(0.15) & 0.225 & 0.052 & 0.121 & 0.032 & 0.123 & 0.022 & 45 & 5 & 92 & 15 & - & - \\
LR(0.2) & 0.305 & 0.061 & 0.107 & 0.036 & 0.117 & 0.021 & 42 & 5 & 94 & 20 & - & - \\
LR(0.3) & 0.438 & 0.071 & 0.085 & 0.037 & 0.116 & 0.014 & 43 & 6 & 98 & 25 & - & - \\
LR(0.4) & 0.552 & 0.075 & 0.066 & 0.037 & 0.129 & 0.009 & 47 & 4 & 109 & 23 & - & - \\
LR(0.5) & 0.592 & 0.099 & 0.048 & 0.035 & 0.150 & 0.009 & 45 & 7 & 106 & 30 & - & - \\
LR(0.6) & 0.638 & 0.107 & 0.035 & 0.028 & 0.164 & 0.013 & 48 & 9 & 113 & 30 & - & - \\
LR(0.7) & 0.692 & 0.135 & 0.025 & 0.023 & 0.185 & 0.022 & 47 & 8 & 115 & 32 & - & - \\
LR(0.8) & 0.747 & 0.164 & 0.015 & 0.017 & 0.214 & 0.039 & 49 & 7 & 122 & 28 & - & - \\
LR(0.9) & 0.759 & 0.195 & 0.007 & 0.013 & 0.256 & 0.062 & 52 & 9 & 134 & 36 & - & - \\
LR & 0.640 & 0.073 & 0.062 & 0.021 & 0.131 & 0.015 & 54 & 3 & 121 & 25 & - & - \\
\midrule\midrule
GBM(0.0001) & 0.003 & 0.003 & 0.444 & 0.028 & 0.443 & 0.028 & 46 & 4 & 94 & 8 & - & - \\
GBM(0.001) & 0.003 & 0.003 & 0.444 & 0.028 & 0.443 & 0.028 & 55 & 15 & 90 & 15 & - & - \\
GBM(0.025) & 0.113 & 0.064 & 0.167 & 0.091 & 0.165 & 0.092 & 79 & 22 & 101 & 24 & - & - \\
GBM(0.05) & 0.149 & 0.054 & 0.128 & 0.043 & 0.125 & 0.040 & 94 & 47 & 97 & 17 & - & - \\
GBM(0.075) & 0.216 & 0.041 & 0.115 & 0.039 & 0.113 & 0.033 & 61 & 17 & 106 & 23 & - & - \\
GBM(0.1) & 0.236 & 0.042 & 0.109 & 0.041 & 0.108 & 0.034 & 123 & 64 & 112 & 19 & - & - \\
GBM(0.15) & 0.283 & 0.031 & 0.098 & 0.039 & 0.101 & 0.029 & 107 & 55 & 113 & 24 & - & - \\
GBM(0.2) & 0.342 & 0.042 & 0.090 & 0.039 & 0.099 & 0.028 & 144 & 45 & 110 & 30 & - & - \\
GBM(0.3) & 0.406 & 0.065 & 0.076 & 0.039 & 0.102 & 0.024 & 148 & 40 & 115 & 20 & - & - \\
GBM(0.4) & 0.470 & 0.063 & 0.061 & 0.037 & 0.107 & 0.024 & 166 & 49 & 113 & 23 & - & - \\
GBM(0.5) & 0.521 & 0.072 & 0.047 & 0.035 & 0.118 & 0.021 & 197 & 65 & 122 & 27 & - & - \\
GBM(0.6) & 0.544 & 0.074 & 0.037 & 0.031 & 0.131 & 0.028 & 176 & 73 & 123 & 29 & - & - \\
GBM(0.7) & 0.592 & 0.096 & 0.025 & 0.029 & 0.159 & 0.019 & 160 & 80 & 126 & 29 & - & - \\
GBM(0.8) & 0.630 & 0.120 & 0.020 & 0.028 & 0.182 & 0.026 & 80 & 16 & 131 & 32 & - & - \\
GBM(0.9) & 0.662 & 0.147 & 0.007 & 0.013 & 0.227 & 0.050 & 81 & 28 & 137 & 33 & - & - \\
GBM & 0.606 & 0.063 & 0.065 & 0.026 & 0.118 & 0.028 & 54 & 10 & 116 & 31 & - & - \\
\bottomrule
\end{tabular}

    \caption{Full results for the MDKP with uncertain capacities (part 1): infeasibility, regret, post-hoc regret, and wall, CPU and GPU time for all methods across all evaluated infeasibility aversion coefficients $\alpha$ and quantile levels $\tau$. Lower is better for all metrics.}

    \label{QSLO::app_CapacityTable_1}
\end{table}

\begin{table}[b!]
    \centering
    \small
\footnotesize
\begin{tabular}{{l|cc|cc|cc|rr|rr|rr}}
\hline
Model&\multicolumn{2}{c}{Infeasibility} & \multicolumn{2}{c}{Regret} & \multicolumn{2}{c}{Post-hoc Regret} & \multicolumn{2}{c}{Wall Time (s)} & \multicolumn{2}{c}{CPU Time (s)} & \multicolumn{2}{c}{GPU Time (s)} \\
&mean & std & mean & std & mean & std & mean & std & mean & std & mean & std \\
\toprule
RF(0.0001) & 0.003 & 0.003 & 0.284 & 0.018 & 0.284 & 0.018 & 29 & 1 & 98 & 8 & - & - \\
RF(0.001) & 0.003 & 0.003 & 0.284 & 0.018 & 0.284 & 0.018 & 29 & 2 & 97 & 9 & - & - \\
RF(0.025) & 0.019 & 0.007 & 0.190 & 0.015 & 0.189 & 0.015 & 30 & 4 & 102 & 17 & - & - \\
RF(0.05) & 0.040 & 0.009 & 0.158 & 0.024 & 0.156 & 0.024 & 29 & 3 & 97 & 13 & - & - \\
RF(0.075) & 0.072 & 0.015 & 0.140 & 0.029 & 0.138 & 0.028 & 30 & 3 & 102 & 13 & - & - \\
RF(0.1) & 0.104 & 0.019 & 0.127 & 0.033 & 0.127 & 0.030 & 31 & 4 & 106 & 19 & - & - \\
RF(0.15) & 0.165 & 0.028 & 0.108 & 0.036 & 0.112 & 0.031 & 31 & 3 & 108 & 14 & - & - \\
RF(0.2) & 0.221 & 0.037 & 0.094 & 0.036 & 0.104 & 0.030 & 32 & 3 & 111 & 16 & - & - \\
RF(0.3) & 0.309 & 0.042 & 0.072 & 0.038 & 0.099 & 0.026 & 34 & 5 & 119 & 22 & - & - \\
RF(0.4) & 0.392 & 0.049 & 0.056 & 0.035 & 0.105 & 0.024 & 34 & 6 & 119 & 26 & - & - \\
RF(0.5) & 0.459 & 0.057 & 0.044 & 0.035 & 0.117 & 0.026 & 36 & 7 & 129 & 31 & - & - \\
RF(0.6) & 0.512 & 0.066 & 0.030 & 0.027 & 0.131 & 0.028 & 36 & 6 & 128 & 25 & - & - \\
RF(0.7) & 0.567 & 0.087 & 0.020 & 0.022 & 0.150 & 0.032 & 37 & 6 & 130 & 27 & - & - \\
RF(0.8) & 0.613 & 0.112 & 0.013 & 0.018 & 0.177 & 0.041 & 37 & 7 & 132 & 30 & - & - \\
RF(0.9) & 0.655 & 0.145 & 0.007 & 0.012 & 0.208 & 0.052 & 39 & 7 & 138 & 30 & - & - \\
RF & 0.454 & 0.076 & 0.047 & 0.031 & 0.113 & 0.026 & 35 & 7 & 125 & 29 & - & - \\
\midrule\midrule
NGBoost(0.0001) & 0.011 & 0.011 & 0.291 & 0.103 & 0.290 & 0.103 & 43 & 7 & 88 & 19 & - & - \\
NGBoost(0.001) & 0.021 & 0.017 & 0.235 & 0.084 & 0.233 & 0.084 & 42 & 2 & 90 & 19 & - & - \\
NGBoost(0.025) & 0.095 & 0.036 & 0.146 & 0.053 & 0.142 & 0.050 & 43 & 4 & 97 & 17 & - & - \\
NGBoost(0.05) & 0.150 & 0.033 & 0.127 & 0.048 & 0.124 & 0.042 & 43 & 3 & 98 & 19 & - & - \\
NGBoost(0.075) & 0.184 & 0.034 & 0.115 & 0.046 & 0.114 & 0.037 & 44 & 3 & 103 & 18 & - & - \\
NGBoost(0.1) & 0.212 & 0.043 & 0.105 & 0.042 & 0.108 & 0.033 & 45 & 5 & 105 & 22 & - & - \\
NGBoost(0.15) & 0.273 & 0.042 & 0.093 & 0.039 & 0.101 & 0.029 & 45 & 5 & 101 & 20 & - & - \\
NGBoost(0.2) & 0.332 & 0.046 & 0.085 & 0.037 & 0.099 & 0.026 & 44 & 6 & 100 & 22 & - & - \\
NGBoost(0.3) & 0.422 & 0.051 & 0.072 & 0.034 & 0.100 & 0.024 & 47 & 6 & 111 & 24 & - & - \\
NGBoost(0.4) & 0.516 & 0.053 & 0.063 & 0.030 & 0.105 & 0.024 & 47 & 5 & 114 & 25 & - & - \\
NGBoost(0.5) & 0.612 & 0.067 & 0.055 & 0.027 & 0.113 & 0.025 & 47 & 8 & 110 & 25 & - & - \\
NGBoost(0.6) & 0.685 & 0.059 & 0.048 & 0.024 & 0.125 & 0.027 & 46 & 5 & 110 & 21 & - & - \\
NGBoost(0.7) & 0.777 & 0.073 & 0.042 & 0.021 & 0.140 & 0.029 & 47 & 5 & 115 & 22 & - & - \\
NGBoost(0.8) & 0.859 & 0.074 & 0.035 & 0.014 & 0.163 & 0.033 & 47 & 5 & 115 & 25 & - & - \\
NGBoost(0.9) & 0.935 & 0.051 & 0.029 & 0.014 & 0.197 & 0.040 & 48 & 7 & 120 & 30 & - & - \\
NGBoost & 0.611 & 0.067 & 0.055 & 0.027 & 0.113 & 0.025 & 56 & 2 & 125 & 25 & - & - \\
\midrule\midrule
TabPFN(0.0001) & 0.000 & 0.000 & 0.515 & 0.047 & 0.515 & 0.047 & 21 & 3 & 63 & 13 & 21 & 3 \\
TabPFN(0.001) & 0.002 & 0.003 & 0.352 & 0.045 & 0.352 & 0.046 & 27 & 5 & 82 & 16 & 26 & 5 \\
TabPFN(0.025) & 0.042 & 0.011 & 0.147 & 0.037 & 0.144 & 0.036 & 37 & 5 & 112 & 16 & 36 & 5 \\
TabPFN(0.05) & 0.087 & 0.015 & 0.126 & 0.041 & 0.122 & 0.038 & 34 & 6 & 111 & 19 & 34 & 6 \\
TabPFN(0.075) & 0.129 & 0.018 & 0.115 & 0.043 & 0.111 & 0.037 & 33 & 5 & 110 & 23 & 32 & 5 \\
TabPFN(0.1) & 0.161 & 0.027 & 0.107 & 0.043 & 0.104 & 0.036 & 34 & 5 & 113 & 19 & 34 & 5 \\
TabPFN(0.15) & 0.209 & 0.044 & 0.094 & 0.043 & 0.098 & 0.033 & 32 & 5 & 105 & 20 & 31 & 5 \\
TabPFN(0.2) & 0.252 & 0.049 & 0.084 & 0.044 & 0.094 & 0.032 & 33 & 6 & 109 & 21 & 32 & 6 \\
TabPFN(0.3) & 0.329 & 0.064 & 0.068 & 0.045 & 0.093 & 0.027 & 34 & 7 & 117 & 29 & 34 & 7 \\
TabPFN(0.4) & 0.394 & 0.066 & 0.053 & 0.043 & 0.098 & 0.023 & 41 & 5 & 144 & 22 & 41 & 5 \\
TabPFN(0.5) & 0.459 & 0.069 & 0.041 & 0.040 & 0.108 & 0.022 & 39 & 7 & 136 & 27 & 39 & 7 \\
TabPFN(0.6) & 0.515 & 0.078 & 0.029 & 0.032 & 0.122 & 0.025 & 39 & 9 & 136 & 34 & 38 & 9 \\
TabPFN(0.7) & 0.571 & 0.099 & 0.020 & 0.026 & 0.139 & 0.031 & 39 & 5 & 135 & 19 & 39 & 5 \\
TabPFN(0.8) & 0.623 & 0.126 & 0.013 & 0.020 & 0.160 & 0.041 & 41 & 7 & 136 & 29 & 41 & 7 \\
TabPFN(0.9) & 0.694 & 0.149 & 0.007 & 0.012 & 0.186 & 0.052 & 43 & 10 & 141 & 37 & 43 & 10 \\
TabPFN & 0.448 & 0.079 & 0.043 & 0.033 & 0.103 & 0.025 & 40 & 8 & 131 & 31 & 40 & 8 \\
\bottomrule
\end{tabular}

    \caption{Full results for the MDKP with uncertain capacities (part 2): infeasibility, regret, post-hoc regret, and wall, CPU and GPU time for all methods across all evaluated infeasibility aversion coefficients $\alpha$ and quantile levels $\tau$. Lower is better for all metrics.}

    \label{QSLO::app_CapacityTable_2}
\end{table}

\begin{table}[b!]
    \centering   
    \begin{minipage}{\textwidth}
\footnotesize
\begin{tabular}{{l|cc|cc|cc|rr|rr|rr}}
\hline
Model&\multicolumn{2}{c}{Infeasibility} & \multicolumn{2}{c}{Regret} & \multicolumn{2}{c}{Post-hoc Regret} & \multicolumn{2}{c}{Wall Time (s)} & \multicolumn{2}{c}{CPU Time (s)} & \multicolumn{2}{c}{GPU Time (s)} \\
&mean & std & mean & std & mean & std & mean & std & mean & std & mean & std \\
\toprule
MSE & 0.523 & 0.011 & 0.171 & 0.005 & 0.216 & 0.003 & 55 & 10 & 37 & 22 & - & - \\
\midrule\midrule
ODECE(0.0)\footnote{20\% of the infeasible instances are unsolvable.} & 1.000 & - & - & - & 61.212 & 34.068 & 97 & 85 & 60 & 80 & - & - \\
ODECE(0.1) & 0.885 & 0.040 & 0.034 & 0.010 & 0.169 & 0.007 & 167 & 18 & 240 & 133 & - & - \\
ODECE(0.2) & 0.807 & 0.010 & 0.055 & 0.013 & 0.172 & 0.004 & 205 & 42 & 244 & 140 & - & - \\
ODECE(0.3) & 0.696 & 0.074 & 0.111 & 0.036 & 0.184 & 0.013 & 185 & 11 & 260 & 150 & - & - \\
ODECE(0.4) & 0.526 & 0.084 & 0.211 & 0.045 & 0.245 & 0.026 & 148 & 6 & 238 & 141 & - & - \\
ODECE(0.5) & 0.374 & 0.041 & 0.302 & 0.050 & 0.316 & 0.032 & 168 & 31 & 241 & 143 & - & - \\
ODECE(0.6) & 0.359 & 0.053 & 0.317 & 0.034 & 0.322 & 0.024 & 181 & 41 & 243 & 143 & - & - \\
ODECE(0.7) & 0.258 & 0.075 & 0.415 & 0.068 & 0.400 & 0.066 & 188 & 25 & 249 & 147 & - & - \\
ODECE(0.8) & 0.132 & 0.030 & 0.557 & 0.073 & 0.534 & 0.068 & 216 & 30 & 266 & 158 & - & - \\
ODECE(0.9) & 0.091 & 0.025 & 0.616 & 0.051 & 0.588 & 0.054 & 174 & 16 & 256 & 152 & - & - \\
ODECE(1.0) & 0.000 & 0.000 & 180.58 & 397.95 & 180.58 & 397.95 & 141 & 57 & 217 & 159 & - & - \\
\midrule\midrule
LR(0.0001) & 0.000 & 0.000 & 0.759 & 0.000 & 0.759 & 0.000 & 11 & 1 & 2 & 0 & - & - \\
LR(0.001) & 0.000 & 0.000 & 0.759 & 0.000 & 0.759 & 0.000 & 12 & 1 & 2 & 0 & - & - \\
LR(0.025) & 0.000 & 0.000 & 0.753 & 0.000 & 0.753 & 0.000 & 10 & 1 & 2 & 0 & - & - \\
LR(0.05) & 0.014 & 0.000 & 0.747 & 0.000 & 0.738 & 0.000 & 10 & 0 & 2 & 0 & - & - \\
LR(0.075) & 0.065 & 0.000 & 0.692 & 0.000 & 0.671 & 0.000 & 12 & 2 & 2 & 0 & - & - \\
LR(0.1) & 0.073 & 0.000 & 0.635 & 0.000 & 0.613 & 0.000 & 11 & 1 & 2 & 0 & - & - \\
LR(0.15) & 0.073 & 0.000 & 0.521 & 0.000 & 0.505 & 0.000 & 11 & 1 & 2 & 0 & - & - \\
LR(0.2) & 0.130 & 0.000 & 0.472 & 0.000 & 0.457 & 0.000 & 11 & 1 & 2 & 0 & - & - \\
LR(0.3) & 0.250 & 0.000 & 0.321 & 0.000 & 0.323 & 0.000 & 13 & 3 & 2 & 0 & - & - \\
LR(0.4) & 0.399 & 0.000 & 0.243 & 0.000 & 0.260 & 0.000 & 11 & 0 & 2 & 0 & - & - \\
LR(0.5) & 0.530 & 0.000 & 0.173 & 0.000 & 0.227 & 0.000 & 12 & 1 & 2 & 0 & - & - \\
LR(0.6) & 0.699 & 0.000 & 0.135 & 0.000 & 0.189 & 0.000 & 12 & 2 & 2 & 0 & - & - \\
LR(0.7) & 0.760 & 0.000 & 0.064 & 0.000 & 0.164 & 0.000 & 11 & 1 & 2 & 0 & - & - \\
LR(0.8) & 0.825 & 0.000 & 0.033 & 0.000 & 0.166 & 0.000 & 13 & 3 & 2 & 0 & - & - \\
LR(0.9) & 0.977 & 0.000 & 0.048 & 0.000 & 0.198 & 0.000 & 14 & 4 & 2 & 0 & - & - \\
LR & 0.663 & 0.000 & 0.232 & 0.000 & 0.266 & 0.000 & 161 & 12 & 5 & 0 & - & - \\
\midrule\midrule
GBM(0.0001) & 0.000 & 0.000 & 0.759 & 0.000 & 0.759 & 0.000 & 4 & 0 & 2 & 0 & - & - \\
GBM(0.001) & 0.000 & 0.000 & 0.759 & 0.000 & 0.759 & 0.000 & 4 & 0 & 2 & 0 & - & - \\
GBM(0.025) & 0.000 & 0.000 & 0.753 & 0.000 & 0.753 & 0.000 & 57 & 0 & 2 & 0 & - & - \\
GBM(0.05) & 0.014 & 0.000 & 0.739 & 0.001 & 0.730 & 0.001 & 169 & 0 & 3 & 0 & - & - \\
GBM(0.075) & 0.075 & 0.014 & 0.635 & 0.001 & 0.618 & 0.006 & 346 & 7 & 5 & 1 & - & - \\
GBM(0.1) & 0.137 & 0.017 & 0.495 & 0.013 & 0.486 & 0.008 & 348 & 5 & 5 & 1 & - & - \\
GBM(0.15) & 0.233 & 0.023 & 0.367 & 0.009 & 0.366 & 0.010 & 349 & 8 & 5 & 1 & - & - \\
GBM(0.2) & 0.304 & 0.021 & 0.316 & 0.008 & 0.317 & 0.004 & 344 & 7 & 4 & 0 & - & - \\
GBM(0.3) & 0.441 & 0.027 & 0.241 & 0.004 & 0.250 & 0.006 & 345 & 6 & 4 & 0 & - & - \\
GBM(0.4) & 0.512 & 0.028 & 0.212 & 0.009 & 0.220 & 0.005 & 350 & 3 & 6 & 0 & - & - \\
GBM(0.5) & 0.591 & 0.018 & 0.183 & 0.008 & 0.218 & 0.004 & 347 & 5 & 6 & 0 & - & - \\
GBM(0.6) & 0.649 & 0.016 & 0.160 & 0.006 & 0.204 & 0.005 & 351 & 3 & 5 & 1 & - & - \\
GBM(0.7) & 0.703 & 0.017 & 0.133 & 0.005 & 0.198 & 0.016 & 351 & 7 & 4 & 0 & - & - \\
GBM(0.8) & 0.778 & 0.024 & 0.084 & 0.010 & 0.174 & 0.005 & 346 & 6 & 4 & 0 & - & - \\
GBM(0.9) & 0.869 & 0.009 & 0.025 & 0.003 & 0.158 & 0.010 & 341 & 9 & 3 & 0 & - & - \\
GBM & 0.638 & 0.018 & 0.210 & 0.003 & 0.258 & 0.003 & 341 & 6 & 4 & 0 & - & - \\
\bottomrule
\end{tabular}

    \caption{Full results for the BAPP with uncertain ore quantities (part 1): infeasibility, regret, post-hoc regret, and wall, CPU and GPU time for all methods across all evaluated infeasibility aversion coefficients $\alpha$ and quantile levels $\tau$. Lower is better for all metrics.}
    \label{QSLO::app_AlloyTable_1}
    \end{minipage}
\end{table}

\begin{table}[b!]
    \centering   
    \begin{minipage}{\textwidth}
\footnotesize
\begin{tabular}{{l|cc|cc|cc|rr|rr|rr}}
\hline
Model&\multicolumn{2}{c}{Infeasibility} & \multicolumn{2}{c}{Regret} & \multicolumn{2}{c}{Post-hoc Regret} & \multicolumn{2}{c}{Wall Time (s)} & \multicolumn{2}{c}{CPU Time (s)} & \multicolumn{2}{c}{GPU Time (s)} \\
&mean & std & mean & std & mean & std & mean & std & mean & std & mean & std \\
\toprule
RF(0.0001) & 0.000 & 0.000 & 0.759 & 0.000 & 0.759 & 0.000 & 696 & 1 & 3 & 0 & - & - \\
RF(0.001) & 0.000 & 0.000 & 0.759 & 0.000 & 0.759 & 0.000 & 699 & 2 & 3 & 0 & - & - \\
RF(0.025) & 0.006 & 0.008 & 0.756 & 0.006 & 0.753 & 0.002 & 698 & 2 & 3 & 0 & - & - \\
RF(0.05) & 0.039 & 0.004 & 0.729 & 0.004 & 0.715 & 0.004 & 696 & 2 & 3 & 0 & - & - \\
RF(0.075) & 0.056 & 0.009 & 0.665 & 0.005 & 0.647 & 0.003 & 701 & 6 & 4 & 0 & - & - \\
RF(0.1) & 0.080 & 0.007 & 0.606 & 0.008 & 0.585 & 0.006 & 696 & 0 & 3 & 0 & - & - \\
RF(0.15) & 0.090 & 0.004 & 0.511 & 0.004 & 0.494 & 0.002 & 696 & 1 & 3 & 0 & - & - \\
RF(0.2) & 0.150 & 0.012 & 0.442 & 0.004 & 0.421 & 0.003 & 699 & 2 & 4 & 0 & - & - \\
RF(0.3) & 0.259 & 0.028 & 0.307 & 0.004 & 0.313 & 0.004 & 698 & 3 & 3 & 0 & - & - \\
RF(0.4) & 0.407 & 0.011 & 0.230 & 0.002 & 0.250 & 0.005 & 696 & 1 & 4 & 1 & - & - \\
RF(0.5) & 0.597 & 0.014 & 0.189 & 0.006 & 0.211 & 0.008 & 695 & 0 & 4 & 0 & - & - \\
RF(0.6) & 0.699 & 0.016 & 0.133 & 0.006 & 0.182 & 0.004 & 694 & 2 & 3 & 0 & - & - \\
RF(0.7) & 0.785 & 0.015 & 0.069 & 0.004 & 0.179 & 0.008 & 695 & 2 & 3 & 0 & - & - \\
RF(0.8) & 0.858 & 0.011 & 0.046 & 0.007 & 0.165 & 0.002 & 696 & 1 & 3 & 0 & - & - \\
RF(0.9) & 0.940 & 0.010 & 0.013 & 0.012 & 0.184 & 0.005 & 695 & 2 & 4 & 0 & - & - \\
RF & 0.576 & 0.010 & 0.183 & 0.006 & 0.204 & 0.007 & 700 & 7 & 5 & 1 & - & - \\
\midrule\midrule
NGBoost(0.0001) & 0.008 & 0.000 & 0.808 & 0.002 & 0.804 & 0.002 & 710 & 21 & 12 & 0 & - & - \\
NGBoost(0.001) & 0.081 & 0.003 & 0.634 & 0.002 & 0.617 & 0.001 & 707 & 6 & 12 & 0 & - & - \\
NGBoost(0.025) & 0.256 & 0.000 & 0.406 & 0.001 & 0.400 & 0.001 & 701 & 7 & 11 & 0 & - & - \\
NGBoost(0.05) & 0.301 & 0.010 & 0.356 & 0.003 & 0.359 & 0.002 & 692 & 3 & 9 & 0 & - & - \\
NGBoost(0.075) & 0.328 & 0.003 & 0.321 & 0.003 & 0.333 & 0.001 & 725 & 10 & 10 & 0 & - & - \\
NGBoost(0.1) & 0.338 & 0.003 & 0.309 & 0.003 & 0.313 & 0.001 & 743 & 31 & 10 & 0 & - & - \\
NGBoost(0.15) & 0.381 & 0.015 & 0.278 & 0.004 & 0.292 & 0.002 & 728 & 23 & 10 & 0 & - & - \\
NGBoost(0.2) & 0.421 & 0.006 & 0.259 & 0.002 & 0.277 & 0.002 & 712 & 14 & 11 & 1 & - & - \\
NGBoost(0.3) & 0.504 & 0.007 & 0.227 & 0.005 & 0.257 & 0.002 & 714 & 27 & 10 & 0 & - & - \\
NGBoost(0.4) & 0.591 & 0.010 & 0.224 & 0.004 & 0.235 & 0.003 & 697 & 12 & 10 & 1 & - & - \\
NGBoost(0.5) & 0.668 & 0.008 & 0.203 & 0.006 & 0.230 & 0.003 & 706 & 13 & 11 & 0 & - & - \\
NGBoost(0.6) & 0.675 & 0.017 & 0.170 & 0.004 & 0.216 & 0.006 & 698 & 14 & 11 & 1 & - & - \\
NGBoost(0.7) & 0.670 & 0.010 & 0.139 & 0.002 & 0.204 & 0.004 & 697 & 10 & 10 & 0 & - & - \\
NGBoost(0.8) & 0.720 & 0.009 & 0.117 & 0.003 & 0.198 & 0.001 & 705 & 16 & 11 & 0 & - & - \\
NGBoost(0.9) & 0.772 & 0.008 & 0.097 & 0.002 & 0.192 & 0.002 & 730 & 14 & 12 & 0 & - & - \\
NGBoost & 0.668 & 0.008 & 0.203 & 0.006 & 0.230 & 0.003 & 704 & 5 & 11 & 0 & - & - \\
\midrule\midrule
TabPFN(0.0001) & 0.000 & 0.000 & 0.844 & 0.000 & 0.844 & 0.000 & 69 & 14 & 167 & 40 & 69 & 14 \\
TabPFN(0.001) & 0.000 & 0.000 & 0.816 & 0.000 & 0.816 & 0.000 & 67 & 9 & 154 & 24 & 67 & 9 \\
TabPFN(0.025) & 0.003 & 0.006 & 0.761 & 0.004 & 0.759 & 0.001 & 69 & 5 & 164 & 14 & 69 & 5 \\
TabPFN(0.05) & 0.060 & 0.004 & 0.746 & 0.002 & 0.723 & 0.000 & 72 & 6 & 169 & 12 & 72 & 6 \\
TabPFN(0.075) & 0.079 & 0.000 & 0.701 & 0.003 & 0.673 & 0.003 & 73 & 9 & 180 & 26 & 73 & 9 \\
TabPFN(0.1) & 0.087 & 0.000 & 0.652 & 0.001 & 0.626 & 0.001 & 76 & 8 & 186 & 23 & 75 & 8 \\
TabPFN(0.15) & 0.087 & 0.000 & 0.556 & 0.002 & 0.539 & 0.002 & 73 & 7 & 178 & 22 & 73 & 7 \\
TabPFN(0.2) & 0.106 & 0.006 & 0.476 & 0.003 & 0.460 & 0.002 & 66 & 5 & 155 & 10 & 65 & 5 \\
TabPFN(0.3) & 0.241 & 0.011 & 0.343 & 0.003 & 0.340 & 0.003 & 69 & 5 & 163 & 17 & 69 & 5 \\
TabPFN(0.4) & 0.380 & 0.022 & 0.249 & 0.004 & 0.257 & 0.003 & 80 & 9 & 183 & 17 & 79 & 9 \\
TabPFN(0.5) & 0.532 & 0.004 & 0.176 & 0.003 & 0.209 & 0.002 & 72 & 5 & 170 & 16 & 72 & 5 \\
TabPFN(0.6) & 0.702 & 0.007 & 0.141 & 0.004 & 0.184 & 0.004 & 75 & 9 & 182 & 27 & 75 & 9 \\
TabPFN(0.7) & 0.756 & 0.004 & 0.079 & 0.001 & 0.175 & 0.003 & 69 & 8 & 165 & 29 & 69 & 8 \\
TabPFN(0.8) & 0.846 & 0.007 & 0.031 & 0.007 & 0.171 & 0.003 & 77 & 6 & 192 & 17 & 77 & 6 \\
TabPFN(0.9) & 0.963 & 0.000 & 0.009 & 0.000 & 0.178 & 0.000 & 77 & 5 & 188 & 14 & 76 & 5 \\
TabPFN & 0.527 & 0.008 & 0.178 & 0.004 & 0.212 & 0.001 & 68 & 5 & 155 & 12 & 68 & 5 \\
\bottomrule
\end{tabular}

    \caption{Full results for the BAPP with uncertain ore quantities (part 2): infeasibility, regret, post-hoc regret, and wall, CPU and GPU time for all methods across all evaluated infeasibility aversion coefficients $\alpha$ and quantile levels $\tau$. Lower is better for all metrics.}
    \label{QSLO::app_AlloyTable_2}
    \end{minipage}
\end{table}

\end{document}